%% file: main.tex
\documentclass[10pt,twocolumn,letterpaper]{article}

\usepackage{cvpr}              
\usepackage[labelfont=bf]{caption}

\input{preamble}

\definecolor{cvprblue}{rgb}{0.21,0.49,0.74}
\usepackage[pagebackref,breaklinks,colorlinks,allcolors=cvprblue]{hyperref}

\def\paperID{14544} 
\def\confName{CVPR}
\def\confYear{2025}

\title{CARL: A Framework for  Equivariant Image Registration}

\author{
Hastings Greer\textsuperscript{1}$\;$
Lin Tian\textsuperscript{1}$\;$
Fran\c{c}ois-Xavier Vialard\textsuperscript{2,3}$\;$
Roland Kwitt\textsuperscript{4}$\;$
Raúl San José Estépar\textsuperscript{5}\\
Marc Niethammer\textsuperscript{6} \vspace{5pt}\\ 
\textsuperscript{1}UNC Chapel Hill $\;$
\textsuperscript{2}LIGM, Universit\'e Gustave Eiffel$\;$
\textsuperscript{3}MOKAPLAN, INRIA Paris \\
\textsuperscript{4}University of Salzburg $\;$
\textsuperscript{5}Harvard Medical School $\;$
\textsuperscript{6}UC San Diego$\;$
}

\begin{document}
\maketitle
\input{sec/0_abstract}
\vskip -2em

\begin{figure}[htp]
	\centering

	\begin{small}
		\begin{tabular}{cc}
			Moving Image                                                                                                                                           & Warped (CARL) \\
			\includegraphics[width=.44\columnwidth, trim={76cm 16cm 18cm 37cm}, clip]{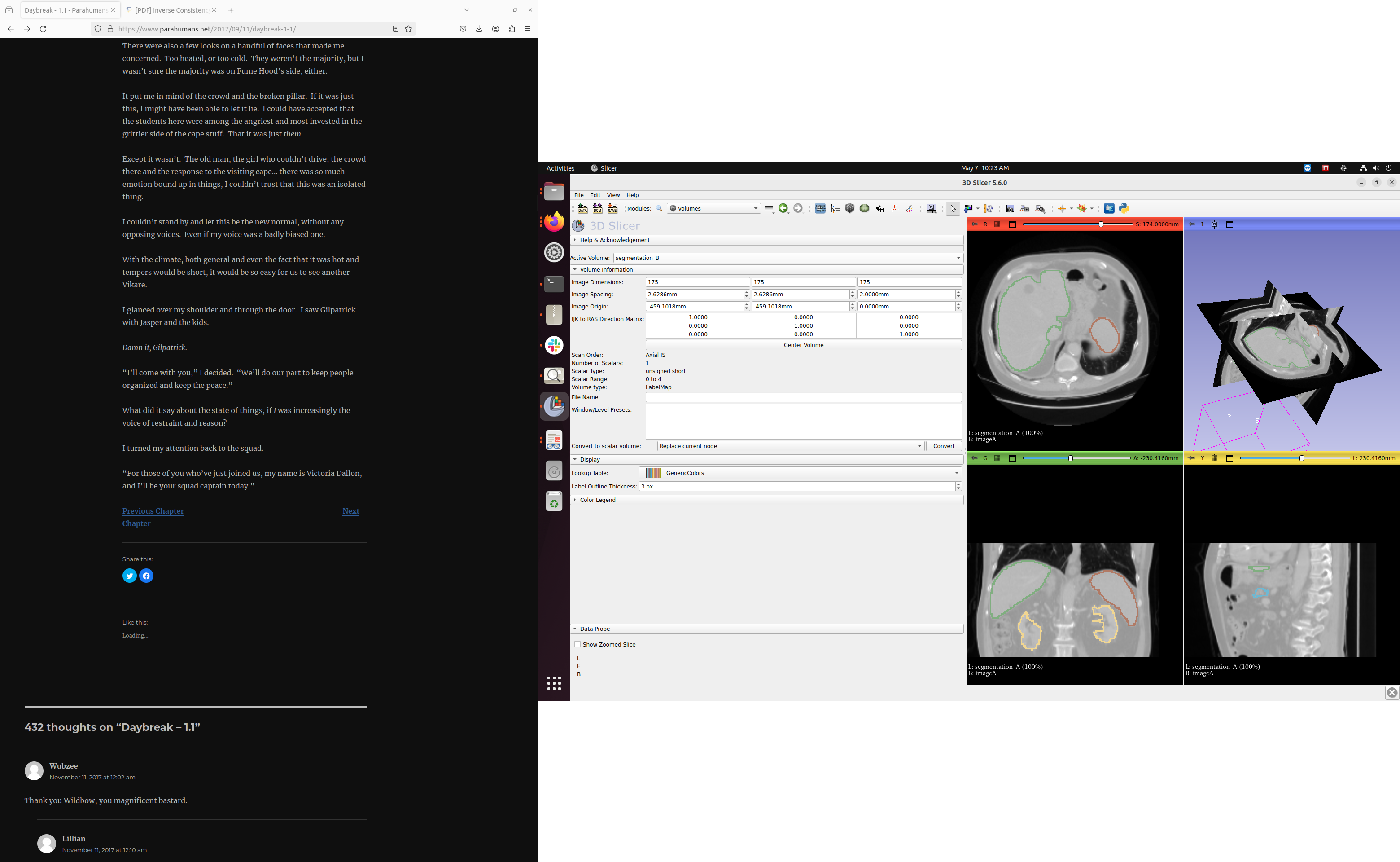}   &
			\includegraphics[width=.44\columnwidth, trim={76cm 16cm 18cm 37cm}, clip]{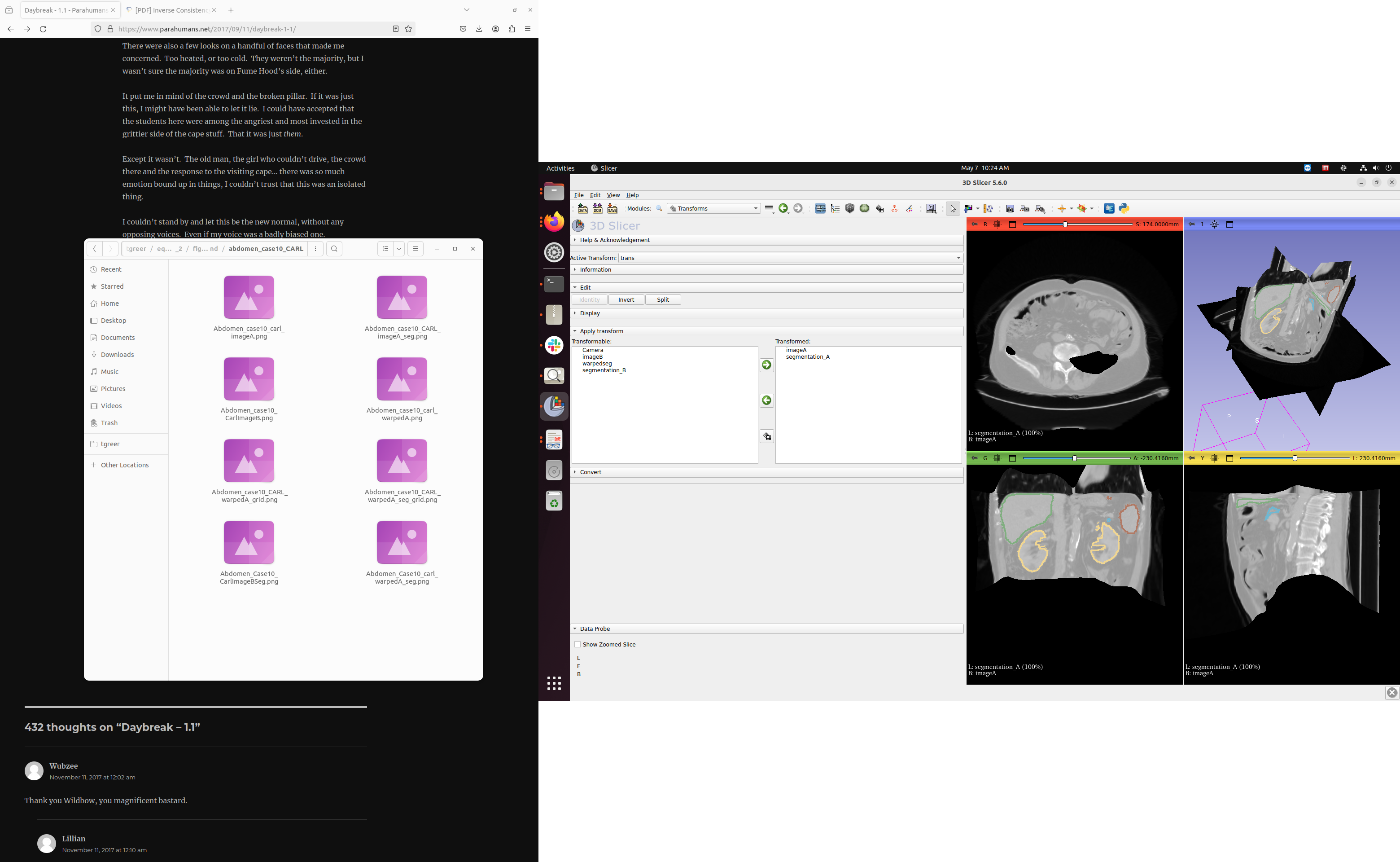}                  \\
			Grid (CARL)                                                                                                                                            & Fixed Image   \\
			\includegraphics[width=.44\columnwidth, trim={76cm 16cm 18cm 37cm}, clip]{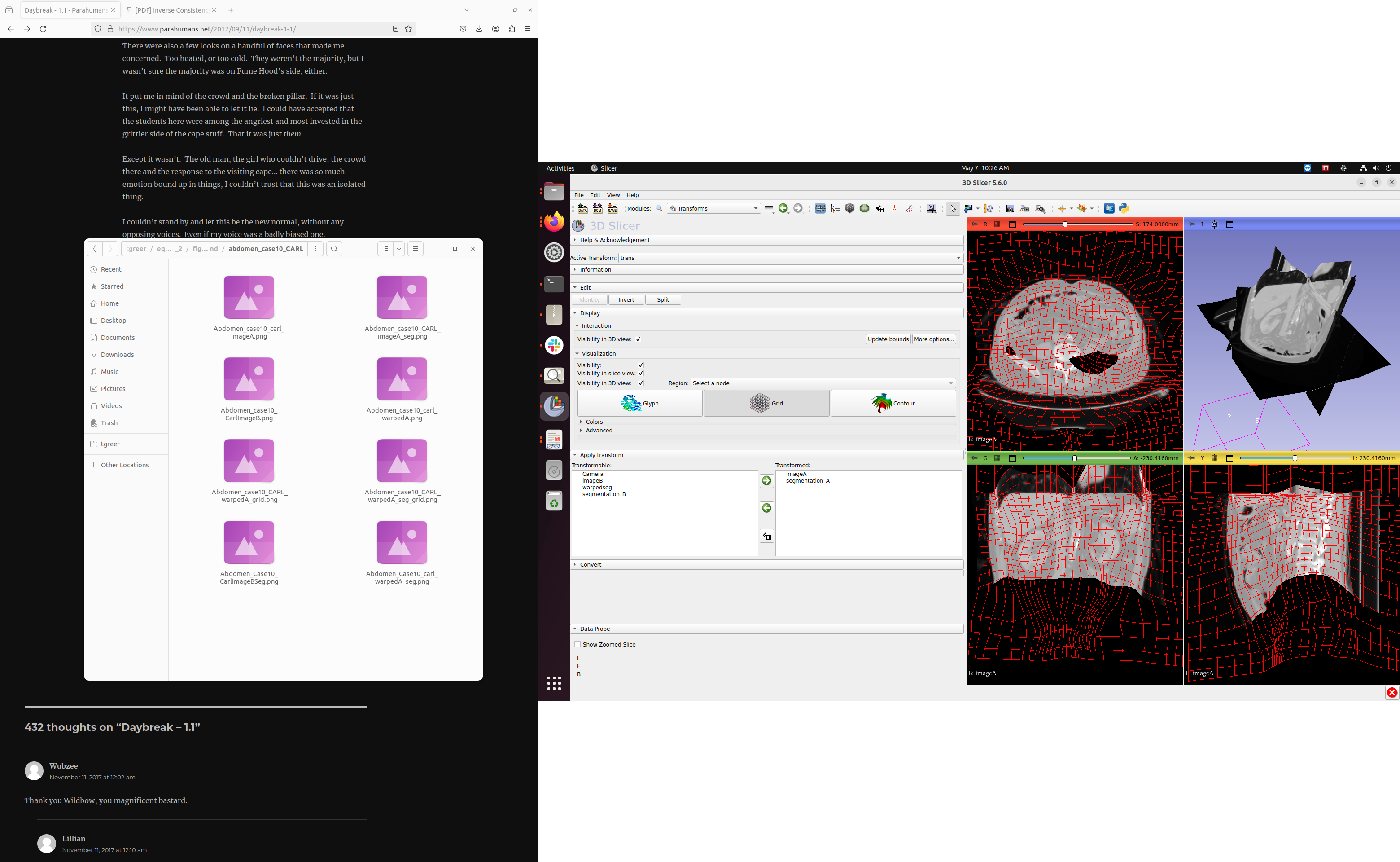} &
			\includegraphics[width=.44\columnwidth, trim={76cm 16cm 18cm 37cm}, clip] {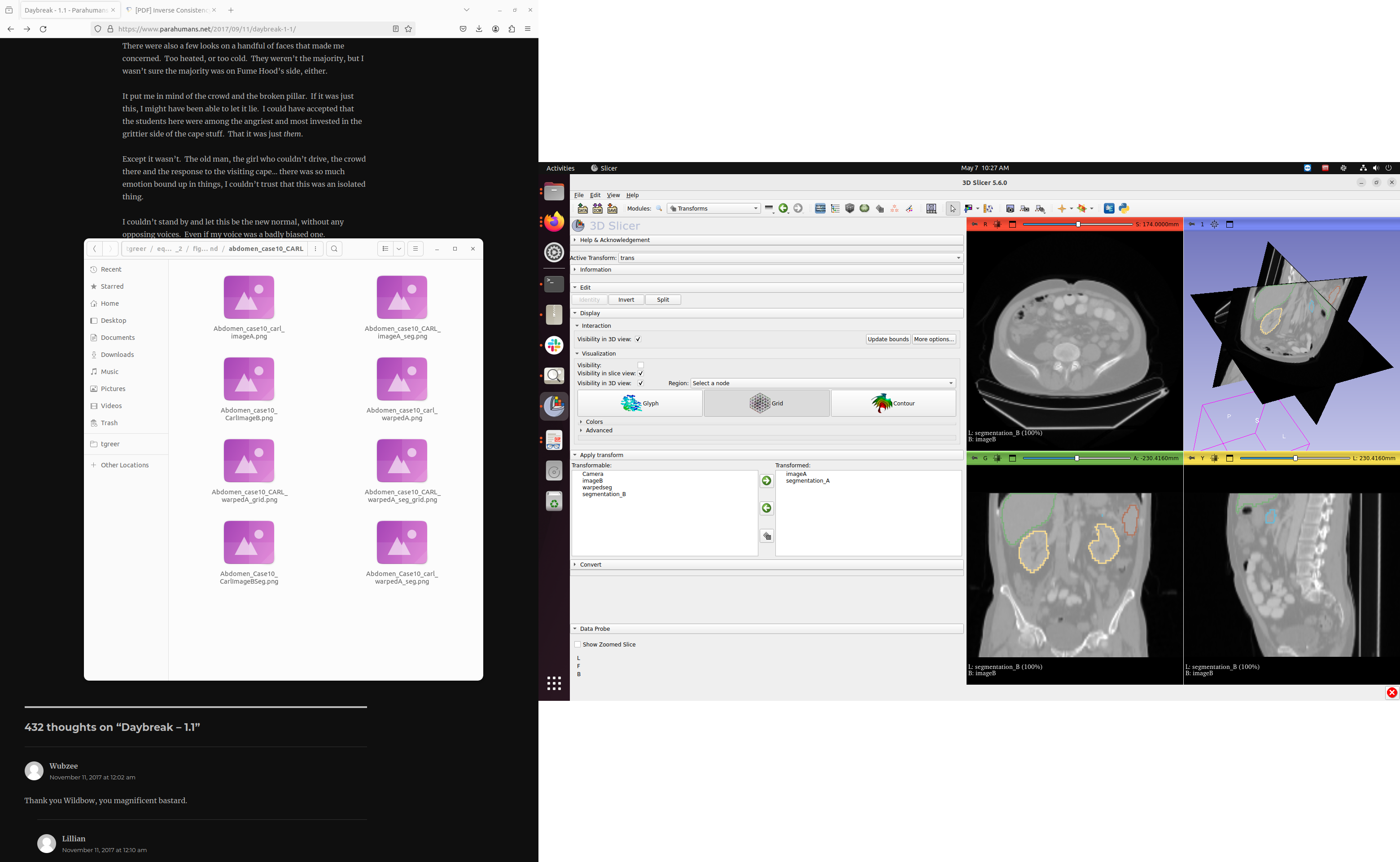} \\

			Moving Image                                                                                                                                           & Warped (CARL\{ROT\}) \\
			\includegraphics[width=.44\columnwidth, trim={75cm 14.4cm 20.4cm 41.3cm}, clip]{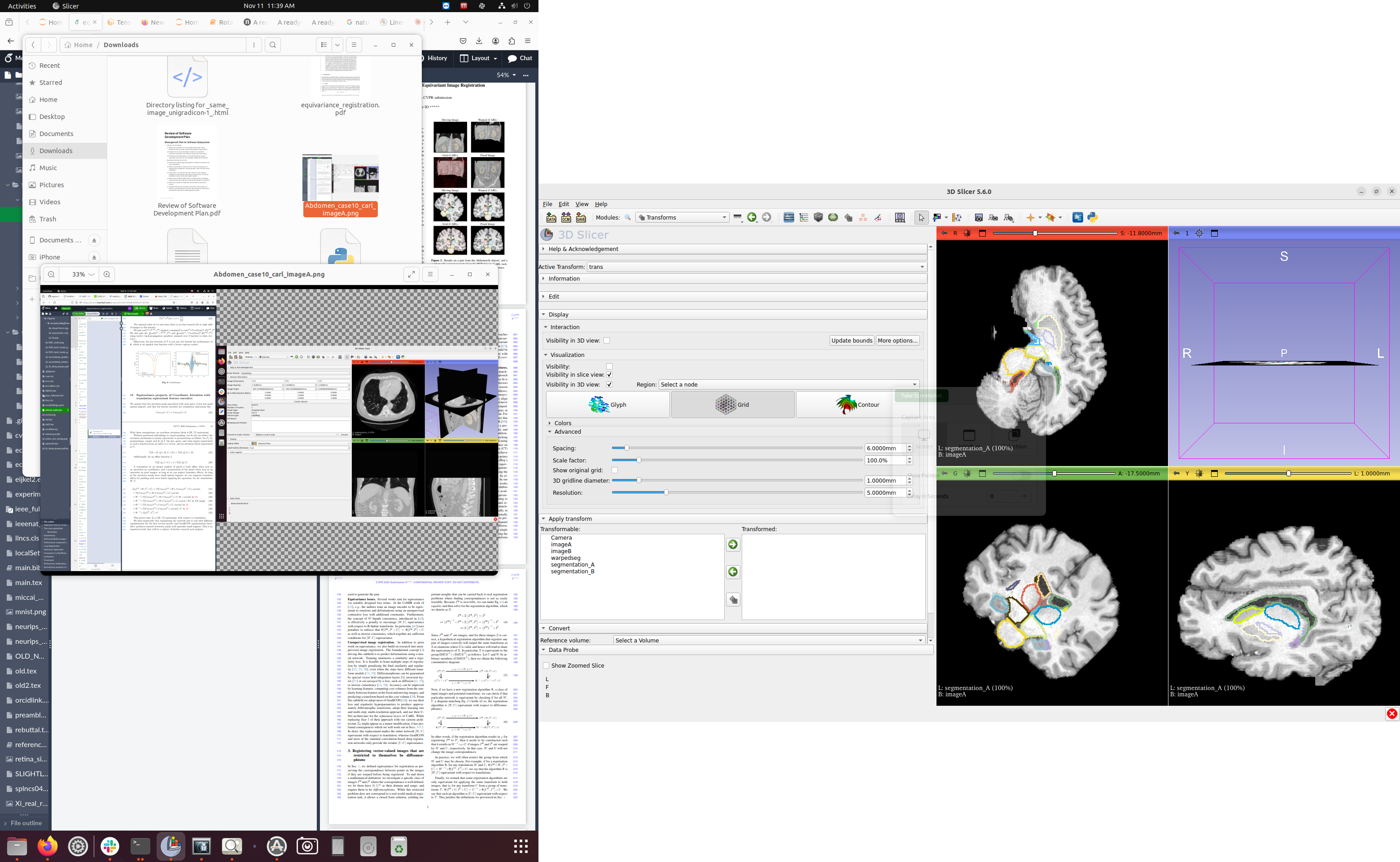}   &
			\includegraphics[width=.44\columnwidth, trim={75cm 14.4cm 20.4cm 41.3cm}, clip]{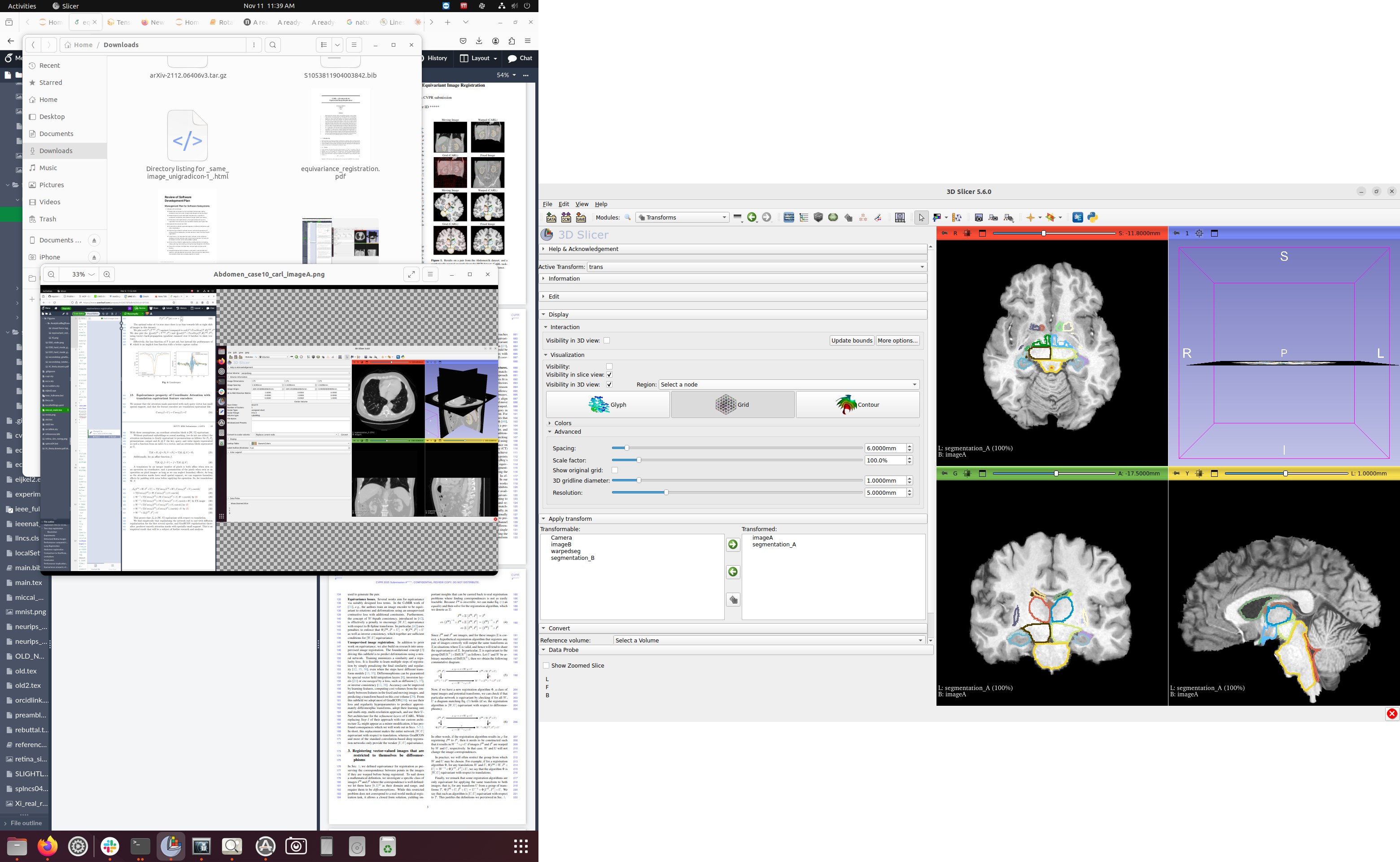}                  \\
			Grid (CARL\{ROT\})                                                                                                                              & Fixed Image   \\
			\includegraphics[width=.44\columnwidth, trim={75cm 14.4cm 20.4cm 41.3cm}, clip]{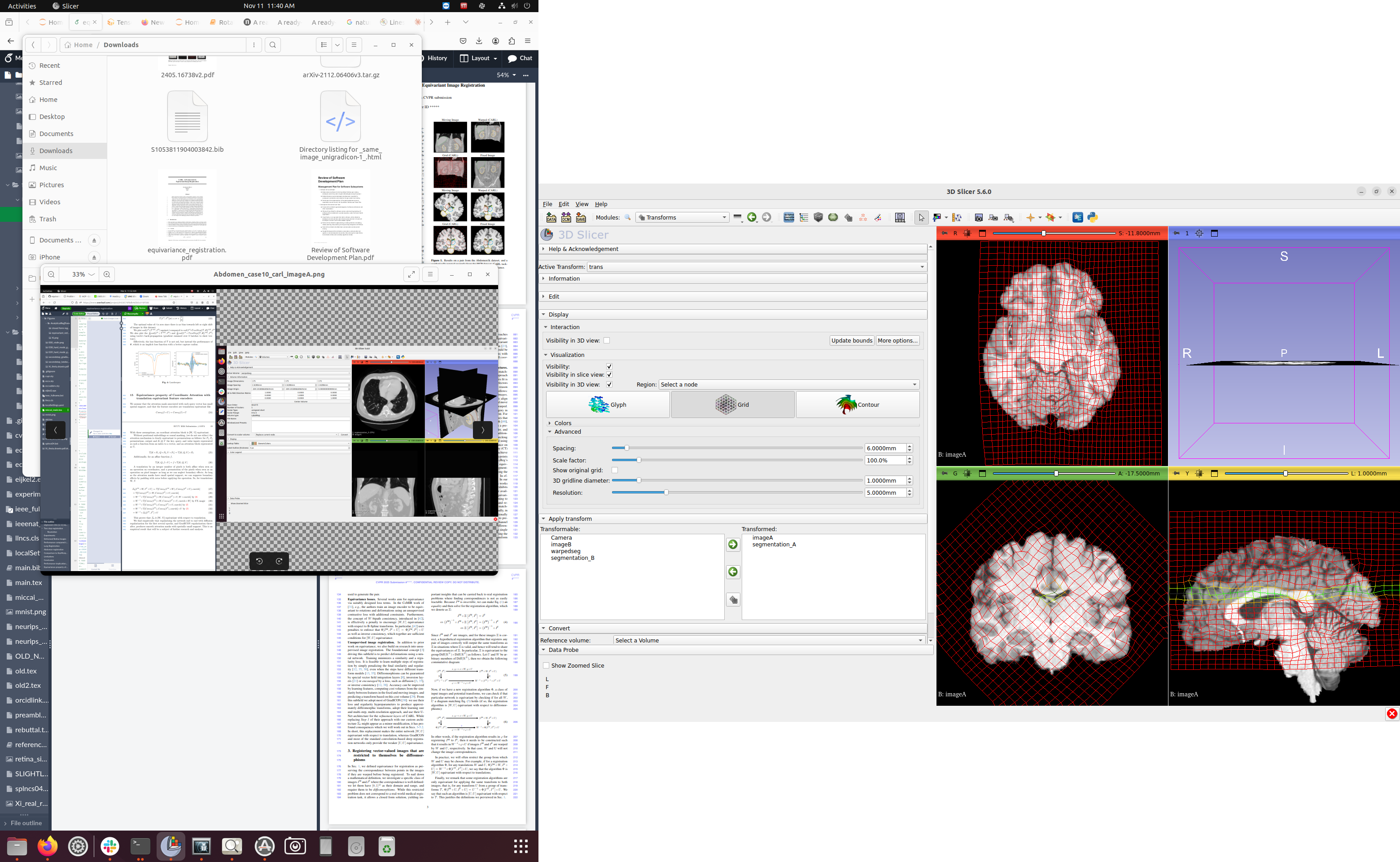} &
			\includegraphics[width=.44\columnwidth, trim={75cm 14.4cm 20.4cm 41.3cm}, clip] {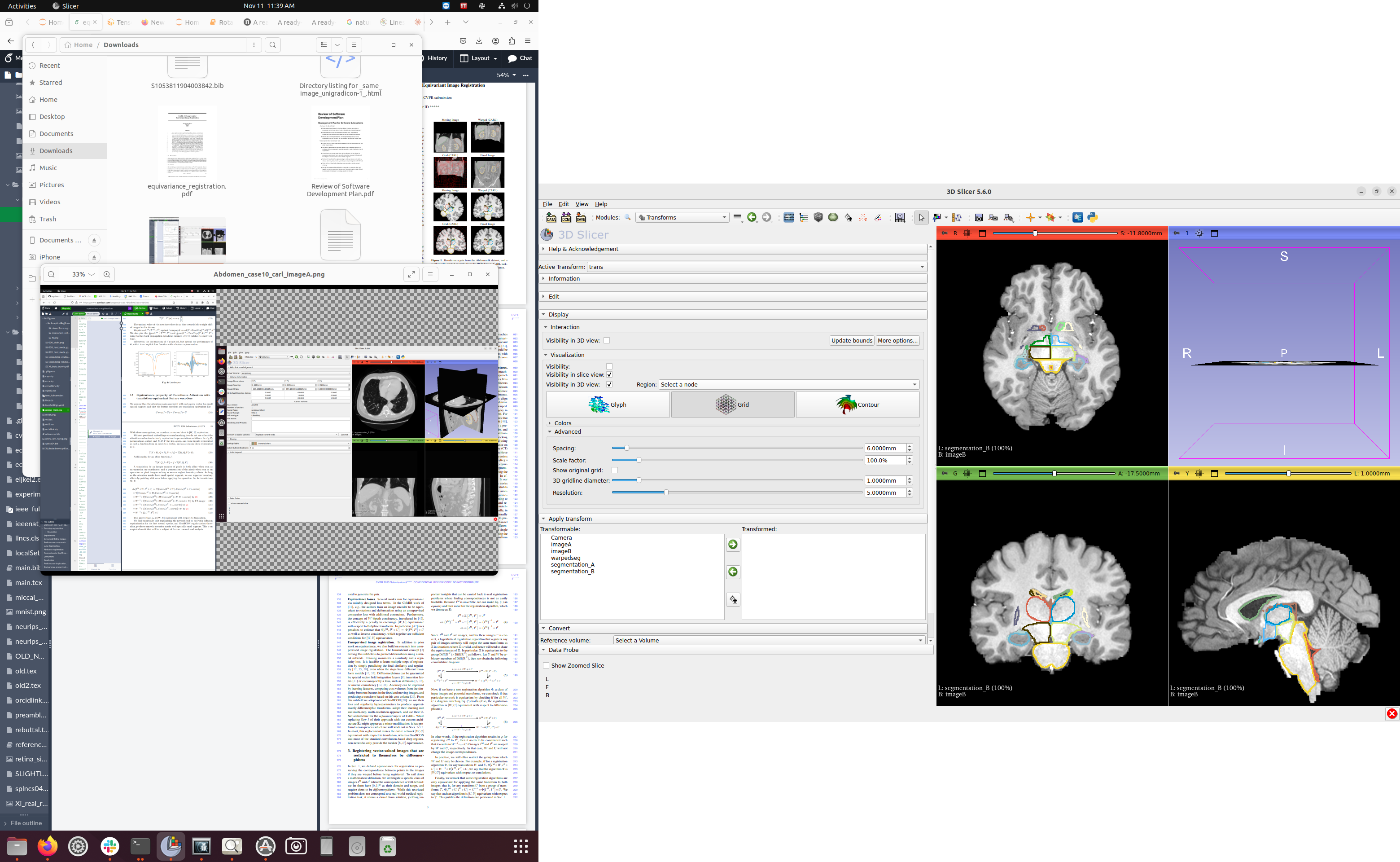} \\
%
		\end{tabular}
	\end{small}
	\caption{Results on a pair from the Abdomen1k dataset, and a synthetically rotated example from the HCP dataset. 
		CARL tackles large displacements and arbitrary rotations via equivariance.}
	\vskip-8pt
\end{figure}
\section{Introduction}

Image registration is an important task in medical image computing~\cite{roleofreg}. There is extensive research into solving this problem with deep learning~\cite{yang2017quicksilver, balakrishnan2019voxelmorph, mok2020large, tian2022}. In particular, unsupervised deep learning methods~\cite{greer2023inverse, Wang2023ARA} have attained strong performance on many benchmarks \cite{hering2022learn2reg}.
We present an equivariant registration layer that can be inserted into an  unsupervised multi-step registration network, which makes the overall registration architecture equivariant to independently translated images.

\noindent
\textbf{Notation.}
\label{section:notation}
We denote the \emph{moving} and \emph{fixed} images as $\ia$ and $\ib$,  respectively. They are functions $\Omega \rightarrow \real^D$, where $\Omega$ is the image domain, and $D$ is typically $1$ for a single-channel imaging modality. We denote a free \emph{transform} with a lowercase Greek letter such as $\varphi$.
Further, we denote a \emph{registration algorithm} as a capital Greek letter, such as $\Phi$, $\Xi$, or $\Psi$. We note that neural networks that perform registration are a subset of registration algorithms.
In our notation, a registration algorithm can be applied to a pair of images by square brackets, \ie, $\Phi\left[\ia, \ib\right]$. 
Hence, when applied to a pair of images, a registration algorithm yields a map $\mathbb{R}^N \rightarrow \mathbb{R}^N$ (ideally a diffeomorphism) that can be post-composed
with the moving image to yield a function that is close to the fixed image, \ie,
\begin{equation}
	\ia \circ \Phi\left[\ia, \ib\right] \sim \ib.
	\label{registration_problem_statement}
\end{equation}
With this notation, we can define what it means for a registration algorithm to be \emph{equivariant}. Conceptually, equivariance means that when the input images are warped, correspondence between the registered images is preserved; it is not required that the correspondence is \emph{correct}, only that it is \emph{the same} before and after applying a warp to the input images. We say a registration algorithm $\Phi$ is \textcolor{blue!75!black}{$[W, U]$ equivariant} with respect to a class of transforms $\mathcal{T}$ if correspondence is preserved when the images are warped \emph{independently}. Formally, this means
\begin{tcolorbox}[top=-2mm,bottom=0mm,boxsep=1mm,colback=blue!5!white,colframe=blue!75!black]
 \vskip -1em
 \begin{align}
		& \forall\ W, U \!\in \mathcal{T}: \label{eqn:WUequiv}                              \\
		& \Phi\left[\ia \circ W, \ib\! \circ U\right] = W^{-1}\! \circ \Phi\left[\ia, \ib\right] \circ U\,.\notag
\end{align}
\end{tcolorbox}
As a weaker form of equivariance, we say a  registration algorithm is \textcolor{green!25!black}{$[U, U]$ equivariant} if correspondence is preserved when the images are warped \emph{together}, \ie,
\begin{tcolorbox}[top=-2mm,bottom=0mm,boxsep=1mm,colback=green!5!white,colframe=green!25!black]
\vskip -1em
	\begin{align}
		& \forall\ U \!\in \mathcal{T}: \label{eqn:UUequiv}  \\
        & \Phi\left[\ia \circ U, \ib \circ U\right] = U^{-1} \circ \Phi\left[\ia, \ib\right] \circ U\,. \notag
	\end{align}
\end{tcolorbox}
\noindent
From our analysis (see Sec.~\ref{sec:diff_to_diff_registration}), it will be clear that both definitions are well justified.

%
%
%
%
%
%
%
\vspace{-0.1cm}
\section{Related work}
\label{section:related_work}

We review related work at the intersection of equivariance (and how to achieve it) and image registration, as well as previous work on unsupervised learning of neural networks for image registration.

\vskip0.5ex
\noindent
\textbf{Equivariant encoders.}
The most impactful equivariant layers in the field of neural networks have been ordinary convolution \cite{fukushimaneocognitronbc} and attention \cite{parikh-etal-2016-decomposable}, which are equivariant to translation and permutation,  respectively. In this work, we focus on using attention and convolution to make a registration algorithm that is by construction $[W, U]$ equivariant to translation. However, there are also more recent approaches to create components with additional geometric equivariances such as to rotation. These include group equivariant convolution
\cite{cohen2016group}, StyleGAN's equivariant convolution \cite{karras2021alias}, and spherical harmonic convolution \cite{moyer2021equivariant}. These
should be explored in future work to expand the deformations with respect to
which our approach (based on what we call \emph{coordinate attention}, see Sec.~\ref{sec:coordinate_attention_block})  is $[W, U]$ equivariant.

\vskip0.5ex
\noindent
\textbf{Keypoint-based equivariant registration architectures.}
Several registration approaches relying on keypoint matching have been proposed~\cite{Lowe2004DistinctiveIF, SURF}. Notably, any approach that aligns keypoints to keypoints using a least squares fit is $[W, U]$ equivariant to rigid motions if the keypoint selectors and descriptors are equivariant to rigid motions. The reason can be derived from the notion of equivariance as follows: \emph{if two points correspond before warping the input images, they should correspond afterwards}. The keypoints align with each other before and after a warp is applied, and move with features on the input images when they are warped. SURF \cite{SURF} and related methods fall under this category in the subfield of non-learning-based 2D image alignment. For medical image registration, learning-based approaches that leverage keypoints include EasyReg \cite{iglesias2023ready}, KeyMorph \cite{Yu22a}, SAME++ \cite{Tian2023SAMEAS} and E-CNN \cite{billot2023equivariant}.
SAME++ \cite{Tian2023SAMEAS} uses a pretrained foundation model as a convolutional encoder, and extracts matched keypoints from it using methods traditionally used in 2D image feature alignment, such as checking for cycle consistency. This alignment is then refined using a trained deep network with overall strong performance on lung, abdomen, and head/neck computed tomography (CT) registration tasks.
EasyReg, E-CNN and KeyMorph achieve equivariant registration by locating a fixed set of keypoints in each image and aligning matching keypoints.
EasyReg's \cite{iglesias2023ready} affine pre-registration is by construction $[W,U]$ equivariant to affine deformations because it works by segmenting the brain images to be registered, and then bringing the centroids of the segmented regions into alignment. In effect, this uses each segmented region as a keypoint. In our experiments (see Sec.~\ref{sec:experiments}) we will see that this strategy works well even for out-of-distribution translations, and exhibits strong performance when detailed segmentations are available for training. In the KeyMorph work of \cite{Wang2023ARA}, equivariance is achieved in an unsupervised setting by learning to predict keypoints in brain images. Unlike SURF and related approaches, KeyMorph avoids the need for a matching step by predicting keypoints in a fixed order. Finally, in the E-CNN work of \cite{billot2023equivariant}, the authors leverage a rotationally equivariant encoder built using spherical harmonics to predict features, and the centers of mass of each feature channel are then used as keypoints and rigidly aligned in a differentiable manner. The network is trained by warping a single image with two different transforms and then penalizing the difference between the network output and the transform used to generate the pair.


\vskip0.5ex
\noindent
\textbf{Equivariance losses.} Several works aim for equivariance via suitably designed loss terms. In the CoMIR work of \cite{Nordling2023ContrastiveLO}, \eg, the authors train an image encoder to be equivariant to rotations and deformations using an unsupervised contrastive loss with additional constraints.
Furthermore, the concept of $W$-bipath consistency, introduced in \cite{Truong_2021_ICCV}, is effectively a penalty to encourage $[W, U]$ equivariance with respect to B-Spline transforms. In particular, \cite{Truong_2021_ICCV} uses penalties to enforce that $\Phi[\ia, \ib \circ U] = \Phi[\ia, \ib] \circ U$ as well as inverse consistency, which together are sufficient conditions for $[W, U]$ equivariance. 

%

\vskip0.5ex
\noindent
\textbf{Unsupervised image registration.}
In addition to prior work on equivariance, we also build on research into unsupervised image registration. The foundational concept~\cite{balakrishnan2019voxelmorph} driving this subfield is to predict deformations using a neural network. Training minimizes a similarity and a regularity loss. It is feasible to learn multiple steps of registration by simply penalizing the final similarity and regularity~\cite{shen2019networks, greer2021icon, tian2022}, even when the steps have different transform models~\cite{greer2023inverse, shen2019networks}. Diffeomorphisms can be guaranteed by special vector field integration layers~\cite{dalca2018unsupervised}, inversion layers~\cite{asymreg} or \emph{encouraged} by a loss, such as diffusion \cite{balakrishnan2019voxelmorph, shen2019networks}, or inverse consistency \cite{greer2021icon, tian2022}. Accuracy can be improved by learning features, computing cost volumes from the similarity between features in the fixed and moving images, and predicting a transform based on this cost volume~
\cite{mok2020large}.
From this subfield we adopt most of GradICON~\cite{tian2022,tian2024unigradicon,demir2024multigradicon}: we use their loss and regularity hyperparameters to produce approximately diffeomorphic transforms, adopt their learning rate and multi-step, multi-resolution approach, and use their U-Net architecture for the \emph{refinement layers} of CARL. While replacing \emph{Step 1} of their approach with our custom architecture $\Xi_\theta$ might appear minor, it has profound consequences which we will work out in Secs.~\ref{sec:diff_to_diff_registration}-\ref{sec:two_step_registration}. In short, this replacement makes the entire network $[W, U]$ equivariant with respect to translation, whereas GradICON and most of the convolution-based deep registration networks only provide the weaker $[U, U]$ equivariance.

\section{Registering vector-valued images that are restricted to be diffeomorphisms}
\label{sec:diff_to_diff_registration}




In Sec.~\ref{section:notation}, we defined equivariance for registration as preserving the correspondence between points in the images if they are warped before being registered. To obtain a mathematical definition, we investigate a specific class of images $\ia$ and $\ib$ where the correspondence is well defined: we let them have $[0, 1]^D$ as their domain and range, and require them
to be \emph{diffeomorphisms}. While this restricted problem does not
correspond to a real-world medical registration task, it allows a closed
form solution, yielding insights that can be carried back to real registration problems where finding correspondences is not trivial. Because $\ia$ is \emph{invertible}, we can make
Eq.~\eqref{registration_problem_statement} an \emph{equality} and then solve for the
registration algorithm, which we denote as $\Xi$:
\begin{equation}
	\begin{split}
		\ia \circ \Xi\left[\ia, \ib\right]                                   & = \ib                       \\
		\Leftrightarrow \left(\ia\right) ^{-1} \circ \ia \circ \Xi\left[\ia, \ib\right] & = \left(\ia\right) ^{-1} \circ \ib     \\
		\Leftrightarrow  \Xi\left[\ia, \ib\right]                            & = \left(\ia\right) ^{-1} \circ \ib\,.
	\end{split}
\end{equation}
Since $\ia$ and $\ib$ are images, and for these images $\Xi$ is correct, a hypothetical registration algorithm that registers any pair of images correctly
will output the same transforms as $\Xi$ in situations where $\Xi$ is valid,
and hence will tend to share the equivariances of $\Xi$. In particular, $\Xi$ is equivariant to the group $\text{Diff}(\real^N) \times \text{Diff}(\real^N)$ as follows. Let $U$ and $W$ be arbitrary members of $\text{Diff}(\real^N)$ (diffeomorphisms), then
we obtain the following commutative diagram:
\begin{equation}
\label{eqn:equivarianceproperty}
	\begin{tikzpicture}[baseline=(current  bounding  box.center),scale=0.78, transform shape,node distance=.7 cm and 4.5cm, auto,
		every edge/.style={draw, -{Stealth[length=2mm]}, thick}]
		\node (tl) {$(I^\text{M}, I^\text{F})$};
		\node (tr) [right=of tl] {$(I^\text{M} \circ W, I^\text{F} \circ U)$};
		\node (bl) [below=of tl] {$(I^M)^{-1} \circ I^F$};
		\node (br) [below=of tr] {$W ^{-1} \circ (I^M)^{-1} \circ I^F \circ U$};

		\draw (tl) edge node[above] {$x, y \mapsto x \circ W, y \circ U$} (tr);
		\draw (tl) edge node[left] {$\Xi$} (bl);
		\draw (tr) edge node[right] {$\Xi$} (br);
		\draw (bl) edge node[below] {$\varphi \mapsto W^{-1} \circ \varphi \circ U$} (br);
	\end{tikzpicture}
\end{equation}
\noindent
Now, if we have a new registration algorithm $\Phi$, a class of input images and potential transforms, we can check if that particular network is equivariant by checking if for all $W$, $U$ a diagram matching \cref{eqn:equivarianceproperty} holds (if so, the registration algorithm is $[W,U]$ equivariant w.r.t. diffeomorphisms):
\begin{equation}
	\begin{tikzpicture}[baseline=(current  bounding  box.center),scale=0.78, transform shape,node distance=.7 cm and 4.5cm, auto,
		every edge/.style={draw, -{Stealth[length=2mm]}, thick}]

		\node (tl) {$(I^\text{M}, I^\text{F})$};
		\node (tr) [right=of tl] {$(I^\text{M} \circ W, I^\text{F} \circ U)$};
		\node (bl) [below=of tl] {$\Phi[I^M , I^F]$};
		\node (br) [below=of tr] {$W ^{-1} \circ \Phi[I^M , I^F] \circ U$};

		\draw (tl) edge node[above] {$x, y \mapsto x \circ W, y \circ U$} (tr);
		\draw (tl) edge node[left] {$\Phi$} (bl);
		\draw (tr) edge node[right] {$\Phi$} (br);
		\draw (bl) edge node[below] {$\varphi \mapsto W^{-1} \circ \varphi \circ U$} (br);        \draw (bl) edge node[above] {{\color{blue}?}} (br);
	\end{tikzpicture}
\end{equation}
In other words, if the registration algorithm results in $\varphi$ for registering $I^{\text{M}}$ to $I^\text{F}$, then it needs to be constructed such that it results in $W^{-1}\circ\varphi\circ U$ if images $I^{\text{M}}$ and $I^{\text{F}}$ are warped by $W$ and $U$, respectively. In that case, $W$ and U will not change the image correspondences.

\vskip1ex
In practice, we will often restrict the group from which $W$ and $U$ may be chosen. For example, if for a registration algorithm
$\Phi$, for any translations $W$ and $U$, $\Phi[\ia \circ W, \ib \circ U] =
	W^{-1} \circ \Phi[I^M, I^F] \circ U$, we say that the algorithm $\Phi$ is $[W, U]$
equivariant with respect to translations.

\vskip1ex
Finally, we remark that some registration algorithms
are only equivariant for applying the same transform to both images; that is,
for any transform $U$ from a group of transforms $\mathcal{T}$, $\Phi[\ia \circ
		U, \ib \circ U] = U^{-1} \circ \Phi[I^M, I^F] \circ U$. We say that such an
algorithm is $[U, U]$ equivariant with respect to $\mathcal{T}$. This justifies the definitions we previewed in Sec.~\ref{section:notation}.

\section{Coordinate attention}
\label{sec:coordinate_attention_block}

Our central contribution when seeking to realize $[W,U]$ equivariance is what we call \emph{coordinate attention}, \ie, a standard attention block where the value vectors are the coordinates of each voxel. This has the effect of computing the center of mass of the attention mask associated with each query. 
As an illustrative example in Appendix \ref{subsection:xinetimplementation}, we use coordinate attention to solve the diffeomorphism-to-diffeomorphism registration problem.

\begin{figure}
	\centering
	\includegraphics[width=.99\columnwidth]{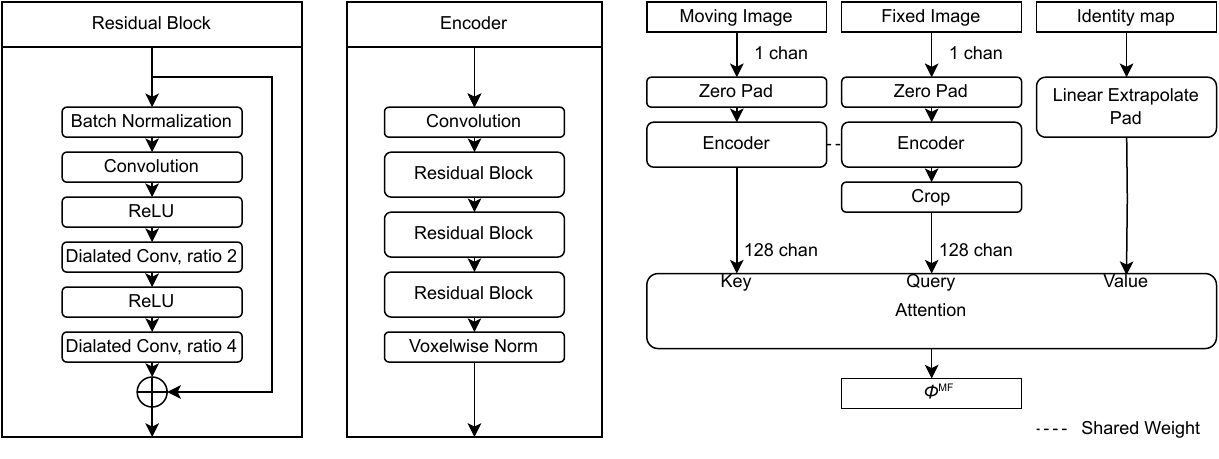}
	\caption{Architecture of $\Xi_\theta$. The specific arrangement of pads and crops allows voxels to be mapped to points outside of $\Omega$ -- this is necessary to represent translation. 
	}
	\label{fig:xitheta}
\end{figure}

\vskip0.5ex
For registering medical images, we use an architecture we call $\Xi_\theta$, a combination of coordinate attention with a feature encoder consisting of convolutional layers all operating at the same resolution, some of which are dilated. The encoder is exactly equivariant to integer pixel translations.
Edge effects are reduced by padding with zeros before running
images through the feature encoder, and cropping afterwards. The architecture is illustrated in Fig.~\ref{fig:xitheta}. 
It is important that the coordinate attention formulation, which is ``center of mass of $\text{softmax}(QV^\top)$'', can be computed using a standard attention operation instead of a custom set of operations on all pairs of feature vectors. Attention has heavily optimized implementations such as
flash attention~\cite{dao2022flashattention} that only require $\mathcal{O}$(\#voxels) memory and have excellent
cache coherency. While flash attention is $\mathcal{O}$(image side length$^6$),
in practice performance is adequate
in our setting when $\Xi_\theta$ is applied to a volume of size $43 \times 43 \times 43$,
which is sufficient for coarse registration. We will show (in Sec.~\ref{twostepproof}) that it is then possible to
add fine-grained registration while preserving $[W, U]$ equivariance. 
\section{Equivariance of Coordinate Attention with convolutional encoders (\texorpdfstring{$\Xi_\theta$)}{XiTheta}}
\label{sec:wuproof}
We assume that the attention mask associated with each query vector has small spatial
support. Finding a training procedure that reliably fulfilled this assumption across different datasets was nontrivial: we find that this assumption is satisfied after regularizing the  network end-to-end with diffusion
regularization for the first several epochs, and using GradICON regularization thereafter. 

We assume that the feature encoders are translation equivariant like
\begin{equation}
	\convo_\theta(I \circ U) = \convo_\theta(I) \circ U
	\label{convdef}\,.
\end{equation}
With these assumptions, we prove that $\Xi_\theta$ is $[W, U]$ equivariant to translations below.

Without positional embeddings or causal masking, (we do not use
either) the attention mechanism is equivariant to permutations as follows: for $P_1, P_2$ permutations; and the output and $K$ (Key), $Q$ (Query), and $V$ (Value) inputs represented each as a function from an index to a vector, and an attention block represented as $\mathbb{T}$,
\begin{equation} \mathbb{T}[K \circ P_1, Q \circ P_2, V \circ P_1] = \mathbb{T}[K, Q, V] \circ P_2\,. \label{transperm}\end{equation}
Additionally, because the attention weights in an attention block sum to 1, for an affine function $f$, we have 
\begin{equation}\mathbb{T}[K, Q, f\circ V] = f \circ \mathbb{T}[K, Q, V]\,. \label{transaff}\end{equation}
A translation $W$ by an integer number of voxels is both affine when seen as an operation on coordinates, $W_{x \mapsto x + r}$, and a permutation of the
voxels when seen as an operation on voxel images $W_{\text{permutation}}$ (as long as we can neglect boundary effects). The map from indices to coordinates, $\text{coords}$, serves as the bridge between these two representations of a transform ($W_{x \mapsto x + r} \circ \text{coords} = \text{coords} \circ W_{\text{permutation}})$. As long as the attention masks have small spatial support, we can suppress boundary effects by padding with zeros before applying the operation. So, for translations $W$ and $U$, we have
\[\Xi_\theta[ I^M, I^F] :=\mathbb{T}[\convo_\theta (I^M ) , \convo_\theta(I^F), \text{coords}]\enspace, \]
from which we establish that $\Xi_\theta$ is $[W, U]$ equivariant with respect to translation as follows: 
\begin{equation}
	\begin{split}
		& \Xi_\theta[ I^M \circ W, I^F \circ U]  \\
        & =\mathbb{T}[\convo_\theta (I^M \circ W) , \convo_\theta(I^F \circ U), \text{coords}]          \\
		& \stackrel{\eqref{convdef}}{=} \mathbb{T}[\convo_\theta (I^M) \circ W , \convo_\theta(I^F) \circ U, \text{coords}] \\
		& \stackrel{\eqref{transaff}}{=} W^{-1} \circ \mathbb{T}[\convo_\theta (I^M) \circ W, \convo_\theta(I^F) \circ U, W \circ \text{coords}] \\
		& = W^{-1} \circ \mathbb{T}[\convo_\theta (I^M) \circ W, \convo_\theta(I^F) \circ U, \text{coords} \circ W]  \\
		& \stackrel{\eqref{transperm}}{=} W^{-1} \circ \mathbb{T}[\convo_\theta (I^M), \convo_\theta(I^F) \circ U, \text{coords}] \\
		& \stackrel{\eqref{transperm}}{=}  W^{-1} \circ \mathbb{T}[\convo_\theta (I^M), \convo_\theta(I^F), \text{coords}] \circ U \\
		& =  W^{-1} \circ \Xi_\theta[ I^M , I^F] \circ U\,.
	\end{split}
\end{equation}
The same argument can also be applied to $[W, U]$ equivariance to axis aligned $\pi$ or $\frac{\pi}{2}$ rotations, provided that $\convo_\theta$ is replaced with an appropriate rotation equivariant encoder.

\subsection{\texorpdfstring{$[U, U]$}{[U,U]} equivariance}
\label{sec:uu_equivariance}
So far, we have pointed out that $\Xi_\theta$ is $[W, U]$ equivariant to translation. Our next goal is to show that our full approach, \emph{Coordinate Attention with Refinement Layers} (CARL), is also $[W, U]$ equivariant to translation. To this end, we (1) show that VoxelMorph-like networks are $[U, U]$ equivariant to translation as they predict displacements, and then (2) show that a multi-step registration algorithm where the first step is $[W, U]$ equivariant and subsequent steps are $[U, U]$ equivariant is overall $[W, U]$ equivariant. Notably, any all-convolutional network is translationally equivariant
\footnote{At
	least, for translations of integer multiples of the least common multiple of all convolution
	strides, and considering the input and output of the convolution to be
	functions.}:
\begin{equation}
	\begin{tikzpicture}[baseline=(current  bounding  box.center),scale=0.9, transform shape,node distance=.7cm and 4.5cm, auto,
		every edge/.style={draw, -{Stealth[length=2mm]}, thick}]
		\node (tl) {$X$};
		\node (tr) [right=of tl] {$\tilde{X}$};
		\node (bl) [below=of tl] {$\phantom{\hat{Y}}$$Y$\hspace{-0.2cm} $\phantom{\hat{Y}}$};
		\node (br) [below=of tr] {$\tilde{Y}$};
		\draw (tl) edge node[above] {$f \mapsto f \circ U$} (tr);
		\draw (tl) edge node[left] {$\text{Conv}_\theta$} (bl);
		\draw (tr) edge node[right] {$\text{Conv}_\theta$} (br);
		\draw (bl) edge node[below] {$f \mapsto f \circ U$} (br);
	\end{tikzpicture}
	\label{eqn:convequi}
\end{equation}
To apply a convolutional network to registration, first it is made to have two inputs by channel-wise concatenation (cat), and it is chosen to have $D$ output channels. Its output can then be interpreted (via interpolation) as a function from the domain of the input images to $\mathbb{R}^N$. If used to predict coordinates, this has a different equivariance than $[U, U]$ equivariance: to be $[U, U]$ equivariant with respect to translation, a registration algorithm needs to have a bottom arrow $f \mapsto U^{-1} \circ f \circ U$ where Eq.~\eqref{eqn:convequi} has $f \mapsto f \circ U$. This is ameliorated in Quicksilver~\cite{yang2017quicksilver}, VoxelMorph~\cite{balakrishnan2019voxelmorph} and their successors by predicting displacements instead of coordinates. Overall, in VoxelMorph we have
\begin{equation}
	\Phi_\theta[\ia, \ib](\vec{x}) := \text{Conv}_\theta[\text{cat}(\ia, \ib)](\vec{x}) + \vec{x}\enspace,
\end{equation}
and by parameterizing our translation transform $U$ by a vector $\vec{r}$, we can compute
\begin{equation}
	\begin{split}
		\Phi_\theta&[\ia  \circ U, \ib \circ U](\vec{x}) \\ &= \text{Conv}_\theta[\text{cat}(\ia \circ U, \ib \circ U)](\vec{x}) + \vec{x}                       \\
		&= \text{Conv}_\theta[\text{cat}(\ia, \ib) \circ U](\vec{x}) + \vec{x}      \\
		&= \left(\text{Conv}_\theta[\text{cat}(\ia, \ib)] \circ U\right)(\vec{x}) + \vec{x}                  \\
		&= \text{Conv}_\theta[\text{cat}(\ia, \ib)](U(\vec{x})) + \vec{x}                         \\
		& = \text{Conv}_\theta[\text{cat}(\ia, \ib)](U(\vec{x})) + (\vec{x} + \vec{r}) - \vec{r}              \\
		&= U^{-1} \circ \left(\text{Conv}_\theta[\text{cat}(\ia, \ib)](\vec{x}) + \vec{x} \right) \circ U  \\ &= U^{-1} \circ \Phi_\theta[\ia, \ib](\vec{x}) \circ U\enspace.
	\end{split}
\end{equation}
\subsection{Two-step registration}
\label{sec:two_step_registration}
\label{twostepproof}
Multi-step registration is common practice, often using multiple algorithms. For example, ANTs \cite{avants2008symmetric} by default performs affine, then deformable registration, and returns the composition of the resulting transforms. Greer \etal~\cite{greer2021icon} propose a general notation for this scheme: the TwoStep operator, which takes as input two registration algorithms, and yields a composite algorithm as follows:
\begin{align}\text{TwoStep}&\{\Phi, \Psi\}[I^M, I^F] \nonumber \\ &= \Phi[I^M, I^F] \circ \Psi[I^M \circ \Phi[I^M, I^F], I^F]\,.\end{align}
We show that if $\Phi$ is $[W, U]$ equivariant and $\Psi$ is $[U, U]$ equivariant for $U,W$
from a class of transforms $\mathcal{T}$, then $\text{TwoStep}\{\Phi, \Psi\}$ is
$[W, U]$ equivariant. In particular, application of the definition of $\Phi$ being $[W, U]$ equivariant, see Eq.~\eqref{eqn:WUequiv}, and $\Psi$ being $[U, U]$ equivariant, see Eq.~\eqref{eqn:UUequiv}, yields
\begin{equation}
	\begin{split}
		& \text{TwoStep}\{\Phi, \Psi\}[I^M \circ W, I^F \circ U]  \\                 
		& \vspace{0.5cm}= \Phi\left[I^M \circ W, I^F \circ U\right] \circ \\
        & \vspace{0.5cm}\phantom{=~} \Psi\left[I^M \circ W \circ \Phi[I^M \circ W, I^F \circ U], I^F \circ U\right] \\
        & \vspace{0.5cm}\stackrel{\eqref{eqn:WUequiv}}{=} W^{-1} \circ \Phi[I^M, I^F] \circ U \circ \\
        & \vspace{0.5cm}\phantom{=~} \Psi\left[I^M \circ \Phi[I^M, I^F] \circ U, I^F \circ U\right] \\
        & \vspace{0.5cm}\stackrel{\eqref{eqn:UUequiv}}{=} W^{-1} \circ \Phi[I^M, I^F] \circ \\ 
        & \vspace{0.5cm}\phantom{=~} \Psi\left[I^M \circ \Phi[I^M, I^F], I^F\right] \circ U \\
        &  \vspace{0.5cm}=W^{-1} \circ \text{TwoStep}\{\Phi, \Psi\}[I^M, I^F] \circ U\enspace.
    \end{split}
\end{equation}
This construction can be applied recursively for an $N$-step $[W, U]$ equivariant network, and can be combined with the downsampling operator (abbreviated as Down) from \cite{tian2022} (see Appendix \ref{downsample}) to make the early stages operate at coarse resolution and subsequent ones at higher resolutions.


\section{Experiments}
\label{sec:experiments}
\noindent
\textbf{Architecture.} We propose the architecture \emph{Coordinate Attention with Refinement Layers} (CARL),
\begin{tcolorbox}[top=0mm,bottom=0mm,boxsep=1mm,colback=red!5!white,colframe=red!75!black]
    \vskip -1em
	\begin{align}
		\text{CARL}   & :=\text{TwoStep}\{\text{TwoStep}\{ \text{Down}\{ \text{TwoStep} \{ \nonumber \\
		              & \text{Down} \{  \Xi_\theta\}\}, \Psi_1\}\}, \Psi_2 \nonumber                 \\
		\}, \Psi_3 \} &
		\label{eqn:CARL}
	\end{align}
\end{tcolorbox}
\noindent
where $\Xi_\theta$ is a $[W, U]$ equivariant neural network composed of a
translation-equivariant feature encoder and a coordinate attention layer as proposed, and
$\Psi_i$ are refinement layers: convolutional networks predicting displacements with architecture and feature counts from \cite{greer2021icon, tian2022}, and so are $[U, U]$ equivariant. To facilitate a direct comparison of performance, the specific arrangement of TwoStep and Downsample layers is identical to GradICON: our only change is to replace the lowest resolution displacement predicting network with $\Xi_\theta$. Following the suggestion of GradICON \cite{tian2022}, the final refinement layer $\Psi_3$ is only added after the first 50,000 steps of training. \emph{The overall network is
	$[W, U]$ equivariant with respect to translation.}

\vskip1ex
\noindent
\textbf{Losses.} In all experiments, we train with LNCC image similarity (LNCC=1-local normalized cross correlation). For regularization, we train initially with a diffusion loss for 1,500 steps, \ie,
\begin{equation}
\begin{split}
	\mathcal{L} & = \text{LNCC}\left(I^\text{M} \circ \text{CARL}[\ia, \ib], \ib\right)  \\ &+  \text{LNCC}\left(I^\text{F} \circ \text{CARL}[\ib, \ia], I^\text{M}\right) \\&+ \lambda ||\nabla (\text{CARL}[\ia, \ib]) - \mathbf{I} ||_F^2
    \end{split}
\end{equation}
and then continue training with the GradICON \cite{tian2022} regularizer for 100,000 steps, \ie,
\begin{equation}
\begin{split}
\mathcal{L} & =  \text{LNCC}(I^M \circ \text{CARL}[\ia, \ib], \ib)  \\&+  \text{LNCC}(I^F \circ \text{CARL}[\ib, \ia], I^M)  \\&+ \lambda ||\nabla (\text{CARL}[\ia, \ib] \circ \text{CARL}[\ib, \ia]) - \mathbf{I} ||_F^2\,,
              \label{gradicon_loss}
\end{split}
\end{equation}
where $\mathbf{I}$ is the identity matrix, and $\lambda>0$.
We use regularization parameter $\lambda = 1.5$ and an Adam optimizer with a learning rate of 0.0001. We remark that diffusion regularization is needed in early steps of training for stability, but \text{GradICON} delivers superior final regularity and correspondence by strongly forcing the network towards invertibility while still allowing large deformations.
\vskip1ex
\noindent
\textbf{Instance Optimization (IO).} Following GradICON~\cite{tian2022}, we evaluate with and without 50 steps of Instance Optimization, \ie, since our method does not use human annotations at train time, at deployment we can further optimize Eq.~\eqref{gradicon_loss} on instances of test data to improve performance.

\subsection{Deformed retina images}
\label{retinaexpr}
\begin{figure*}[t!]
\centering
\includegraphics[width=1.0\textwidth]{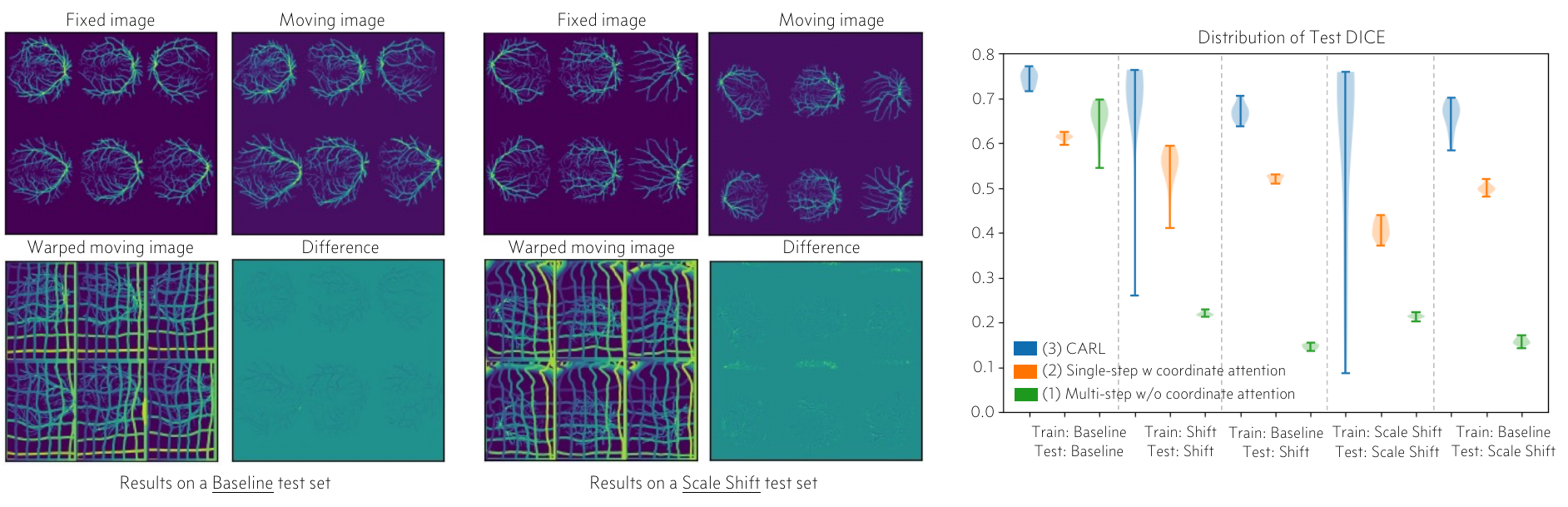}
\caption{\label{fig:synthetic_experiment}Equivariance allows CARL to generalize out-of-distribution. The \emph{left four images} show the performance of CARL on a test set with the same distribution as the training set, where images are \emph{aligned} in scale and translation. The \emph{middle four images} show generalization to a test set \textbf{Scale Shift} where images are \emph{misaligned} in scale and translation. The \emph{right-most} figure shows the mean Dice distribution on the test set, over multiple training runs.}
\vskip-0.5em
\end{figure*}

We first demonstrate the importance of several details of
our implementation on a dataset consisting of synthetically warped retina
segmentations from \cite{tian2022}. In particular, for this experiment,
we train three network variants:
\begin{itemize}
	\item[1)] $\text{TwoStep}\{\text{Down}\{\text{TwoStep}\{\text{Down}\{\Psi_0\},\Psi_1\}\}, \Psi_2 \}$, a multi-step network with each step predicting a displacement field. This is the 1st stage architecture proposed in GradICON~\cite{tian2022}. We expect this network to \emph{not} be $[W, U]$ equivariant w.r.t. translation.
	\item[2)] $\text{Down}\{\text{Down}\{\Xi_\theta\}\}$, a single-step coordinate attention network . We expect this network to be $[W, U]$ equivariant w.r.t translation, but less powerful than 1).
	\item[3)] CARL, Eq.~\eqref{eqn:CARL}. We expect this  network to be powerful and $[W, U]$ equivariant to translation.
\end{itemize}
Each is assessed \emph{without} instance optimization. To assess differences between these network types, we create three variants of the retina dataset:
\begin{itemize}
	\item[1)] {\bf Baseline}. The train
		set is constructed by warping 14 segmentations from the DRIVE dataset \cite{staal2004ridge}, and the
		test set is constructed by warping the remaining 6 segmentations, with elastic warp parameters as suggested in~\cite{tian2022}.
	\item[2)] {\bf Shift}. The same as Baseline, but with fixed images shifted by 200 pixels ($\sim \frac{1}{3}$ of retina diameter).
	\item[3)] {\bf Scale Shift}. The same as Shift, but with fixed images also scaled to 80\% size (See Fig. \ref{fig:synthetic_experiment}).
\end{itemize}

\vskip0.5ex
\noindent
Results for this experiment are shown in Fig.~\ref{fig:synthetic_experiment}. We see that when training/testing on the \textbf{Baseline} dataset, as expected, all three networks have adequate performance, although the importance of the refinement layers is highlighted when $\text{Down}\{\text{Down}\{\Xi_\theta\}\}$ falls meaningfully behind. When trained and tested on the \textbf{Shift} variant, the equivariant networks succeed, while the displacement predicting network fails. Additionally, we see that the equivariant networks \emph{generalize} to the \textbf{Shift} test dataset when trained only on \textbf{Baseline}. Finally, we see that the equivariant networks also learn on, \emph{and generalize to}, the \textbf{Scale Shift} dataset. This result illustrates an important point that we did not prove but observe empirically: while $\Xi_\theta$ is formally only $[W, U]$ equivariant to integer translations, in practice it is $[W, U]$ equivariant to deformations that are locally approximately translations, such as scaling. We do see a large variance \textbf{between training runs} when training CARL on the \textbf{Shift} and \textbf{Scale Shift} datasets. That is, specifically when training on Shift or Scale Shift, most trained CARL networks have high performance across all test examples, but for a small fraction of training runs, performance is poor across the whole test set.

\subsection{Performance comparison to other methods}
We compare our method to existing approaches from the literature on three datasets. First, we
evaluate on the 10 DirLab COPDGene test cases as this is a popular and
difficult lung registration benchmark that allows us to compare our performance to
diverse and highly optimized methods. Second, we evaluate on the HCP brain
dataset, which allows us to directly compare to two existing equivariant
registration methods, EasyReg and KeyMorph. Third, we evaluate on a novel
registration task: registering images from the Abdomen1k dataset without
registration-specific preprocessing. This dataset has large translational
differences and different fields of view, which causes other unsupervised
registration approaches to converge poorly. Our strong equivariance makes this
task possible with the LNCC + regularization loss. We aim to present a picture of how our approach will work when applied as specified to new datasets. To this end, \emph{we use the same hyperparameters for all three datasets}.

\vskip0.5ex
\noindent
\textbf{Lung registration.}
To evaluate CARL against the current state of the art for general registration, we use the DirLab lung registration task, which is well standardized and has gained considerable attention in the community. In particular, we train on 999 inhale/exhale CT pairs of images from the COPDGene dataset, and evaluate based on landmark distance in the 10 standard DirLab pairs. We follow the data preparation and augmentation suggested in GradICON~\cite{tian2022}. Results are listed in Table~\ref{tab:exp_results}, showing the commonly-used mean target registration error (mTRE) score as well as the percentage of negative values for the Jacobian determinant ($\%|J|_{<0}$), indicative of the number of folds. Performance matches state of the art, exceeded significantly only by PTVReg and RRN which have hyper parameters selected specifically for the 10 challenge cases.

\begin{figure}[htp]
	\centering
	\includegraphics[width=\columnwidth, trim={0cm 0cm 0cm 0.43cm}, clip]{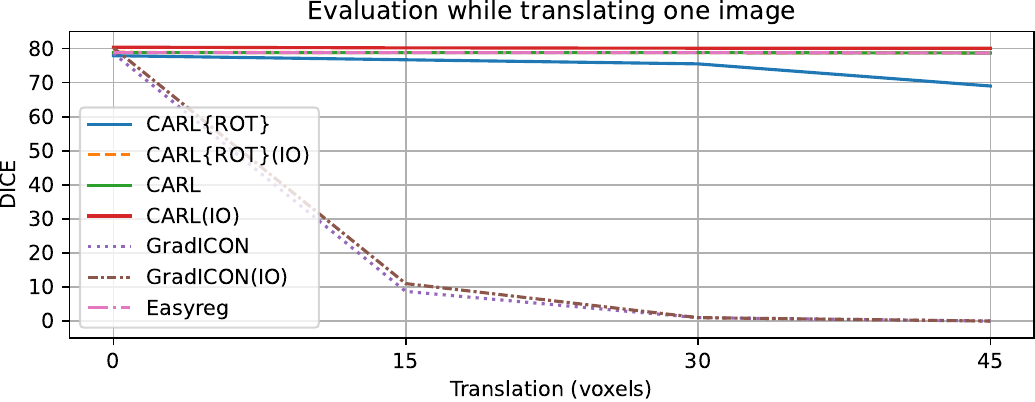}
	\includegraphics[width=\columnwidth, trim={0cm 0cm 0cm 0.43cm}, clip]{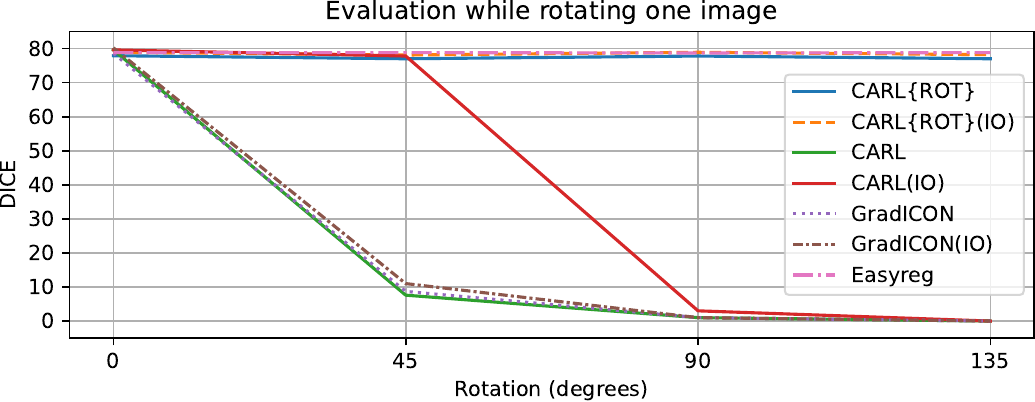}
	\caption{HCP evaluation while translating or rotating one image. CARL and EasyReg are unaffected by translation due to $[W,U]$ equivariance, and CARL\{ROT\} is additionally unaffected by rotation. GradICON DICE drops significantly when images are transformed differently due to its $[U,U]$ equivariance.}
	\label{fig:hcp_translation}
	\vskip-0.8em
\end{figure}

\vskip1ex
\noindent
\textbf{Brain registration.}
\label{brainreg}
The existing equivariant registration methods that we compare to are presented as brain registration methods. We compare to them on T1w MRI images from the Human Connectome Project (HCP) dataset~\cite{van2012human,rushmore2022anatomically,rushmore_r_jarrett_2022_dataset}. In detail, we perform evaluation and preprocessing as in \cite{greer2023inverse}, measuring DICE of midbrain structures \cref{fig:box_hcp} and find that CARL yields state-of-the-art performance. We also artificially translate or rotate one image by varying amounts and re-evaluate performance, verifying that GradICON sharply drops in performance under this shift, while our approach and EasyReg are unaffected (see Fig.~\ref{fig:hcp_translation}).

\vskip1ex
\noindent
\textbf{Abdomen registration.}
Abdominal CT-CT datasets can capture the diversity of fields of view of abdominal CT
scans in clinical practice, and the Abdomen1k~\cite{Abdomen1K9497733} dataset is curated specifically to do so. Since our method is robust to these differences of
field of view because it is $[W, U]$ equivariant to translation, we showcase its
performance for inter-subject registration of the Abdomen1K \cite{Abdomen1K9497733} dataset. We train on the first 800 cases of the Abdomen1K dataset, validate on cases 900-1000, and test on cases 800-900. We opt for a minimalist preprocessing: we resample all images to [175, 175, 175]
\footnote{This leads to nonisometric voxel spacing. We tried isometric voxel spacing, padding with background pixels if necessary, but non-isometric, non padded preprocessing had uniformly better performance}
and then divide by the maximum intensity. We test by randomly sampling 30 image pairs and then computing the mean Dice score over manual annotations of the liver, kidney, pancreas, and spleen. We report (Appendix, \ref{abdomen_fulldata}) the exact pairs to allow future works to compare to our result. The most important comparison is to {GradICON}, as our final network, similarity and regularity loss are identical to GradICON's except that we  replace the first U-Net based $\Psi$ with our custom $\Xi_\theta$. This change produces a dramatic improvement in non-instance-optimization registration accuracy and an improvement in instance-optimized registration accuracy. Results are listed in Table \ref{tab:exp_results}. We also check whether the model trained on the Abdomen1k dataset generalizes to the Learn2Reg Abdomen CT-CT challenge set \cite{hering2022learn2reg}. We find excellent generalization: performance is better than other unsupervised approaches including the keypoint-based SAME++, but does not match the best supervised methods trained using the Learn2Reg CT-CT training set \emph{and segmentations}.

\begin{table*}[htp]
	\centering
	\captionof{table}{Performance comparison on Abdomen1K, HCP, DirLab, and Learn2Reg Abdomen CT-CT dataset.}\label{tab:exp_results}
	\vskip0.1ex
	\begin{adjustbox}{width=.98\textwidth,center}
		\begin{tabular}{lccp{1em}lccp{1em}lcc}\toprule
			Method                                                    & $\%$DICE$\uparrow$ & $\%|J|_{<0}\downarrow$ & & Method                                                  & mTRE$\downarrow$ & $\%|J|_{<0}\downarrow$                        & & Method                                                        & $\%$DICE$\uparrow$ & Unsupervised \\ \midrule
			\multicolumn{3}{c}{\cellcolor[RGB]{205,224,238}Abdomen1K}                                           & & \multicolumn{3}{c}{\cellcolor[RGB]{255,228,206}DirLab}                                                                     & & \multicolumn{3}{c}{\cellcolor{blue!10}L2R Abd. CTCT Test set}                                              \\
			ANTs~\cite{avants2008symmetric}                           & 45.4           & 0                      & & ANTs~\cite{avants2008symmetric}                         & 1.79             & 0                                             & & ConvexAdam~\cite{siebert2021fast}                             & 69             &                       \\
			VoxelMorph~\cite{balakrishnan2019voxelmorph}              & 59.3           & -                      & & Elastix~\cite{klein2009elastix}                         & 1.32             & -                                             & & LapIRN~\cite{mok2020large}                                    & 67             &                       \\
			GradICON~\cite{tian2022}                                  & 49.6           & 7e-1                   & & VoxelMorph~\cite{balakrishnan2019voxelmorph}            & 9.88             & 0                                             & & Estienne~\cite{estienne2021deep}                              & 69             &                       \\
			GradICON(IO)~\cite{tian2022}                              & 71.0           & 4e-2                   & & LapIRN~\cite{mok2020large}                              & 2.92             & 0                                             & & corrField~\cite{hansen2021graphregnet,heinrich2015estimating} & 49             & yes                             \\
			ConstrICON~\cite{greer2023inverse}                        & 65.2           & 7e-4                   & & RRN~\cite{he2021recursive}                              & 0.83             & -                                             & & PIMed~\cite{hering2022learn2reg}                              & 49             &                              \\
			ConstrICON(IO)~\cite{greer2023inverse}                    & 66.8           & 2e-3                   & & Hering el al~\cite{hering2021cnn}                       & 2.00             & 6e-2                                          & & PDD-Net~\cite{heinrich2019closing}                            & 49             &                              \\
			uniGradICON~\cite{tian2024unigradicon}                    & 43.5           & 9e-1                   & & GraphRegNet~\cite{hansen2021graphregnet}                & 1.34             & -                                             & & Joutard~\cite{hering2022learn2reg}                            & 40             &                              \\
			uniGradICON(IO)~\cite{tian2024unigradicon}                & 54.8              &    2e-3                   & & PLOSL~\cite{wang2022PLOSL}                              & 3.84             & 0                                             & & NiftyReg~\cite{modat2010fast}                                 & 45             & yes                             \\
			\textbf{CARL}                                             & 75.7           & 4e-1                   & & PLOSL(IO)~\cite{wang2022PLOSL}                          & 1.53             & 0                                             & & Gunnarsson~\cite{gunnarsson2020learning}                      & 43             &                              \\
			\textbf{CARL(IO)}                                         & 77.2           & 2e-3                   & & PTVReg~\cite{vishnevskiy2017isotropic}                  & 0.83             & -                                             & & \multicolumn{3}{c}{\cellcolor{blue!10}L2R Abd. CTCT Val. set}                                              \\
			\multicolumn{3}{c}{\cellcolor[RGB]{208,234,208}HCP}                                                 & & GradICON~\cite{tian2022}                                & 1.93             & 3e-4                                          & & ConvexAdam~\cite{siebert2021fast}                             & 71                                              \\
			ANTs~\cite{avants2008symmetric}                           & 77.2           & 0                      & & GradICON(IO)~\cite{tian2022}                            & 1.31             & 2e-4                                          & & Estienne~\cite{estienne2021deep}                              & 65                                              \\
			GradICON~\cite{greer2023inverse}                          & 78.6           & 1e-3                   & & ConstrICON~\cite{greer2023inverse}                      & 2.03             & 7e-6                                          & & Estienne~\cite{estienne2021deep}                              & 52             & yes                            \\
			GradICON(IO)~\cite{greer2023inverse}                      & 80.2           & 5e-4                   & & ConstrICON(IO)~\cite{greer2023inverse}                  & 1.62             & 3e-6                                          & & SAME++(IO)~\cite{Tian2023SAMEAS}                              & 49             & yes                            \\
			ConstrICON~\cite{greer2023inverse}                        & 79.3           & 4e-6                   & & uniGradICON~\cite{tian2024unigradicon}                  & 2.26             & 9e-5                                          & & uniGradICON    ~\cite{tian2024unigradicon}                    & 48             & yes                            \\
			ConstrICON(IO)~\cite{greer2023inverse}                    & 80.1           & 0                      & & uniGradICON(IO)~\cite{tian2024unigradicon}              & 1.40             & 9e-5                                          & & uniGradICON(IO)~\cite{tian2024unigradicon}                    & 52             & yes                            \\
			uniGradICON~\cite{tian2024unigradicon}                    & 76.2           & 6e-5                   & & \textbf{CARL}                                           & 1.88             & 4e-6                                          & & \textbf{CARL}                                                 & 50             & yes                            \\
			uniGradICON(IO)~\cite{tian2024unigradicon}                & 78.9           & 2e-4                   & & \textbf{CARL(IO)}                                       & 1.34             & 6e-6                                          & & \textbf{CARL(IO)}                                             & 56             & yes                            \\
			EasyReg~\cite{iglesias2023ready}                          & 78.8           & -                      & & \multicolumn{3}{c}{\cellcolor{white}}                                                                                      & & \multicolumn{2}{c}{\cellcolor{white}}                                                                           \\
			KeyMorph~\cite{Yu22a}                                     & 67.2           & 0                      & & \multicolumn{3}{c}{\cellcolor{white}}                                                                                      & & \multicolumn{2}{c}{\cellcolor{white}}       \\
                        \textbf{CARL}                                             & 79.6           & 6e-3                   & & \multicolumn{3}{c}{\cellcolor{white}}                                                                                      & & \multicolumn{2}{c}{\cellcolor{white}}\\ 
                        \textbf{CARL(IO)}                                         & 80.4           & 1e-3                   & & \multicolumn{3}{c}{\cellcolor{white}}                                                                                      & & \multicolumn{2}{c}{\cellcolor{white}}\\ 
                        \textbf{CARL\{ROT\}}                                      & 77.8           & 1e-3                   & & \multicolumn{3}{c}{\cellcolor{white}}                                                                                      & & \multicolumn{2}{c}{\cellcolor{white}}\\ 
                        \textbf{CARL\{ROT\}(IO)}                                  & 78.9           & 3e-3                   & & \multicolumn{3}{c}{\cellcolor{white}}                                                                                      & & \multicolumn{2}{c}{\cellcolor{white}}\\ 
   \bottomrule
		\end{tabular}
	\end{adjustbox}
	\vskip-0.1em
\end{table*}

\subsection{Rotation equivariance}

Our approach is formally $[W, U]$ equivariant to translations. In addition, we find empirically that it can be made nearly $[W, U]$ equivariant to rotations by data augmentation. The following changes to the training procedure are required: \emph{First}, $\Xi_\theta$ is made formally equivariant to axis aligned $\pi$ rotations by evaluating the encoder four times, once for each possible $\pi$ rotation, and averaging. If the encoder is equivariant in the natural sense to transforms from a class $\mathcal{T}$, then $\Xi_\theta$ becomes $[W, U]$ equivariant to that class as mentioned in \cref{sec:coordinate_attention_block}. \emph{Second}, the receptive field of the encoder is broadened by adding an additional dilated convolution to the residual blocks with dilation 8. This empirically stabilizes training, but slightly damages translational equivariance via increased boundary effects. In addition, we change the extrapolation used when evaluating displacement field transforms at locations outside $[0, 1]$ (see \cref{sec:improved_extrapolation} in the Appendix). This change allows the Jacobian of the final transform to be continuous across this boundary when the displacement field represents a large rotation. \emph{Finally}, the dataset is augmented with random rotations from $\text{SO}(3)$. Because this augmentation requires large rotations to align the images, pretraining with naive diffusion regularization (\ie $\mathcal{L}_\text{reg}(\varphi) = ||\nabla \varphi - \boldsymbol{I}||^2_F$) is no longer appropriate as diffusion forbids transforms with Jacobians far from the identity map. During the pretraining step, it is therefore necessary to apply the diffusion regularizer to $R^{-1} \circ \varphi \circ Q$ instead of $\varphi$, where $R$ and $Q$ are the augmentations applied to the input images and $\varphi$ is the output of the overall architecture. As the network becomes $[W, U]$ equivariant to arbitrary rotations, $\nabla (R^{-1} \circ \varphi \circ Q)$ becomes close enough to the identity matrix for diffusion regularization to be appropriate: see \cref{sec:augmentation}. We refer to CARL trained using this procedure as CARL\{ROT\}. While baseline CARL can handle rotations of up to 45 degrees with instance optimization, CARL\{ROT\} performs correct registration when the input images are arbitrarily rotated (\cref{fig:hcp_translation}).

\subsection{Limitations}
\label{subsection:limitations}


Registering a pair of images with a forward pass takes 2.1 seconds. Instance optimization (IO) is slower: 209 seconds on an NVIDIA A100 GPU. 
However, we consider the runtime of IO a minor issue at the moment, and focus on accuracy, as we anticipate that hardware and software for computing the flash attention kernel will continue to improve.

\vskip -1.5em

\section{Conclusion}
We demonstrated that our proposed multi-step architecture has a novel combination of strong equivariance (by construction to translation and optionally for rotation by judicious data augmentation) and precise handling of complex and high-resolution deformations. We show that it has close to SOTA accuracy on lung registration when compared to a competitive leaderboard, SOTA performance on brain registration with empirically verified equivariance to translations and rotations, and significantly improves registration accuracy over previous unsupervised registration approaches for the challenging registration of raw abdomen scans with differing fields of view. 

\textbf{Acknowledgments}
This research was, in part, funded by the National Institutes of Health (NIH) under awards 1R01AR082684, 1R21MH132982, 1R01HL149877, and RF1MH126732. The views and conclusions contained in this document are those of the authors and should not be interpreted as representing official policies, either expressed or implied, of the NIH.

	{
		\small
		\bibliographystyle{ieeenat_fullname}
		\bibliography{references}
	}

\input{sec/X_suppl}

\end{document}

%% file: preamble.tex
\usepackage{adjustbox}
\usepackage{colortbl}
\usepackage{graphicx}

\usepackage{amsmath}
\usepackage{multirow}
\usepackage{amssymb}
\usepackage{inconsolata}
\usepackage{pifont}
\usepackage{tikz}
\usepackage[most]{tcolorbox}
\usepackage{array}

\definecolor{verylightgray}{gray}{0.98}

\DeclareMathOperator{\convo}{Conv}

\usetikzlibrary{arrows.meta}
\usetikzlibrary{positioning}

\newcommand{\lsim}{\mathcal{L}_{\text{sim}}}

\newcommand{\real}[0]{\mathbb{R}}

\newcommand{\ia}[0]{{I^\text{M}}}
\newcommand{\ib}[0]{{I^\text{F}}}

%% file: sec/0_abstract.tex
\begin{abstract}
Image registration estimates spatial correspondences between image pairs. These estimates are typically obtained via numerical optimization or regression by a deep network. A desirable property is that a correspondence estimate (e.g., the true oracle correspondence) for an image pair is maintained under deformations of the input images. Formally, the estimator should be equivariant to a desired class of image transformations. In this work, we present careful analyses of equivariance properties in the context of multi-step deep registration networks. Based on these analyses we 1) introduce the notions of $[U,U]$ equivariance (network equivariance to the same deformations of the input images) and $[W,U]$ equivariance (where input images can undergo different deformations); we 2) show that in a suitable multi-step registration setup it is sufficient for overall $[W,U]$ equivariance if the first step has $[W,U]$ equivariance and all others have $[U,U]$ equivariance; we 3) show that common displacement-predicting networks only exhibit $[U,U]$ equivariance to translations instead of the more powerful $[W,U]$ equivariance; and we 4) show how to achieve multi-step $[W,U]$ equivariance via a coordinate-attention mechanism combined with displacement-predicting networks. Our approach obtains excellent practical performance for 3D abdomen, lung, and brain medical image registration. We match or outperform state-of-the-art (SOTA) registration approaches on all the datasets with a particularly strong performance for the challenging abdomen registration.
\end{abstract}

%% file: sec/X_suppl.tex
\clearpage
\setcounter{page}{1}
\maketitlesupplementary

\begin{figure*}[tp!]
    \centering
    \includegraphics[width=.96\textwidth]{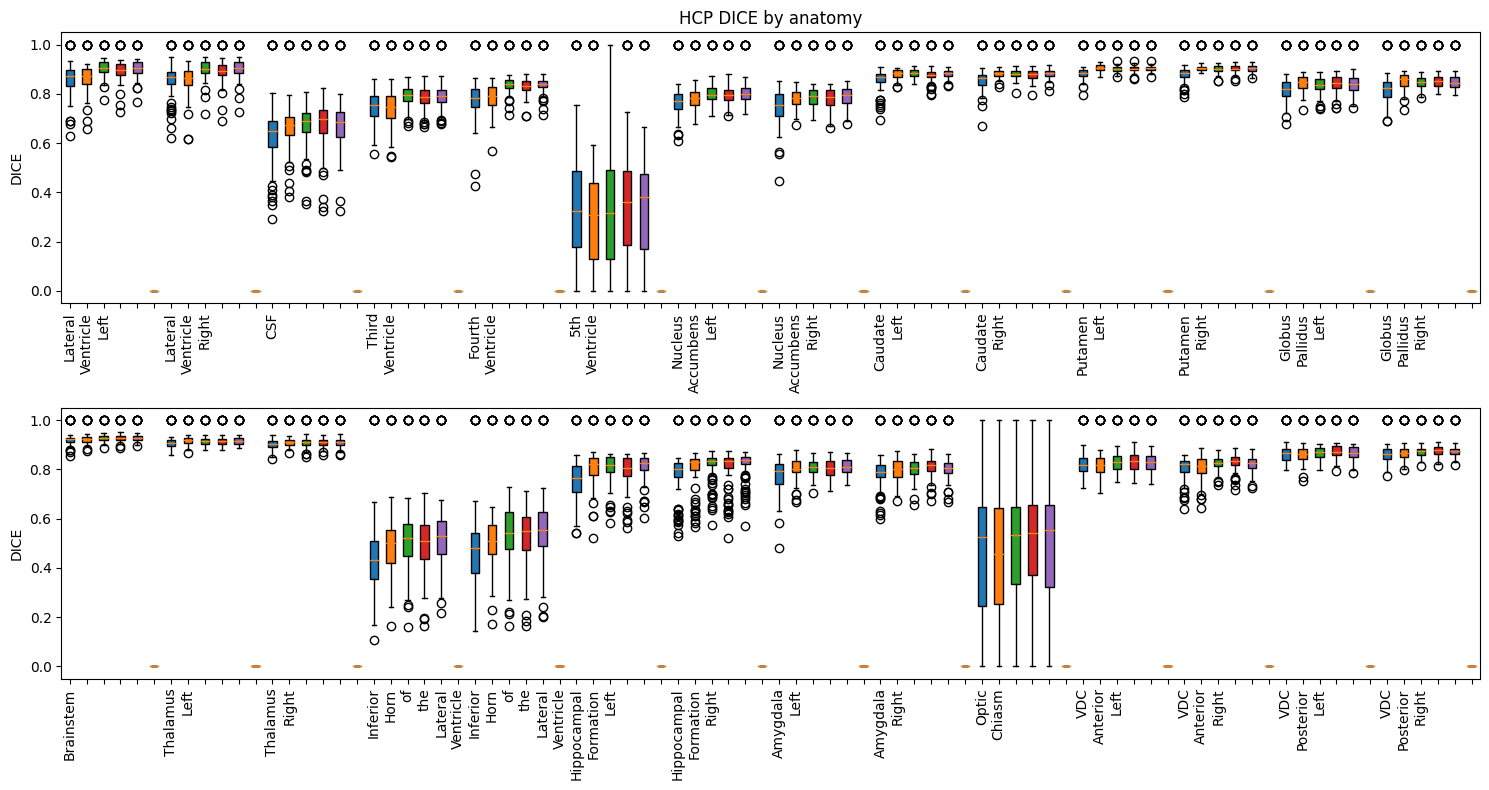}
    \includegraphics[width=.43\textwidth]{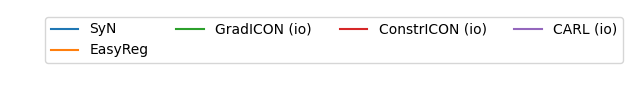}
    \caption{Per structure DICE scores on the HCP dataset. CARL ranks well on most structures.}
    \label{fig:box_hcp}
\end{figure*}
\begin{figure*}[tp!]
    \centering
    \includegraphics[width=.96\textwidth]{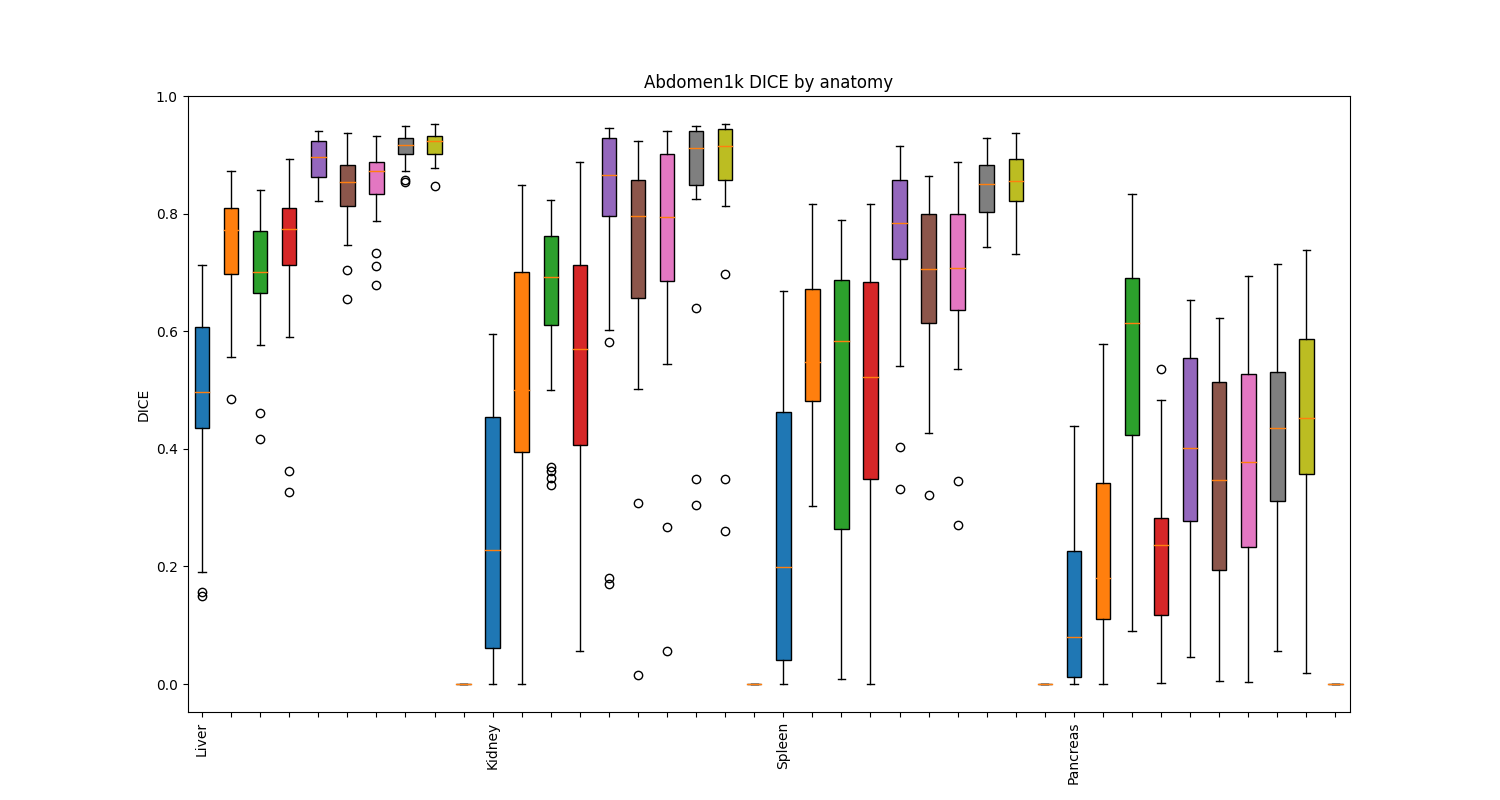}
    \includegraphics[width=.43\textwidth]{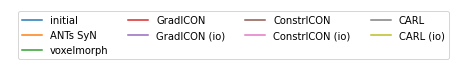}
    \caption{Per structure DICE scores on the Abdomen1k dataset. We observe that CARL is dramatically ahead of competing methods on liver, kidney, and spleen registration, but performs meaninfully worse than Voxelmorph on the pancreas.}
    \label{fig:box_abd}
\end{figure*}

\section{Per Structure DICE Box Plot}
To provide a more comprehensive picture of how different registration algorithms perform for different brain structures or organs (instead of purely reporting averages) Figs.~\ref{fig:box_hcp} and~\ref{fig:box_abd} show anatomy-specific boxplots. We observe especially strong performance for CARL on the Abdomen1k dataset (Fig.~\ref{fig:box_abd}) with excellent registration results for liver, kidney, and spleen. 

\section{Resolution, Downsampling \& Coordinates}
\label{downsample}

We use an internal convention that regardless of resolution, images have coordinates ranging from (0, 0, 0) to (1, 1, 1). Thus, a transform, a function from $[0, 1]^D \rightarrow \mathcal{R}^N$, can be applied to an image of any resolution. This allows us to construct a multiresolution, multi-step registration algorithm using TwoStep and the operator Downsample defined in \cite{greer2021icon} as
\begin{align}
&\text{Downsample}\{\Phi\}[\ia, \ib] \nonumber \\
&= \Phi[ \text{averagePool}(\ia, 2), \text{averagePool}(\ib, 2)]\,. \\
&\text{TwoStep}\{\Phi, \Psi\}[I^M, I^F] \nonumber \\ &= \Phi[I^M, I^F] \circ \Psi[I^M \circ \Phi[I^M, I^F], I^F]\,.\end{align}

\section{Implementing the diffeomorphism-to-diffeomorphism case}
\label{subsection:xinetimplementation}

\begin{figure*}[htp]
    \centering
    \includegraphics[width=.5\textwidth]{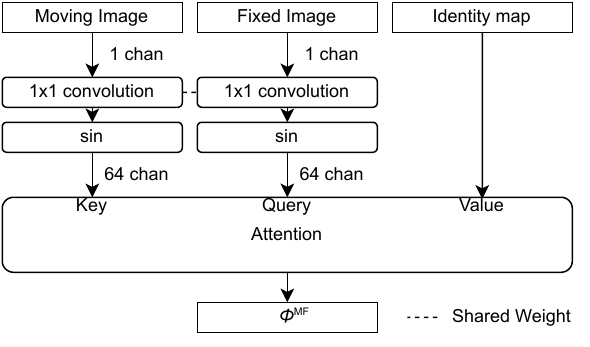}
    \includegraphics[width=.43\textwidth, trim={17cm 0cm 0cm 0cm}, clip]{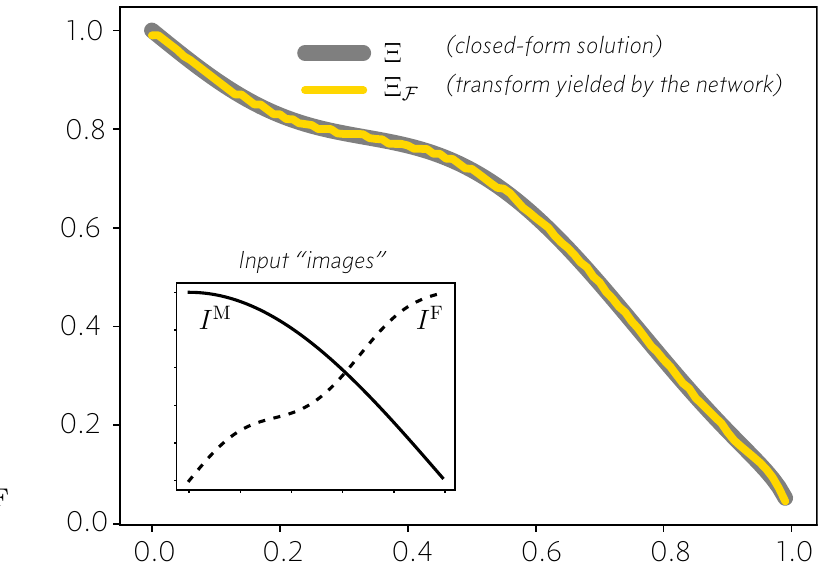}
    \caption{\label{xinet}\emph{Left}: Neural network $\Xi_\mathcal{F}$ implementing $\Xi$. \emph{Right}: Result of registering the 1-dimensional "images" $\ia : [0, 1] \rightarrow[0, 1], x \mapsto \cos(\frac{\pi}{2} x)$ and $\ib : [0, 1] \rightarrow[0, 1], x \mapsto x + 0.07 \sin(3\pi x)$ via $\Xi$ and $\Xi_\mathcal{F}$, illustrating that the resulting maps are equivalent. $\Xi$ is computable here as these images are invertible and smooth. The neural network output (gold) closely matches the analytical solution (i.e., $\Xi[\ia, \ib] = \ia^{-1} \circ \ib = \frac{2}{\pi} \cos^{-1}(x + 0.07 \sin(3\pi x))$, gray).\label{Fig:analytincal} \emph{Best-viewed in color.}}
\end{figure*}

We can use coordinate attention to solve the diffeomorphism-to-diffeomorphism registration problem with a neural network $\Xi_\mathcal{F}$ (shown in the left part of Fig.~\ref{xinet}).


The input functions $I^\text{M}$, $I^\text{F}$, and the output transform are approximated as arrays of voxels. The functional $\Xi$ such that $\Xi[I^M, I^F]:= (I^M)^{-1} \circ I^F$ (which only operates on images that are diffeomorphic) can be directly implemented, without training, using standard neural network components. We refer to this implementation as $\Xi_\mathcal{F}$. The intention is to map each voxel in the moving image into a high dimensional vector that will have a large dot product with the corresponding voxel in the fixed image with the same value, and then compute the attention matrix with the embedded fixed image voxels as the queries and the embedded moving image voxels as the keys. Subsequently, we can compute the center of mass of the attention masks (i.e., where each fixed image voxel matches on the moving image) by setting the \emph{values} to be the raw coordinates of the moving image voxels. We choose for the embedding a $1\times 1$ convolution with large weights followed by a sine-nonlinearity, which has the desired property of two vectors having a large dot product only when their input intensities are similar. Because our images are diffeomorphisms, we know a-priori that the input intensity of our moving image will only be close to intensities of the fixed image in a small region. We verify that this network, without any training, reproduces $\Xi$ when applied to input images that are diffeomorphisms, see~Fig.~\ref{Fig:analytincal} (right).
\vspace{-0.2cm}
\subsection{Limitation on equivariance feasibility}

In Sec.~\ref{subsection:xinetimplementation}, we turned images into features using $1\times 1$
convolution followed by a sine nonlinearity which, since it is a function applied pointwise, is perfectly equivariant. This
worked since the images to be registered were diffeomorphisms, and hence each intensity vector was unique. However, since, as we are about to prove, we cannot achieve equivariance to arbitrary diffeomorphisms for registering
real images, we have to sacrifice some equivariance
in order to expand the set of valid inputs. This drives our choice to target translation and rotation equivariance out of the set of possible diffeomorphisms.

\textbf{Claim.} It is impossible to have an algorithm that is $[W, U]$
equivariant to any non-identity class of transforms and can be applied to arbitrary
images.

\vskip1ex
\noindent
\textbf{Counterexample.}  Assume that $\Phi$ is a $[W, U]$ equivariant algorithm for all $W, U \in$ diffeomorphisms, and that is valid for all input images. We ask it to register the images $\ia, \ib := 0$. Then, for a non identity $W$ and  $U$ picked to be identity,
\begin{align}
    \Phi[\ia, \ib] &= W^-1 \circ \Phi[\ia \circ W, \ib \circ U] \circ U\\
    \Phi[\ia, \ib] &= W^-1 \circ \Phi[\ia \circ W, \ib]\\
    \Phi[0, 0] &= W^ {-1} \circ \Phi[0, 0]\\
    id &= W^{-1} ~ ,
    \end{align}where 0 indicates an image that is zero everywhere. This yields a contradiction.

\vskip1ex

\textbf{Claim.} It is impossible to have an algorithm that is $[W, U]$
equivariant to rotations and can be applied to rotationally symmetric images.

\vskip1ex
\noindent
\textbf{Counterexample.}  Assume that $\Phi$ is a $[W, U]$ equivariant algorithm for all $W, U \in$ rotations, and that $\Phi$ is valid for input images including at least one rotationally symmetric image $\ia$ (such that for a non identity $W,~\ia \circ W = \ia$.) We ask it to register the images $\ia, \ib$. Then, for a non identity $W$ with respect to which $\ia$ is symmetric and  $U$ picked to be identity,
\begin{align}
    \Phi[\ia, \ib] &= W^{-1} \circ \Phi[\ia \circ W, \ib \circ U] \circ U\\
    \Phi[\ia, \ib] &= W^{-1} \circ \Phi[\ia \circ W, \ib]\\
    \Phi[\ia, \ib] &= W^ {-1} \circ \Phi[\ia, \ib]\\
    id &= W^{-1} ~ ,
    \end{align} This yields a contradiction, as we assumed W was not identity.

We conclude that there is a tradeoff. If there is a valid input image $I$ and a nonzero transform $T$ such that $I \circ T = I$, then $T$ cannot be in the class of transforms with respect to which $\Phi$ is $[W, U]$ equivariant. For a simple example, an algorithm that registers images of perfect circles cannot be $[W, U]$ equivariant to rotations. For a practical example, since brain-extracted brain images have large areas outside the brain that are exactly zero, algorithms that register such preprocessed brain images cannot be $[W, U]$ equivariant to transforms that are identity everywhere in the brain but have deformations outside the brain. To modify $\Xi_\mathcal{F}$ so that it can apply to a broader class of images other than "images that happen to be diffeomorphisms", we thus have to restrict the transforms with respect to which it is $[W, U]$ equivariant. 

\subsection{Guarantee given equivariance}
\label{guarantee}
While it is unfortunate that we cannot achieve [W, U] equivariance to arbitrary diffeomorphisms for arbitrary input images, there is great advantage to expanding the group of transforms $\mathcal{T}$ with respect to which our algorithm is [W, U] equivariant. The advantage is as follows: for any input image pair $\ia, \ib$ where the images can be made to match exactly by warping $\ia$ by a transform $U$ in $\mathcal{T}$, an algorithm that outputs the identity map when fed a pair of identical images and is $[W, U]$ equivariant with respect to $\mathcal{T}$ will output U for $\ia, \ib$. We see this as follows.

Assume $\Phi$ outputs the identity transform when fed identical fixed and moving images and $[W, U]$ equivariant with respect to $\mathcal{T}$, and $\ia \circ U = \ib$ for a $U \in \mathcal{T}$. Then

\begin{align}
    \Phi[&\ia, \ib] \\
    &= \Phi[\ia, \ia \circ U] \\
    &= \Phi[\ia, \ia] \circ U \\
    &= U\,,
\end{align}
where $\Phi[\ia, \ia \circ U]=\Phi[\ia, \ia] \circ U$ holds because of the $[W,U]$ equivariance with $W$ being the identity transform. 

We note that before training, the CARL architecture emphatically does not have the property of outputting the identity map when fed identical images: instead, it learns this property from the regularizer during training.

\section{Performance implications of two step registration}
We observed in Sec.~\ref{retinaexpr} that $\Xi_\theta$ trains significantly better as the beginning of a multistep algorithm than on its own. Here, we examine why that may be, while removing as much complexity as possible. Our finding suggests that two step registration assists training by functioning as a similarity
measure with better capture radius.

First, we briefly train a single step network $\Phi$ on the \textbf{Baseline} task from Sec.~\ref{retinaexpr}, i.e. we stop training before convergence. Then, we examine the loss
landscape of a trivial "fixedTranslation" neural network $\tau$ to
register \textbf{Baseline}. This network has a single parameter, $t$, and it ignores
its input images and always shifts images to the right by $t$: that is
\begin{equation}
	\tau[\ia, \ib](\vec{r}) = \vec{r} + \begin{bmatrix}t \\ 0\end{bmatrix}\,.
\end{equation}

The optimal value of $t$ is zero since there is no bias towards left or right
shift of images in this dataset- but if we were to train $\tau$ on LNCC similarity, how well would the gradients drive t to zero?

\begin{figure}[htp]
	\centering
	\includegraphics[width=.45\textwidth]{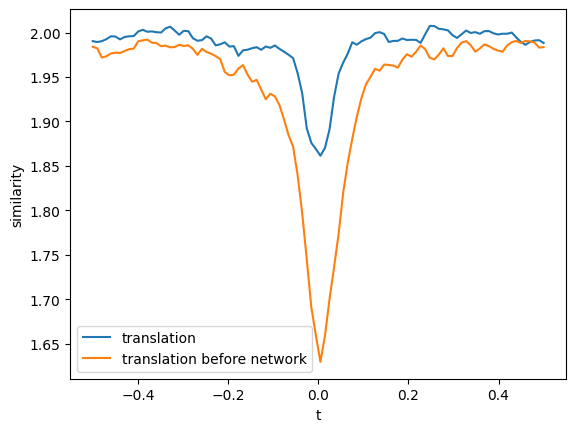}
	\includegraphics[width=.45\textwidth]{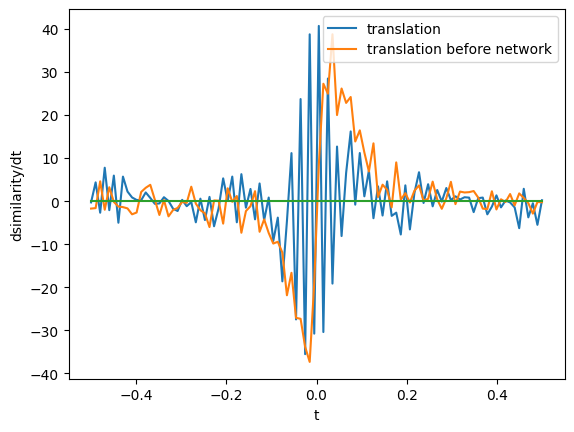}
	\caption{The loss of $TwoStep\{\tau, \Phi\}$ (i.e., translation before network) as a function of t is much better behaved than the loss of $\tau$ (i.e., translation) as a function of t. The capture radius of the former is larger and the loss is overall smoother close to the correct solution as shown on the bottom.}
	\label{fig:landscapces}
\end{figure}

We plot $LNCC(\ia \circ \tau[\ia, \ib], \ib)$ against $t$ compared to $LNCC(\ia
	\circ TwoStep\{\tau, \Phi\}[\ia, \ib], \ib)$. We also plot
$\frac{\partial}{\partial t} LNCC(\ia \circ \tau[\ia, \ib], \ib)$ and
$\frac{\partial}{\partial t}LNCC(\ia \circ TwoStep\{\tau, \Phi\}[\ia, \ib],
	\ib)$ using PyTorch's back-propagation. Fig.~\ref{fig:landscapces} shows the result indicating that multi-step registration results in better capture radius.

We conjecture that when two step network $\text{TwoStep}\{\tau, \Phi\}$ is trained with an LNCC loss, the loss function seen by $\tau$ is not simply LNCC, but instead the loss function seen by $\tau$ is actually the
performance of $\Phi$ (which is measured by LNCC), which is an implicit loss function with a better capture
radius than the original LNCC loss function.

\section{Computational Budget}

Each 100,000 step training run of CARL takes 14 days on 4 RTX A6000 GPUs or 6 days on four A100 GPUS. In total, 336 GPU days were spent developing the CARL architecture and training the final models.

An additional 45 GPU days were spent training comparison methods. 14 server days were spent training KeyMorph variants, although the published KeyMorph code is io-bound and did not significantly load the server's GPU.

\section{Comparison Methods Details}

\subsection{Abdomen1k}
For Abdomen1k, we trained all methods using their published code and default hyperparameters.

\subsection{DirLab Lung}
On the DirLab challenge set, all results of comparison methods are taken from the literature. Results of ANTs, Elastix, Voxelmorph, and LapIRN are from \cite{tian2022}. The remainder are from their original publications.

\subsection{HCP}
On HCP, we evaluated ANTs using code from \cite{tian2022}. Results of GradICON, and ConstrICON are from the ConstrICON publication. We evaluated KeyMorph by training a model on the HCP 
Dataset using KeyMorph's published code and hyperparameters for the IXI dataset. We evaluated Easyreg using its published weights, which are advertised to be appropriate for the HCP dataset. We measured the equivariance of the GradICON method using GradICON's published code and weights.

\section{Abdomen 1k Test Pairs}
\label{abdomen_fulldata}

We used the following 30 image pairs to evaluate our abdomen registration experiments.

00817 00872,
00808 00832,
00815 00863,
00857 00860,
00883 00848,
00826 00812,
00862 00803,
00849 00855,
00877 00800,
00857 00834,
00829 00875,
00813 00840,
00803 00802,
00803 00883,
00869 00801,
00848 00887,
00827 00854,
00803 00867,
00828 00856,
00863 00870,
00829 00844,
00829 00886,
00828 00858,
00837 00802,
00853 00871,
00882 00812,
00823 00880,
00837 00815,
00842 00864,
00854 00864

\section{Extension to Rotational Equivariance}
\label{sec:augmentation}

The first solution to obtain a registration network that exhibts rotation equivariance that comes to mind is to simply augment the training dataset with random rotations, and see if the network can still register it. This conceptually works fine for our main training with GradICON regularization, but breaks when directly applied to our diffusion regularized pretraining (which is empirically required for training a coordinate attention layer).
That is , the training loss would look like
\begin{align}
    R, Q &\sim \text{Uniform}(\text{Rotations}) \\
    I^M, I^F &\sim \text{Dataset} \nonumber \\
    \hat{I}^M, \hat{I}^F &:= (I^M \circ R), (I^F \circ Q) \nonumber \\
    \text{minimize}:& ~\lsim(\hat{I}^M \circ \Phi[\hat{I}^M, \hat{I}^F], \hat{I}^F) + \mathcal{L}_\text{reg}(\Phi[\hat{I}^M, \hat{I}^F]) \nonumber
\end{align}
This cannot be reliably trained with a diffusion regularizer, because to align $\hat{I}^M$ to $\hat{I}^F$  will require transforms with Jacobians of the transformation map that are very far from the identity map as they will need to express large-scale rotations. 

Our proposed solution is to move the augmentation "inside" the losses, in the following sense:

First, expand our augmented images $\hat{I}^M, \hat{I}^F$
\begin{align}
    R, Q &\sim \text{Uniform}(\text{Rotations}) \nonumber \\
    I^M, I^F &\sim \text{Dataset} \nonumber \\
    &\text{minimize}:  \nonumber \\
    &\lsim((I^M \circ R) \circ \Phi[I^M \circ R, I^F \circ Q], I^F \circ Q) \nonumber \\
    &+ \mathcal{L}_\text{reg}(\Phi[I^M \circ R, I^F \circ Q])
\end{align}
In this expanded form, change to the following:
\begin{align}
    R, Q &\sim \text{Uniform}(\text{Rotations}) \nonumber \\
    I^M, I^F &\sim \text{Dataset} \nonumber \\
    &\text{minimize}:  \nonumber \\
    &\lsim(I^M \circ (R \circ \Phi[I^M \circ R, I^F \circ Q] \circ Q^{-1}, I^F) \nonumber \\
    &+ \mathcal{L}_\text{reg}(R \circ \Phi[I^M \circ R, I^F \circ Q] \circ Q^{-1})
\end{align}
Then, collect like terms. It becomes clear that the augmentation is now \emph{inside} the loss and connected to the network. 
\begin{align}
    R, Q &\sim \text{Uniform}(\text{Rotations}) \nonumber \\
    I^M, I^F &\sim \text{Dataset} \nonumber \\
    \hat{\Phi}[I^M, I^F] &:= R \circ \Phi[I^M \circ R, I^F \circ Q] \circ Q^{-1} \\   
    \text{minimize}:& ~\lsim(I^M, \hat{\Phi}[I^M, I^F], I^F) + \mathcal{L}_\text{reg}(\hat{\Phi}[I^M, I^F]) \nonumber
\end{align}
Now, while $\Phi$ outputs large rotations, on a rotationally aligned dataset $\hat{\Phi}$ outputs transforms with Jacobians near the identity, and so can be trained with diffusion regularization.

\section{Improved Extrapolation of Displacement Fields}
\label{sec:improved_extrapolation}

Displacement fields $(\text{disp})$ are stored as grids of vectors associated with coordinates in $[0, 1]^D$. In ICON \cite{greer2021icon}, Greer et al. noted that the method of extrapolating when evaluating a transform outside of this region is important when composing transforms, since transforms such as translations and rotations move some coordinates from inside $[0, 1]^D$ to outside it. They propose coordinate by coordinate clipping before interpolating into  \emph{the displacement field}

\begin{equation}
    \varphi_\text{disp}(x) = x + \text{interpolate}(\text{disp}, \text{clip}(x, 0, 1))\,.
\end{equation}

This formulation has a discontinuous Jacobian on the boundary of $[0, 1]^D$, and in particular results in non-invertible transforms on the boundaries for large rotations.

We instead propose 

\begin{align}
\text{clip}(x) &= x - 
    \begin{cases}
     x& \text{if } x < 0\\
    0,              & \text{otherwise}
\end{cases}
    -
    \begin{cases}
     (x - 1)& \text{if } x > 1\\
    0,              & \text{otherwise}
\end{cases}\\
    \text{reflect}(x) &= x - 
    \begin{cases}
     2x& \text{if } x < 0\\
    0,              & \text{otherwise}
\end{cases}
    -
    \begin{cases}
     2(x - 1)& \text{if } x>  1\\
    0,              & \text{otherwise}
\end{cases}\\
    \varphi_\text{disp}(x) &= x + 2~\text{interpolate}(\text{disp}, \text{clip}(x)) \\
    & - \text{interpolate}(\text{disp}, \text{reflect}(x))
\end{align}

which is identical inside $[0, 1]^D$ but has continuous Jacobian over the boundary.

\begin{figure}
    \centering
    \includegraphics[width=0.7\linewidth]{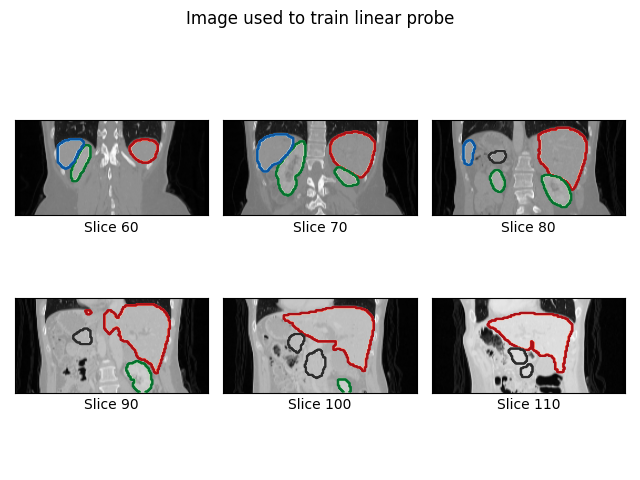}
    \includegraphics[width=0.7\linewidth]{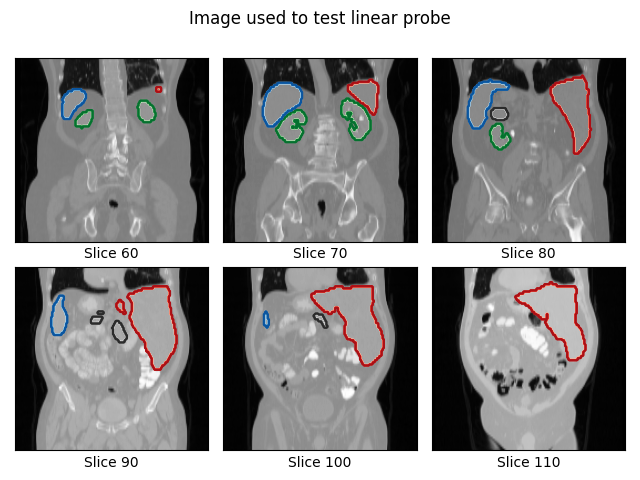}
    \includegraphics[width=0.7\linewidth]{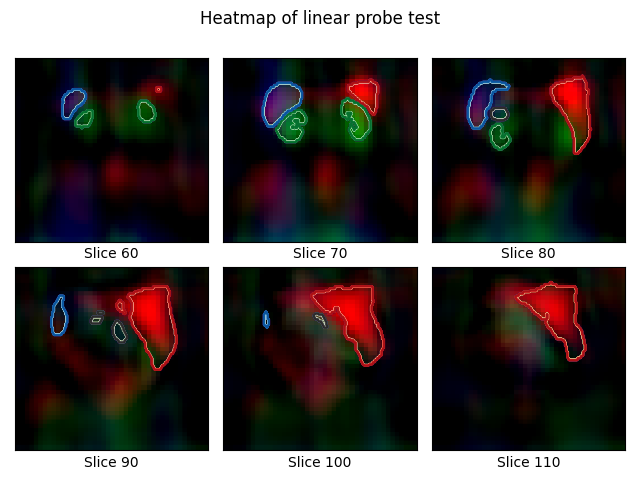}
    \includegraphics[width=0.7\linewidth]{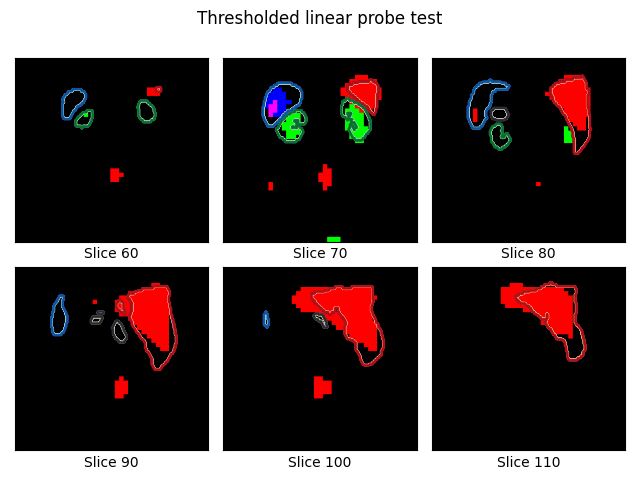}
    \caption{A linear probe is used to aid interpretability of the features learned by the convolutional encoder. The linear probe is trained on one image, and then its output heat maps are visualized on another image. The red, green, and blue channels are used to indicate the liver, kidney, and spleen respectively. The grey channel is used to indicate the pancreas, although no direction is found in the features that segments it.}
    \label{fig:linear_probe_interpretability}
\end{figure}
\section{Investigation of internal features of $\Xi_\theta$}

We use interpretability techniques to investigate the features learned by the convolutional encoder $\convo$ of $\Xi_\theta$. Two images are selected from the Abdomen1k test set, and independently encoded using the convolutional network. We convert these images into features using $\convo$. From these voxelwise features, we train linear probes to segment the kidneys, liver, pancreas, and spleen of a single train image by minimizing least squares error between ground truth and predicted label, and then visualize the probe's output on a second test image. This linear probe suggests (see \cref{fig:linear_probe_interpretability}) that the features, which were learned without any segmentations, include directions that measure liver-ness, spleen-ness, and kidney-ness, but there is no pancreas-direction. This may explain why our model is less accurate at registering the pancreas than the other organs.

\subsection{Verification that $\Xi_\theta$'s attention masks are compact}

\begin{figure}
    \centering
    \includegraphics[width=.9\linewidth, trim={1cm 0cm 1cm 0cm}, clip]{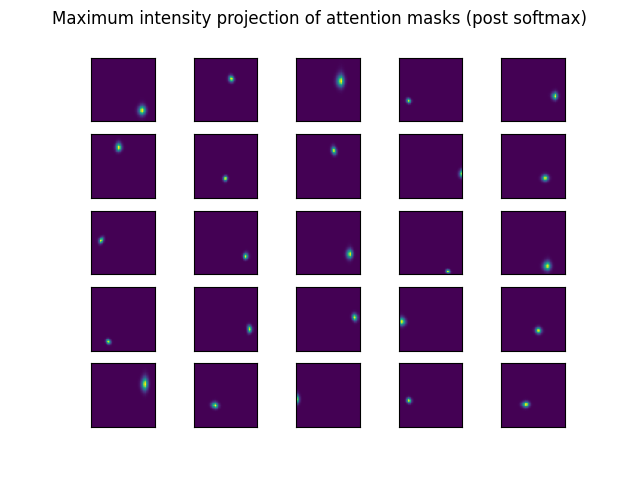}
    \caption{Sample attention masks from inside the coordinate attention block of CARL trained on Abdomen1k. The masks are compact, justifying the claim on which we build \cref{sec:wuproof}.}
    \label{fig:attention_samples}
\end{figure}


As an assumption in \cref{sec:wuproof} was that post training, attention masks are spatially compact. We verify this by computing the attention masks associated with the query vectors of 25 random voxels when registering the pair in \cref{fig:linear_probe_interpretability}, maximum intensity projecting them to get 2-D heatmaps, and plotting. As expected, we see in \cref{fig:attention_samples} that each pixel in the moving image attends to a small region in the fixed image. As a result, these attention masks will not immediately interact with the boundary of the padded feature volume when an image is translated. This property is required for [W, U] equivariance.

\onecolumn

\section{Fully elaborated proof that Coordinate attention is [W, U] equivariant}

Previously, we elided the difference between two definitions of an image: a function from $[0, 1]^D \rightarrow \mathbb{R}$ suitable for composition with transforms, and a function from voxel indices to intensities suitable for discrete convolution and attention. Here, we fully make this distinction explicit. We will continue to consider \emph{images} to be continuous, and discretize them as necessary by composing them with or interpolating them at the function coords which maps voxel indices to coordinates in $[0, 1]^D$. This explicit style makes clear that the proof is formally correct, and also more directly maps to the implementation. We use the linear interpolation function $\text{interpolate}(\text{points}, \text{values}, x)$ where x is the spatial location where we evaluate, and points and values are the locations where we know the value of the function (typically a grid). 

\textbf{Assumptions:} We assume that the feature encoders are translation equivariant like
\begin{equation}
	\convo_\theta(I \circ U) = \convo_\theta(I) \circ U\,.
    \label{convperm}
\end{equation}

Without positional embeddings or causal masking, (we do not use
either) the attention mechanism is equivariant to permutations as follows: for $P_1, P_2$ permutations; and the output and $K$ (Key), $Q$ (Query), and $V$ (Value) inputs represented each as a function from an index to a vector, and an attention block represented as $\mathbb{T}$,
\begin{equation} \mathbb{T}[K \circ P_1, Q \circ P_2, V \circ P_1] = \mathbb{T}[K, Q, V] \circ P_2\,. \label{transperm2}\end{equation}

In plain language, changing the order of the queries causes the order of the output of the attention operation to have its order changed in the same way, and changing the order of the keys and values has no effect as long as they are changed together.

Additionally, because the attention weights in an attention block sum to 1, for an affine function $f$, we have 

\textbf{Lemma}:
\begin{equation}\mathbb{T}[K, Q, f\circ V] = f \circ \mathbb{T}[K, Q, V]\,. \label{transaff2}\end{equation}

\textbf{Proof of lemma}: Once attention weights $w_i$ are computed, for each output token we produce a weighting function W

\begin{equation}
    W(\textbf{x}_i \dots) = \sum_j w_j \textbf{x}_j
\end{equation}

where $w_i$ sum to 1.

We also have an affine function f, that is 
\begin{equation}
    f(\textbf{x}) = A\textbf{x} + \textbf{b}
\end{equation}

We then observe that b is preserved and hence $f\circ W = W \circ f$ as long as $w_i$ sum to 1.

Finally, we assume that the attention mask associated with each query vector has small spatial
support. Finding a training procedure that reliably fulfilled this assumption across different datasets was nontrivial: we find that this assumption is satisfied after regularizing the  network end-to-end with diffusion
regularization for the first several epochs, and using GradICON regularization thereafter. This is a crucial empirical result that we find evidence for in \cref{fig:attention_samples}

With these assumptions, we prove that $\Xi_\theta$ is $[W, U]$ equivariant to translations below.

\textbf{Proof}: A translation $W$ by an integer number of voxels is both affine when seen as an operation on coordinates, $W_{x \mapsto x + r}$, and a permutation of the
voxels when seen as an operation on voxel images $W_{\text{permutation}}$ (as long as we can neglect boundary effects). The map from indices to coordinates, $\text{coords}$, serves as the bridge between these two representations of a transform ($W_{x \mapsto x + r} \circ \text{coords} = \text{coords} \circ W_{\text{permutation}})$. As long as the attention masks have small spatial support (and hence do not interact with the boundary), we can suppress boundary effects by padding with zeros before applying the operation. So, for translations $W$ and $U$, we have
\[\Xi_\theta[ I^M, I^F](x) :=\text{interpolate}(\text{coords}, \mathbb{T}[\convo_\theta (I^M \circ \text{coords}) , \convo_\theta(I^F \circ \text{coords}), \text{coords}], x)\enspace, \]
from which we establish that $\Xi_\theta$ is $[W, U]$ equivariant with respect to translation as follows: 
\begin{equation}
	\begin{split}
		& \Xi_\theta[ I^M \circ W, I^F \circ U](x)  =\text{interpolate}(\text{coords}, \mathbb{T}[\convo_\theta (I^M \circ W_{x \mapsto x + r} \circ \text{coords}) , \convo_\theta(I^F \circ U_{x \mapsto x + r} \circ \text{coords}), \text{coords}], x)          \\
        &=\text{interpolate}(\text{coords}, \mathbb{T}[\convo_\theta (I^M \circ \text{coords} \circ W_{\text{permutation}}) , \convo_\theta(I^F \circ \text{coords} \circ U_{\text{permutation}}), \text{coords}], x)          \\
		& \stackrel{\eqref{convperm}}{=} \text{interpolate}(\text{coords}, \mathbb{T}[\convo_\theta (I^M \circ \text{coords}) \circ W_{\text{permutation}} , \convo_\theta(I^F \circ \text{coords}) \circ U_{\text{permutation}}, \text{coords}], x)\\
		& \stackrel{\eqref{transaff2}}{=} \text{interpolate}(\text{coords}, W_{x \mapsto x + r}^{-1} \circ \mathbb{T}[\convo_\theta (I^M\circ \text{coords}) \circ W_{\text{permutation}}, \convo_\theta(I^F\circ \text{coords}) \circ U_{\text{permutation}}, W_{x \mapsto x + r} \circ \text{coords}], x) \\
		&= W_{x \mapsto x + r}^{-1} \circ \text{interpolate}(\text{coords}, \mathbb{T}[\convo_\theta (I^M\circ \text{coords}) \circ W_{\text{permutation}}, \convo_\theta(I^F\circ \text{coords}) \circ U_{\text{permutation}}, W_{x \mapsto x + r} \circ \text{coords}], x) \\
		& = W_{x \mapsto x + r}^{-1} \circ \text{interpolate}(\text{coords},  \mathbb{T}[\convo_\theta (I^M\circ \text{coords}) \circ W_{\text{permutation}}, \convo_\theta(I^F\circ \text{coords}) \circ U_{\text{permutation}}, \text{coords} \circ W_{\text{permutation}}] , x) \\
        & \stackrel{\eqref{transperm2}}{=} W_{x \mapsto x + r}^{-1} \circ \text{interpolate}(\text{coords},  \mathbb{T}[\convo_\theta (I^M\circ \text{coords}), \convo_\theta(I^F\circ \text{coords}) \circ U_{\text{permutation}}, \text{coords} ] , x) \\
        & \stackrel{\eqref{transperm2}}{=} W_{x \mapsto x + r}^{-1} \circ \text{interpolate}(\text{coords},  \mathbb{T}[\convo_\theta (I^M\circ \text{coords}), \convo_\theta(I^F\circ \text{coords}), \text{coords} ] \circ U_\text{permutation}, x) \\
        & = W_{x \mapsto x + r}^{-1} \circ \text{interpolate}(\text{coords} \circ U^{-1}_\text{permutation},  \mathbb{T}[\convo_\theta (I^M\circ \text{coords}), \convo_\theta(I^F\circ \text{coords}), \text{coords} ], x) \\
        & = W_{x \mapsto x + r}^{-1} \circ \text{interpolate}((U_{x \mapsto x + r})^{-1} \circ \text{coords},  \mathbb{T}[\convo_\theta (I^M\circ \text{coords}), \convo_\theta(I^F\circ \text{coords}), \text{coords} ], x) \\
        & = W_{x \mapsto x + r}^{-1} \circ \text{interpolate}(\text{coords},  \mathbb{T}[\convo_\theta (I^M\circ \text{coords}), \convo_\theta(I^F\circ \text{coords}), \text{coords} ], U_{x \mapsto x + r} (x)) \\
		& =  W^{-1} \circ \Xi_\theta[ I^M , I^F] \circ U\,.
	\end{split}
\end{equation}
Again, the same argument can also be applied to $[W, U]$ equivariance to axis aligned $\pi$ or $\frac{\pi}{2}$ rotations, provided that $\convo_\theta$ is replaced with an appropriate rotation equivariant encoder.

\begin{figure}[]
  \centering
  \includegraphics[width=.5\linewidth]{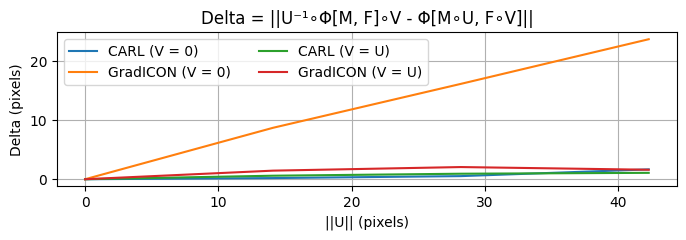}
  \vskip-1em 
   \caption{We directly confirm CARL is $[W, U]$ and GradICON is $[U, U]$ equivariant to translation on the deformed retina dataset.}
   \label{fig:onecol}
   \vskip-0.2cm
\end{figure}
\begin{table}[htp!]
    \centering
 \captionof{table}{CARL: ablation of final refinement layer (no IO)}
    \vskip-0.2cm
    \label{tab:ixi}
    \begin{tabular}{@{}lcccc@{}}
        \toprule
        & Abdomen1k DICE & HCP DICE & DirLab mTRE & L2R DICE \\
        \midrule
        w/o final refinement & 74.1 & 78.8 & 2.58 & 49 \\
        with final refinement & 75.7 & 79.6 & 1.88 & 50 \\
        \bottomrule
        
    \end{tabular}
\end{table}

\begin{figure}[htp]
	\centering

		\begin{tabular}{cccc}
			Moving Image & Warped (CARL) & Grid (CARL) & Fixed Image   \\ 
			\includegraphics[width=.20\textwidth, trim={0cm 19cm 19cm 1.5cm}, clip]{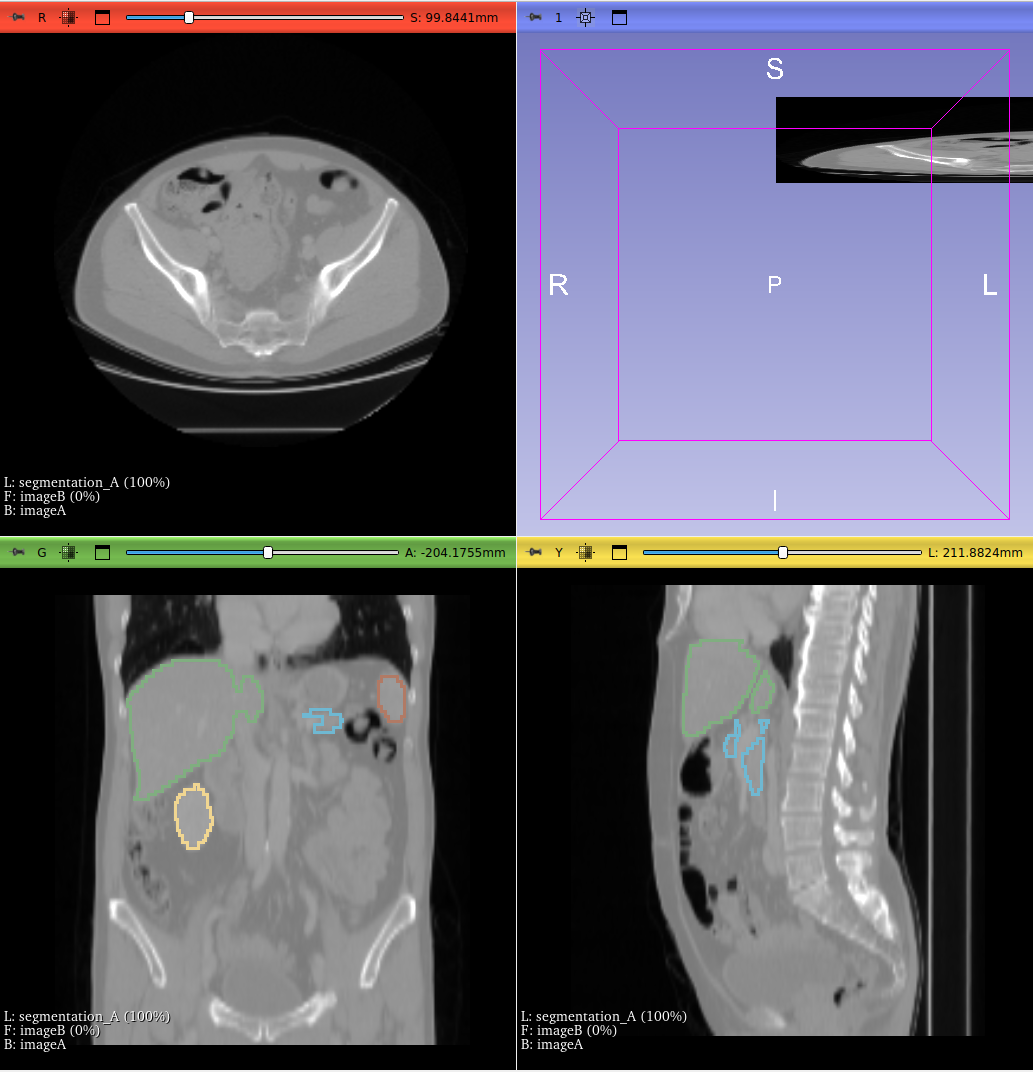}  &
			\includegraphics[width=.20\textwidth, trim={0cm 19cm 19cm 1.5cm}, clip]{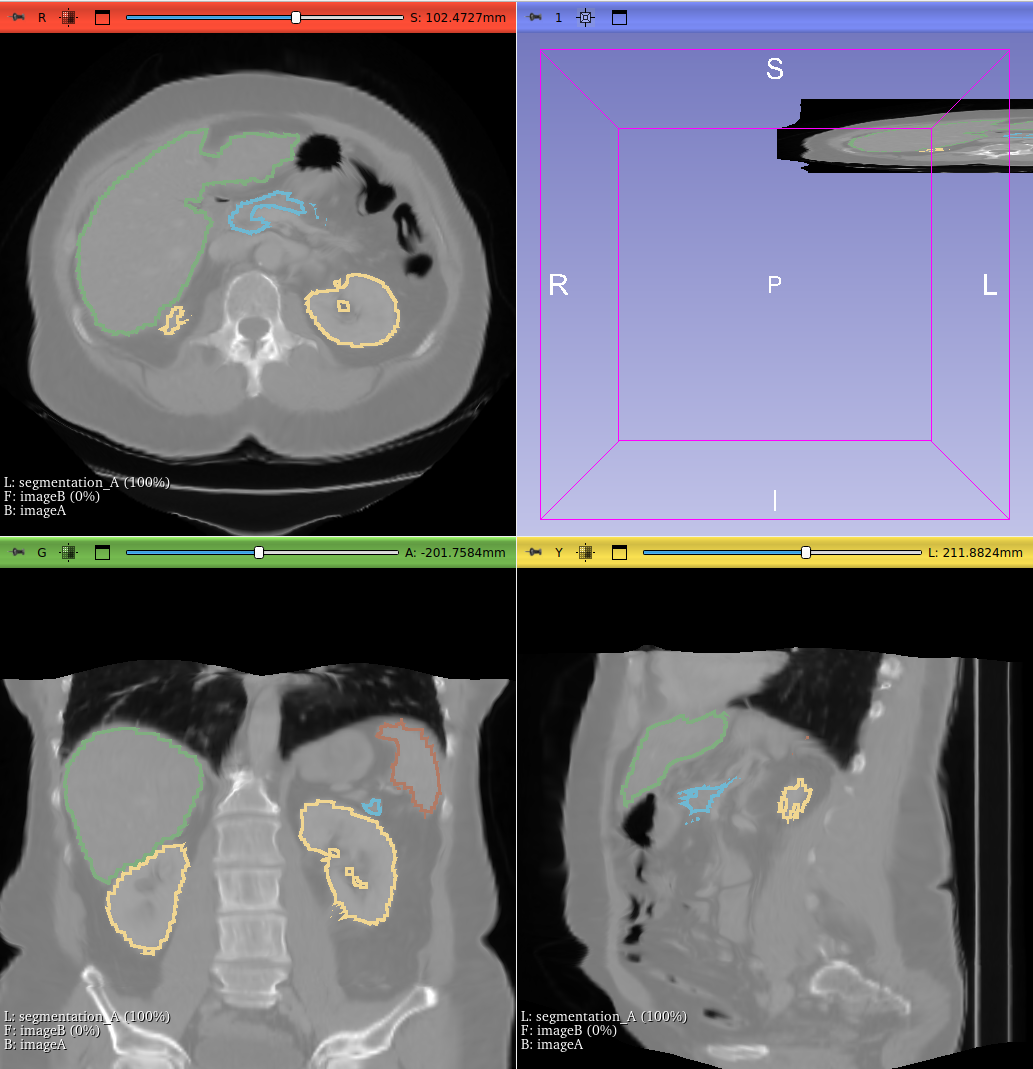} &
			\includegraphics[width=.20\textwidth, trim={0cm 19cm 19cm 1.5cm}, clip]{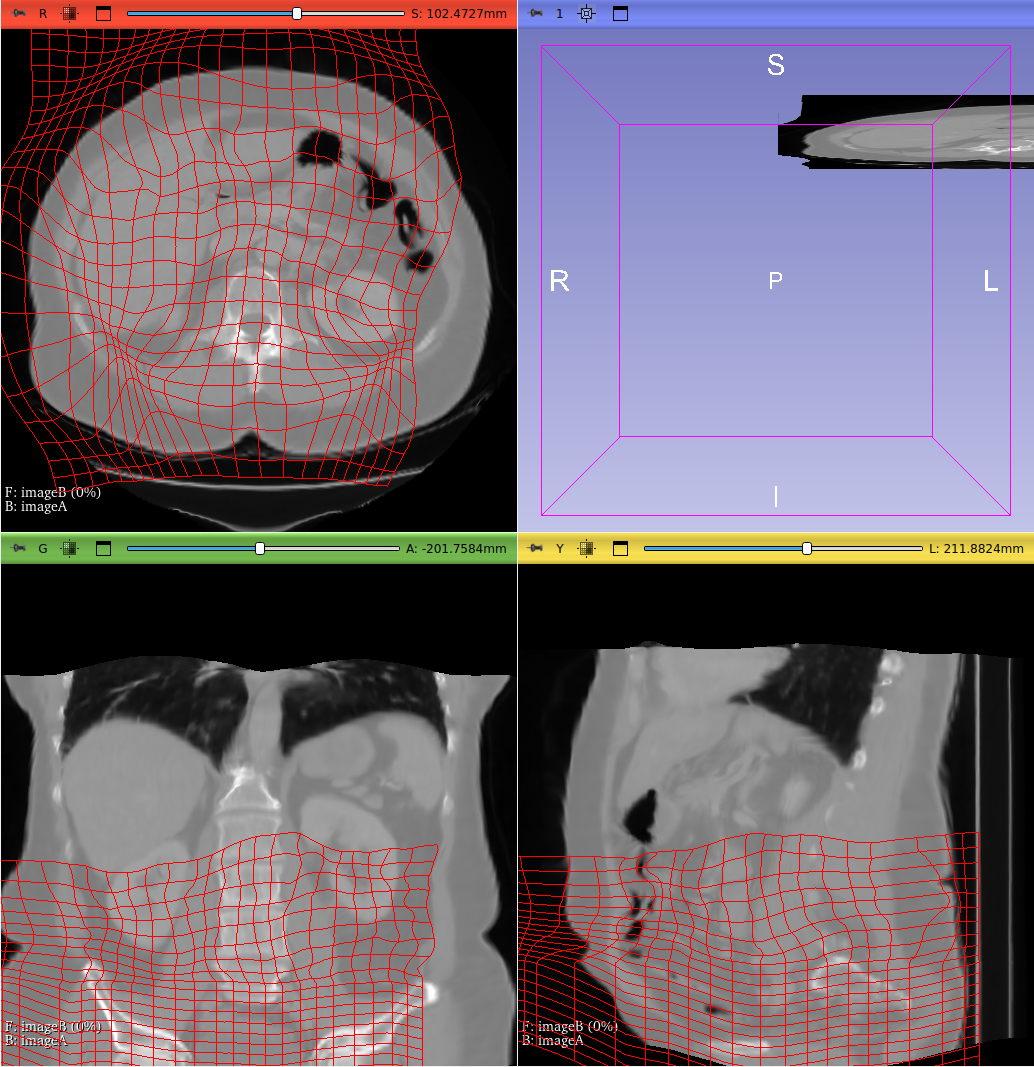} &
			\includegraphics[width=.20\textwidth, trim={0cm 19cm 19cm 1.5cm}, clip]{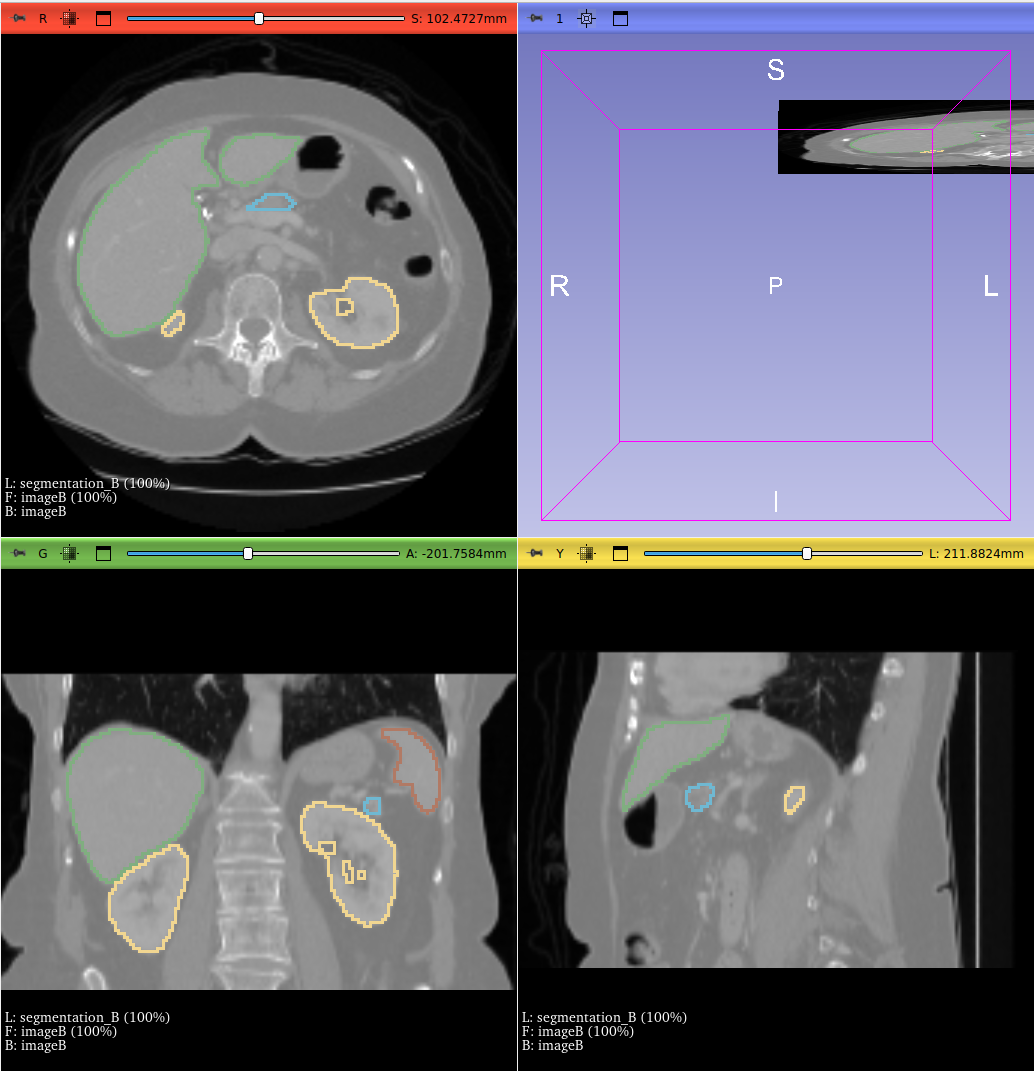}	\\
            \includegraphics[width=.20\textwidth, trim={0cm 1cm 19cm 20cm}, clip]{supfigs/abdomen_A.png}  &
			\includegraphics[width=.20\textwidth, trim={0cm 1cm 19cm 20cm}, clip]{supfigs/abdomen_warped_A.png} &
			\includegraphics[width=.20\textwidth, trim={0cm 1cm 19cm 20cm}, clip]{supfigs/abdomen_grid_A.png} &
			\includegraphics[width=.20\textwidth, trim={0cm 1cm 19cm 20cm}, clip]{supfigs/abdomen_B.png}	\\
            \includegraphics[width=.20\textwidth, trim={19cm 1cm 1cm 20cm}, clip]{supfigs/abdomen_A.png}  &
			\includegraphics[width=.20\textwidth, trim={19cm 1cm 1cm 20cm}, clip]{supfigs/abdomen_warped_A.png} &
			\includegraphics[width=.20\textwidth, trim={19cm 1cm 1cm 20cm}, clip]{supfigs/abdomen_grid_A.png} &
			\includegraphics[width=.20\textwidth, trim={19cm 1cm 1cm 20cm}, clip]{supfigs/abdomen_B.png}	\\
            \includegraphics[width=.20\textwidth, trim={0cm 19cm 19cm 1.5cm}, clip]{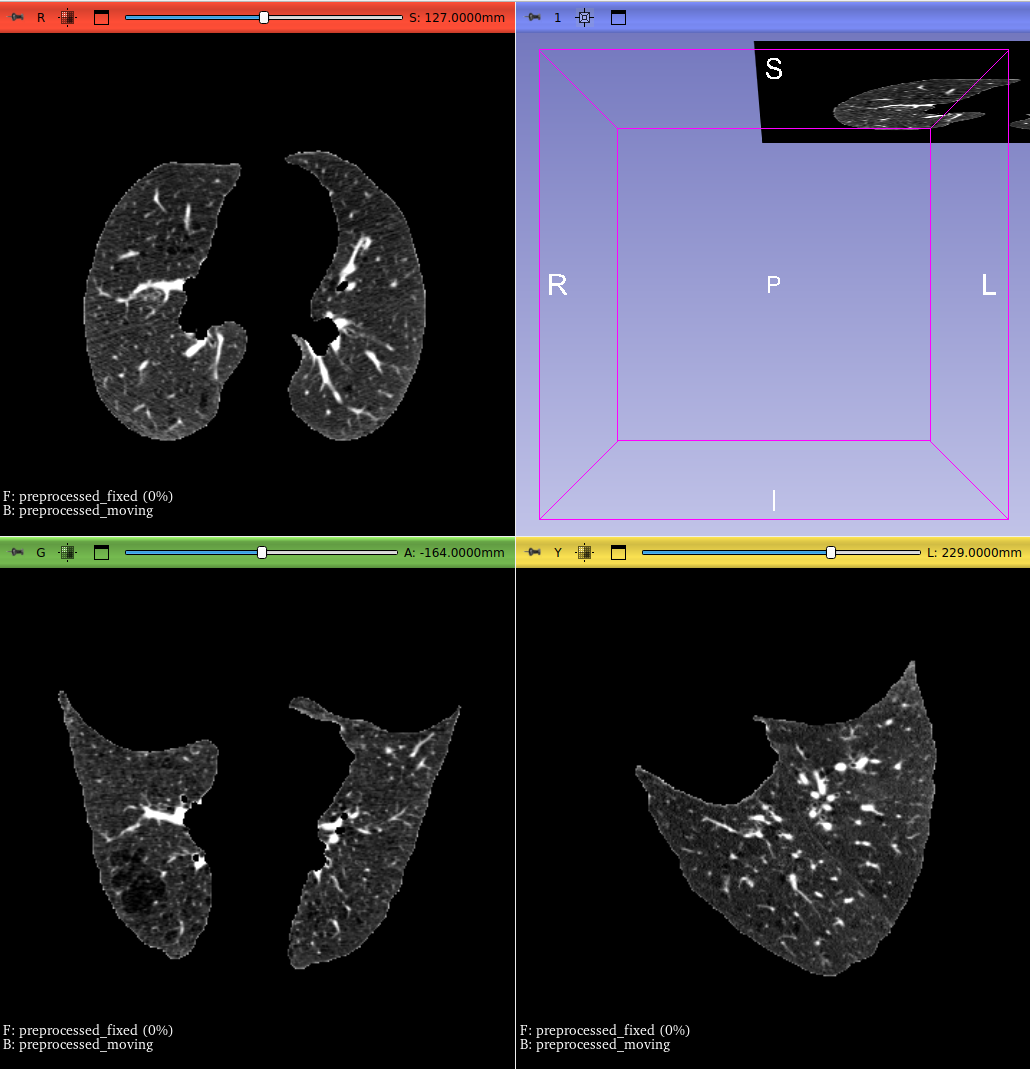}  &
			\includegraphics[width=.20\textwidth, trim={0cm 19cm 19cm 1.5cm}, clip]{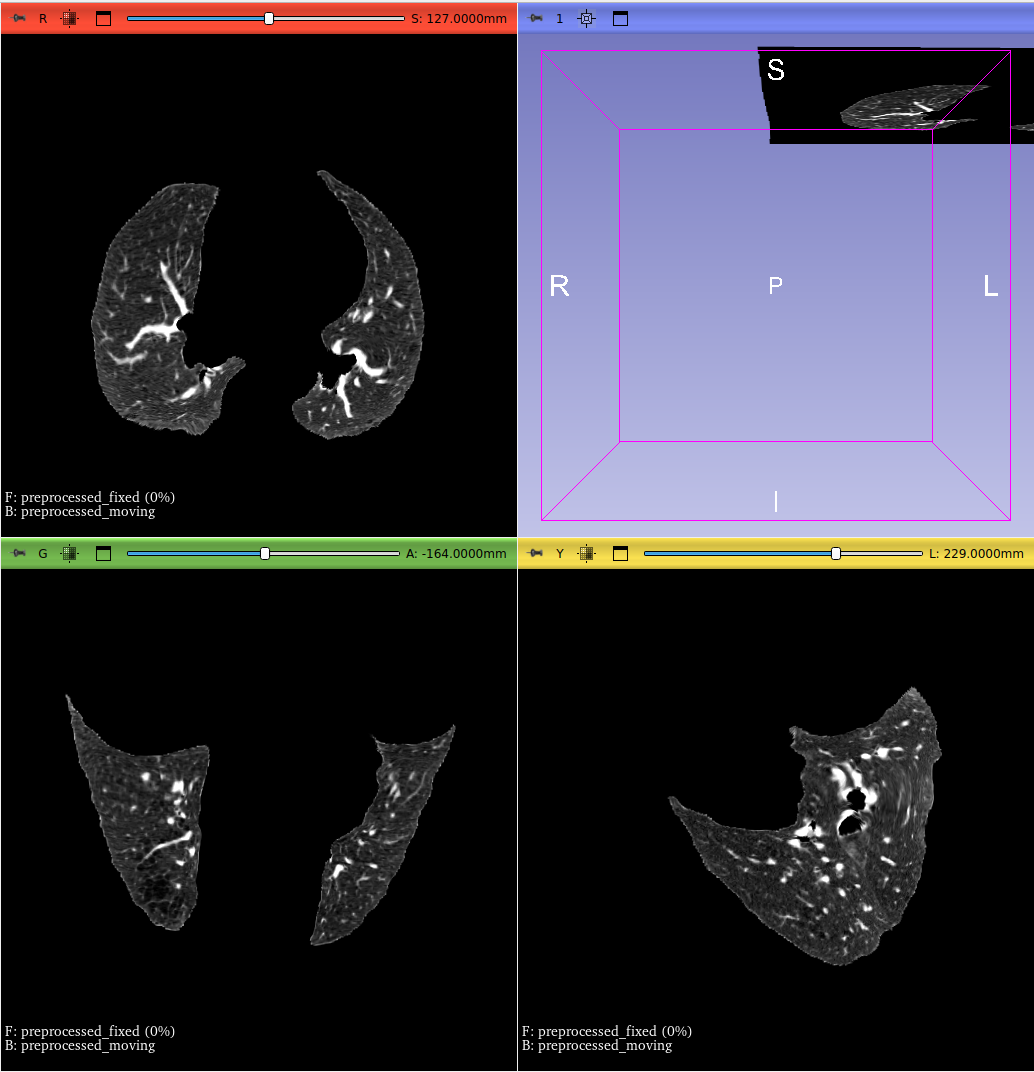} &
            \includegraphics[width=.20\textwidth, trim={0cm 19cm 19cm 1.5cm}, clip]{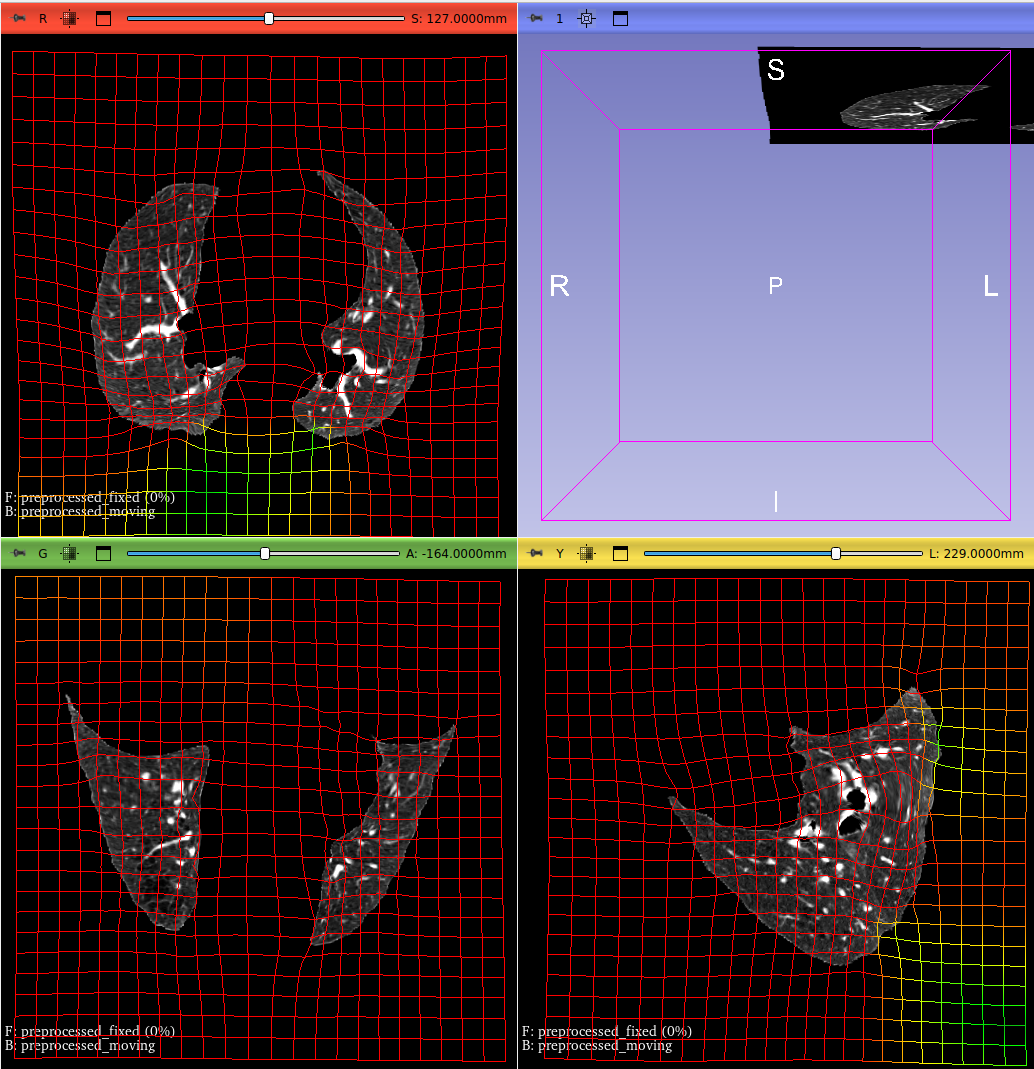} &
            \includegraphics[width=.20\textwidth, trim={0cm 19cm 19cm 1.5cm}, clip]{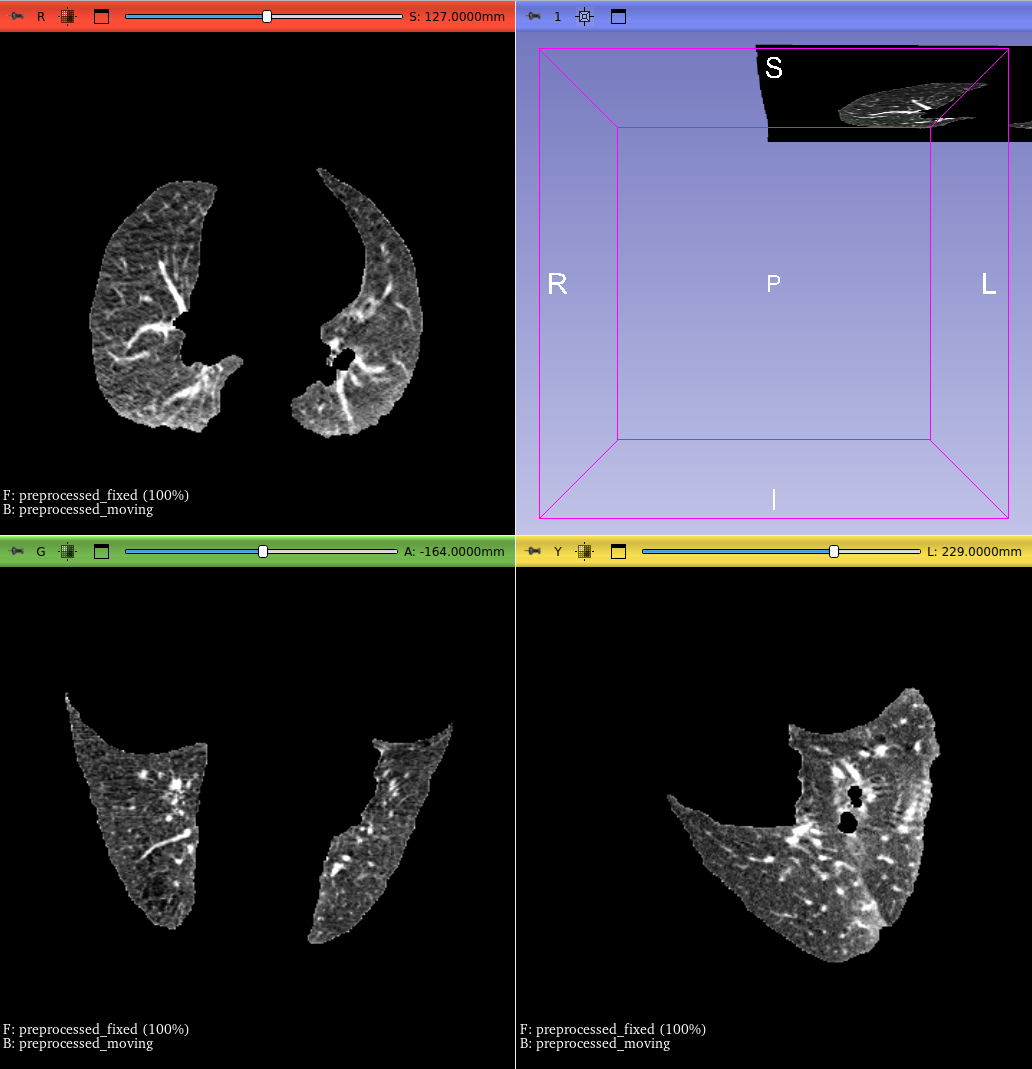}    \\
            \includegraphics[angle=180, origin=c, width=.20\textwidth, trim={0cm 2cm 19cm 20cm}, clip]{supfigs/lung_A.png}  &
            \includegraphics[angle=180, origin=c, width=.20\textwidth, trim={0cm 2cm 19cm 20cm}, clip]{supfigs/lung_warped_A.png} &
            \includegraphics[angle=180, origin=c, width=.20\textwidth, trim={0cm 2cm 19cm 20cm}, clip]{supfigs/lung_grid_A.png} &
            \includegraphics[angle=180, origin=c, width=.20\textwidth, trim={0cm 2cm 19cm 20cm}, clip]{supfigs/lung_B.png}    \\
            \includegraphics[angle=180, origin=c, width=.20\textwidth, trim={19cm 2cm 1cm 20cm}, clip]{supfigs/lung_A.png}  &
            \includegraphics[angle=180, origin=c, width=.20\textwidth, trim={19cm 2cm 1cm 20cm}, clip]{supfigs/lung_warped_A.png} &
            \includegraphics[angle=180, origin=c, width=.20\textwidth, trim={19cm 2cm 1cm 20cm}, clip]{supfigs/lung_grid_A.png} &
            \includegraphics[angle=180, origin=c, width=.20\textwidth, trim={19cm 2cm 1cm 20cm}, clip]{supfigs/lung_B.png}    \\
		\end{tabular}

	\caption{Detailed figures of our results on Abdomen1k (cases 00817 00872) and DirLAB case 1}
\end{figure}

\begin{figure}[htp]
	\centering

		\begin{tabular}{cccc}
			Moving Image & Warped (CARL\{ROT\} IO) & Grid (CARL\{ROT\} IO) & Fixed Image   \\ 
			\includegraphics[width=.20\textwidth, trim={0cm 19cm 19cm 1.5cm}, clip]{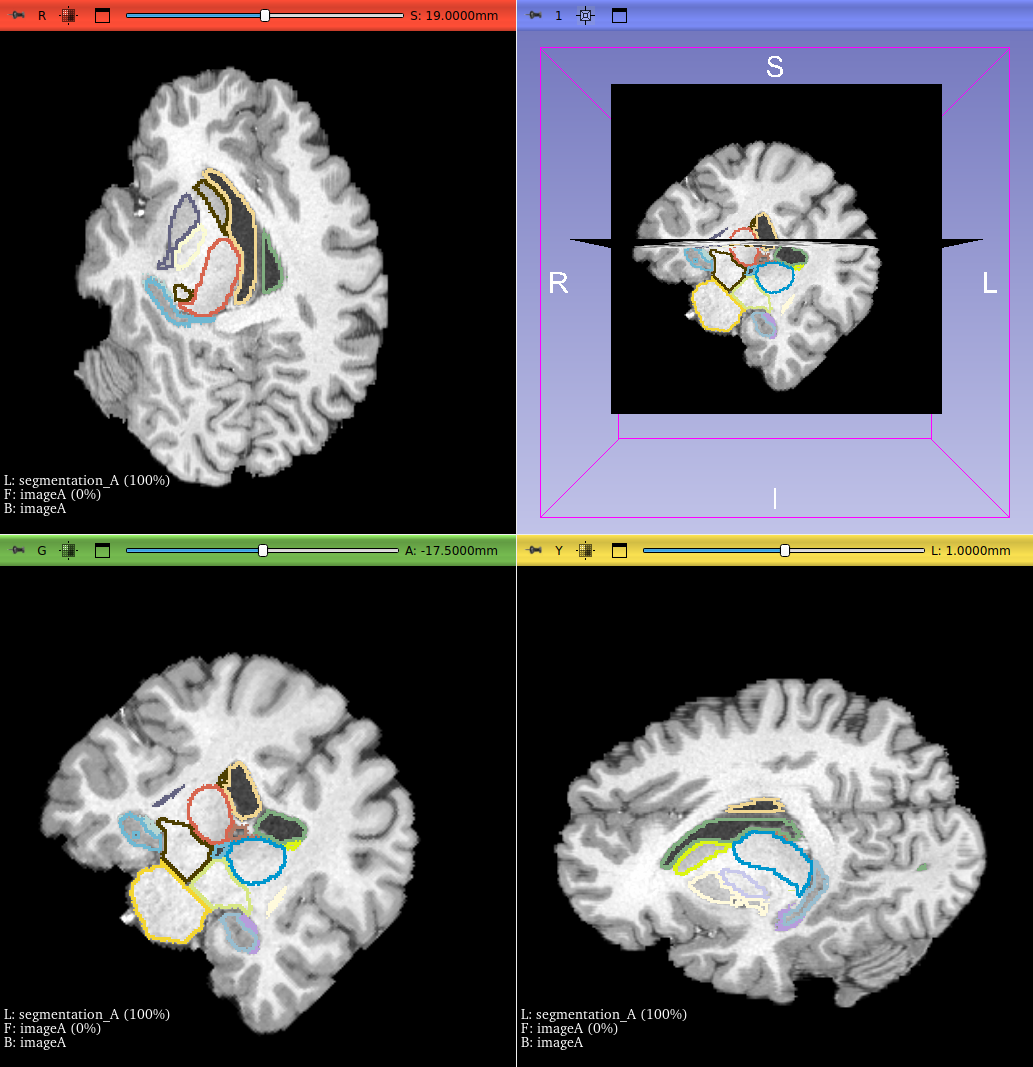}  &
			\includegraphics[width=.20\textwidth, trim={0cm 19cm 19cm 1.5cm}, clip]{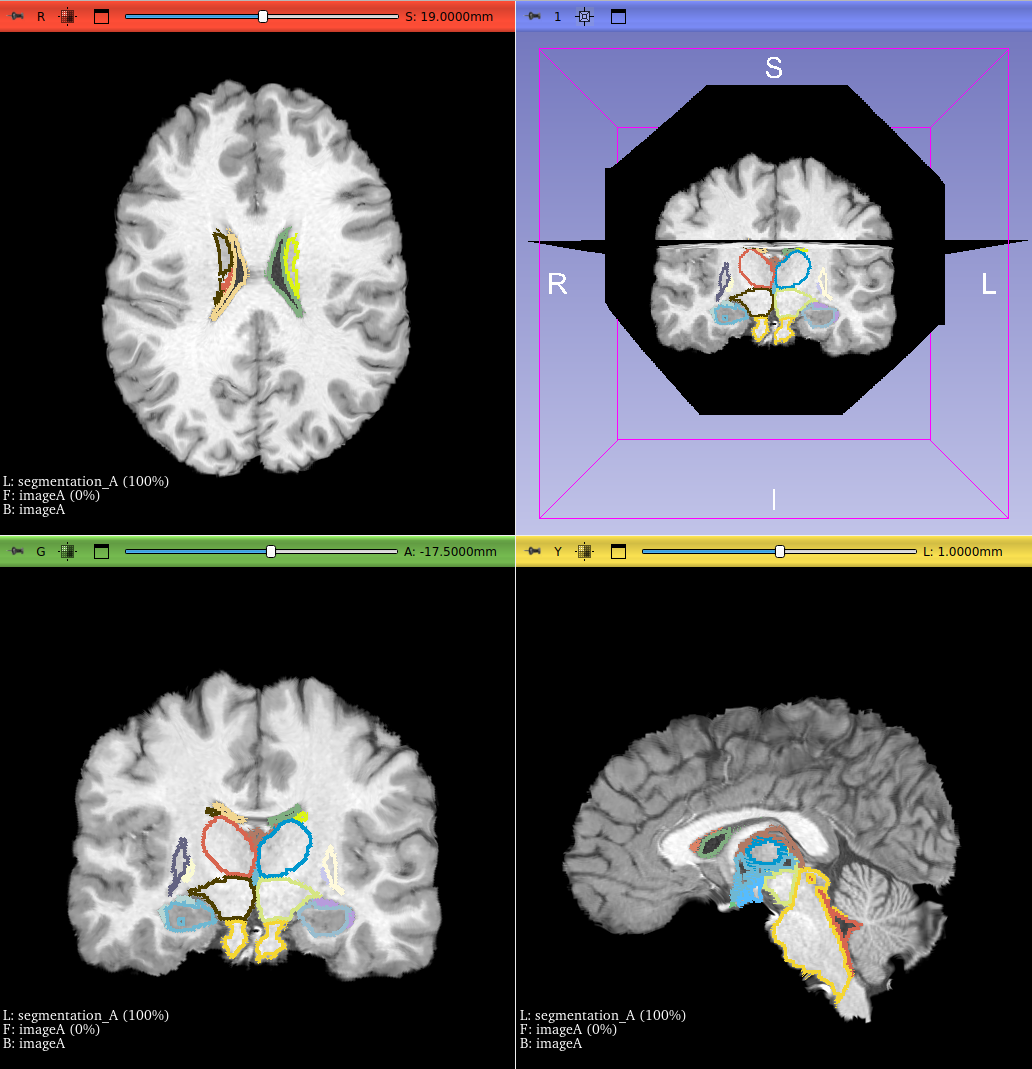} &
			\includegraphics[width=.20\textwidth, trim={0cm 19cm 19cm 1.5cm}, clip]{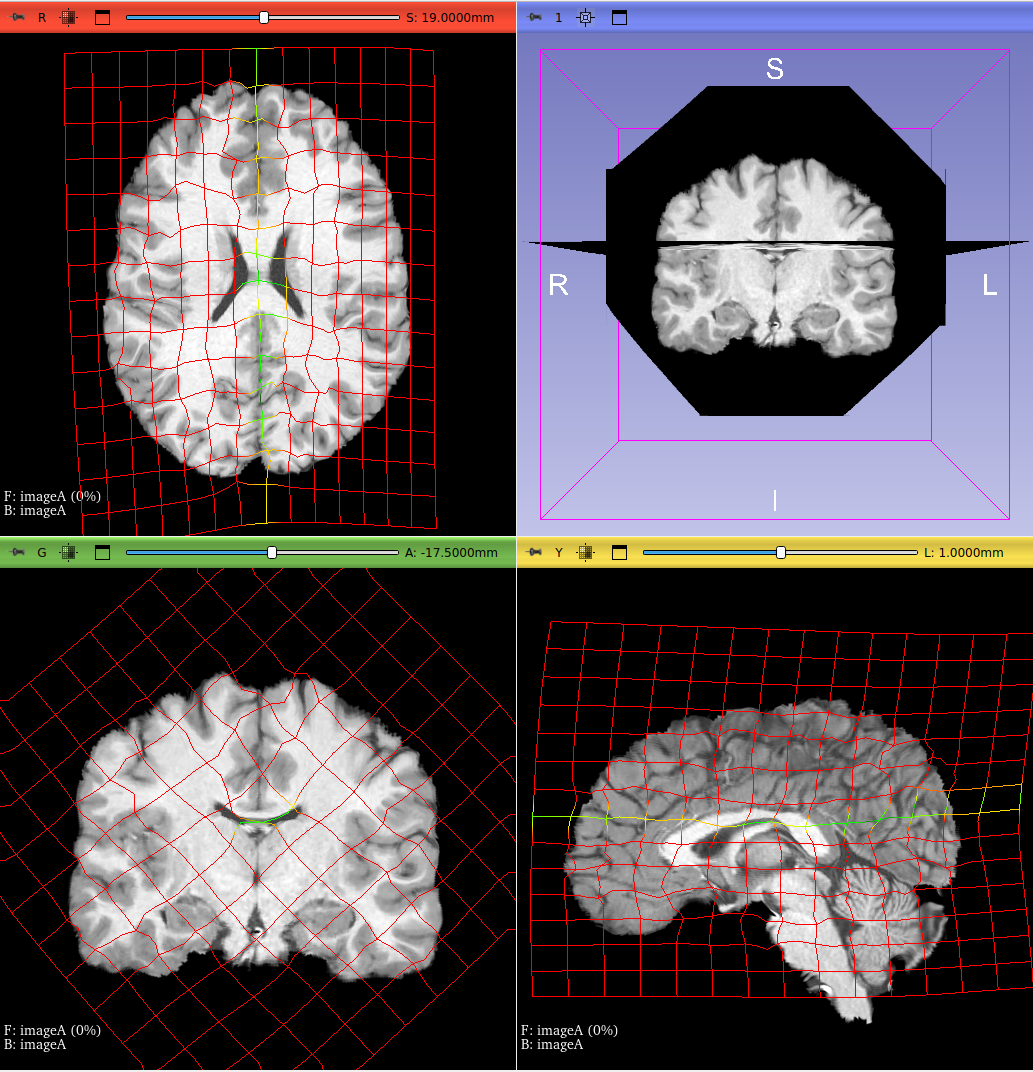} &
			\includegraphics[width=.20\textwidth, trim={0cm 19cm 19cm 1.5cm}, clip]{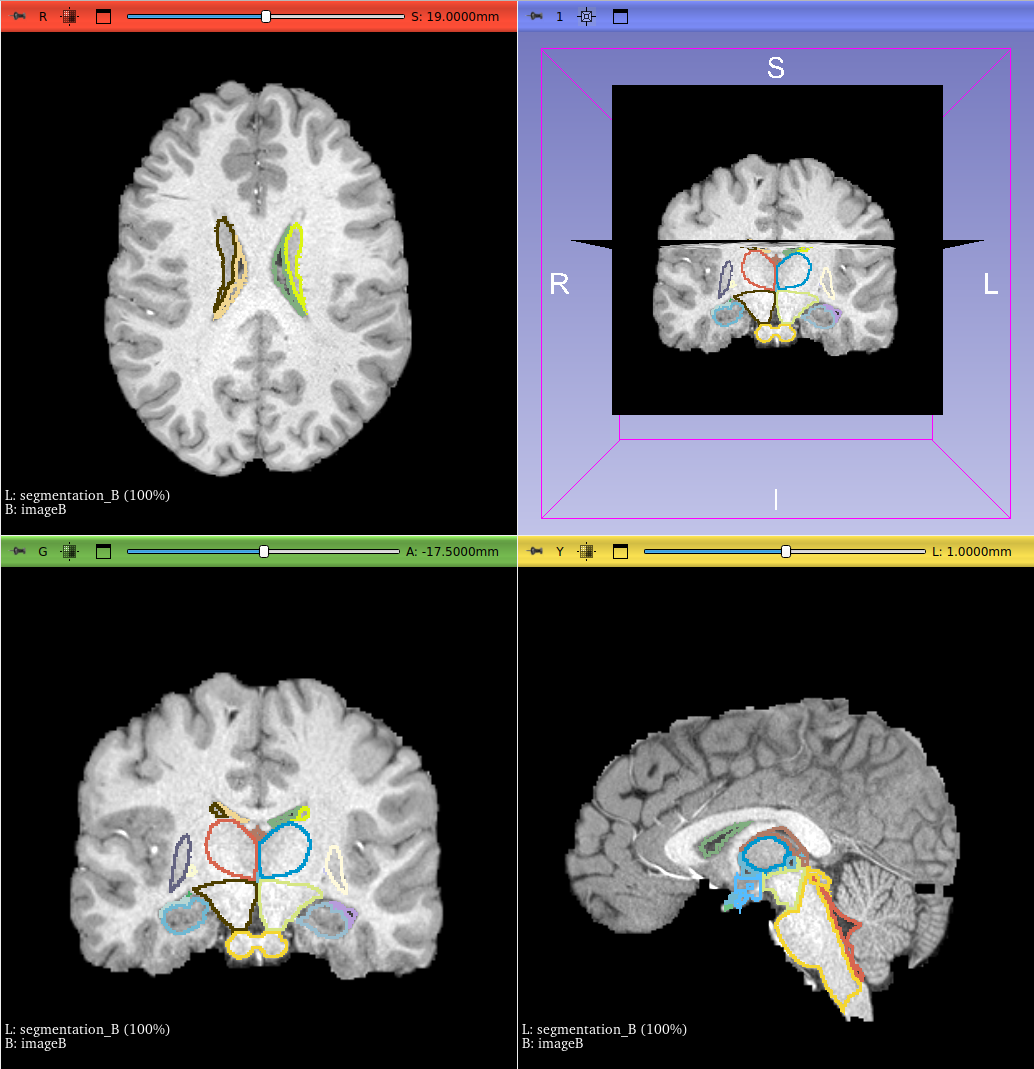}	\\
            \includegraphics[width=.20\textwidth, trim={0cm 1cm 19cm 20cm}, clip]{supfigs/hcp_A.png}  &
			\includegraphics[width=.20\textwidth, trim={0cm 1cm 19cm 20cm}, clip]{supfigs/hcp_warped_A.png} &
			\includegraphics[width=.20\textwidth, trim={0cm 1cm 19cm 20cm}, clip]{supfigs/hcp_grid_A.png} &
			\includegraphics[width=.20\textwidth, trim={0cm 1cm 19cm 20cm}, clip]{supfigs/hcp_B.png}	\\
            \includegraphics[width=.20\textwidth, trim={19cm 1cm 1cm 20cm}, clip]{supfigs/hcp_A.png}  &
			\includegraphics[width=.20\textwidth, trim={19cm 1cm 1cm 20cm}, clip]{supfigs/hcp_warped_A.png} &
			\includegraphics[width=.20\textwidth, trim={19cm 1cm 1cm 20cm}, clip]{supfigs/hcp_grid_A.png} &
			\includegraphics[width=.20\textwidth, trim={19cm 1cm 1cm 20cm}, clip]{supfigs/hcp_B.png}	\\
			Moving Image & Warped (CARL IO) & Grid (CARL IO) & Fixed Image   \\ 
            \includegraphics[width=.18\textwidth, trim={0cm 19cm 19cm 1.5cm}, clip]{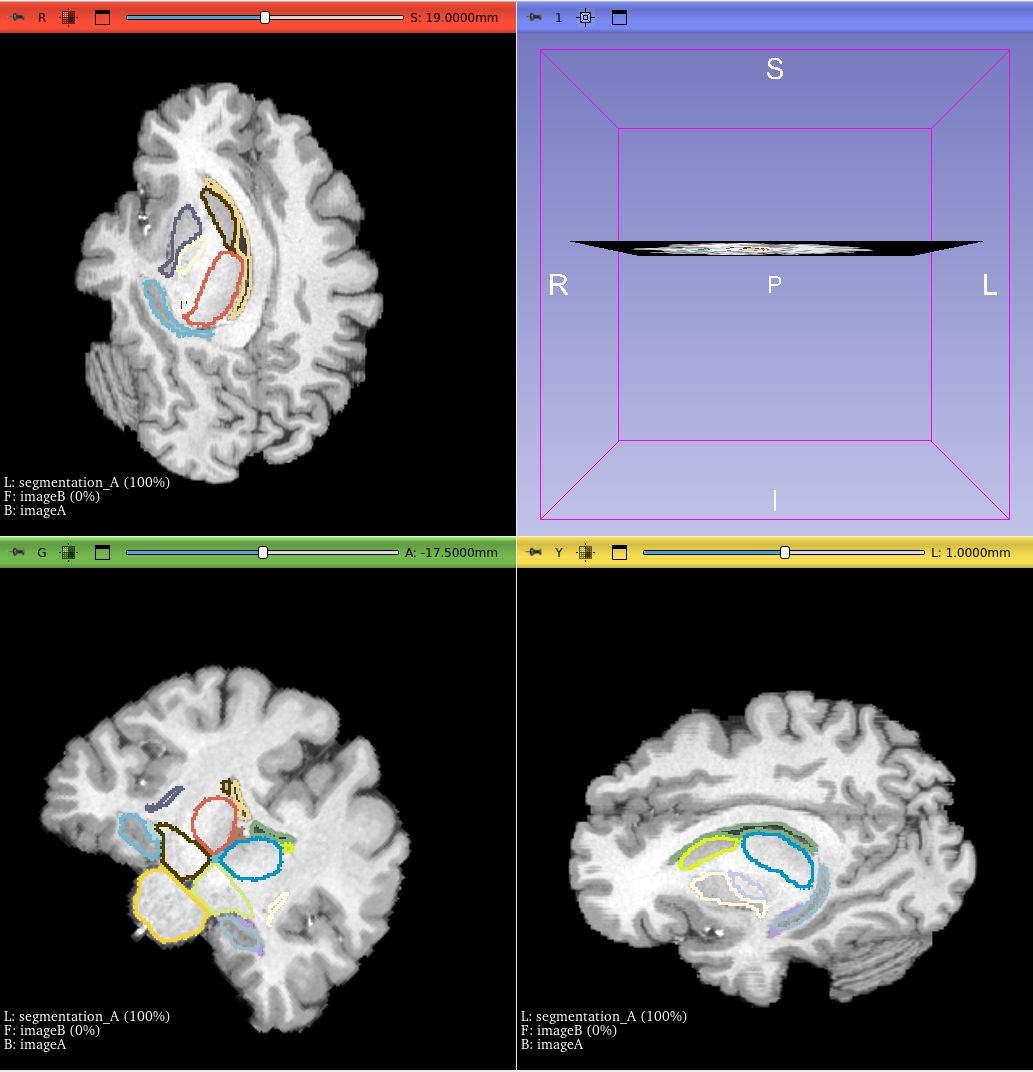}  &
			\includegraphics[width=.18\textwidth, trim={0cm 19cm 19cm 1.5cm}, clip]{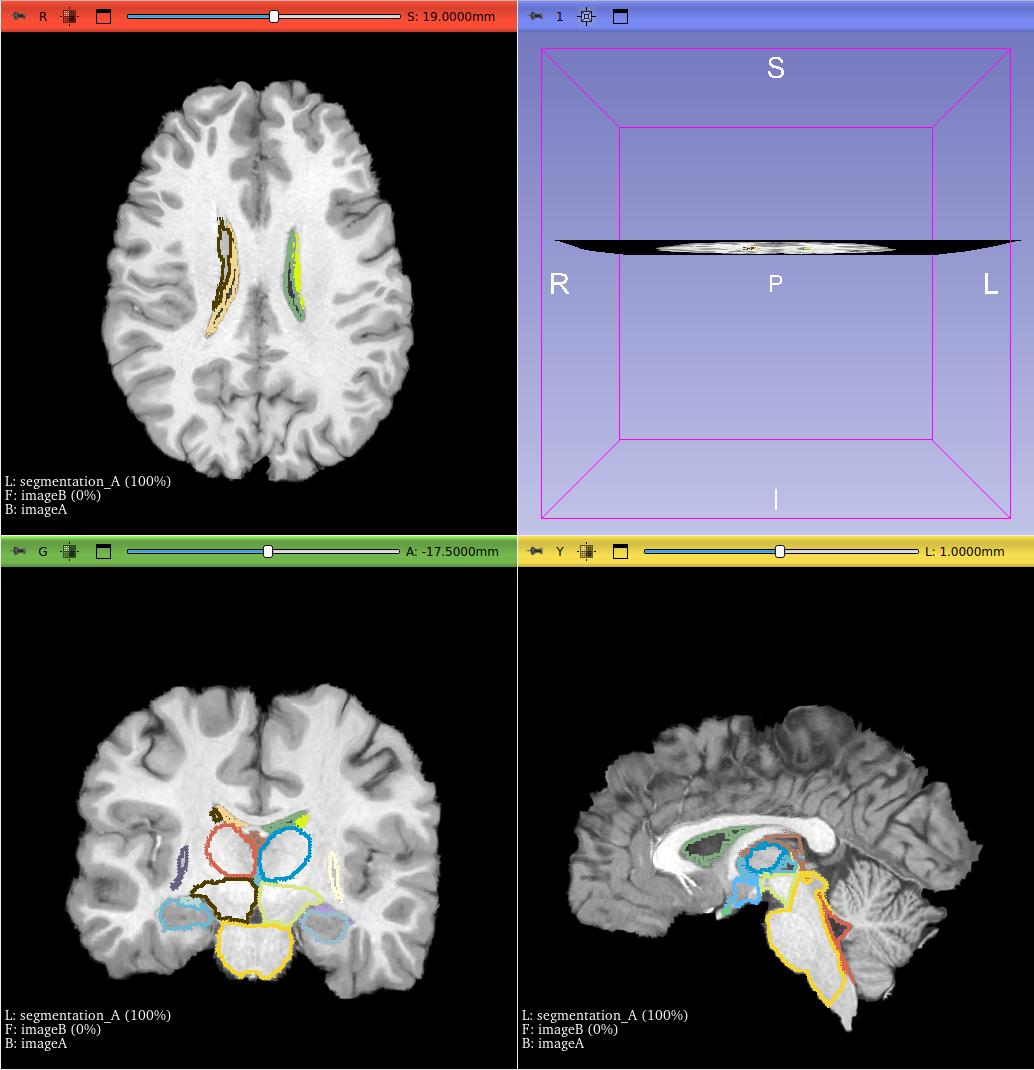} &
			\includegraphics[width=.18\textwidth, trim={0cm 19cm 19cm 1.5cm}, clip]{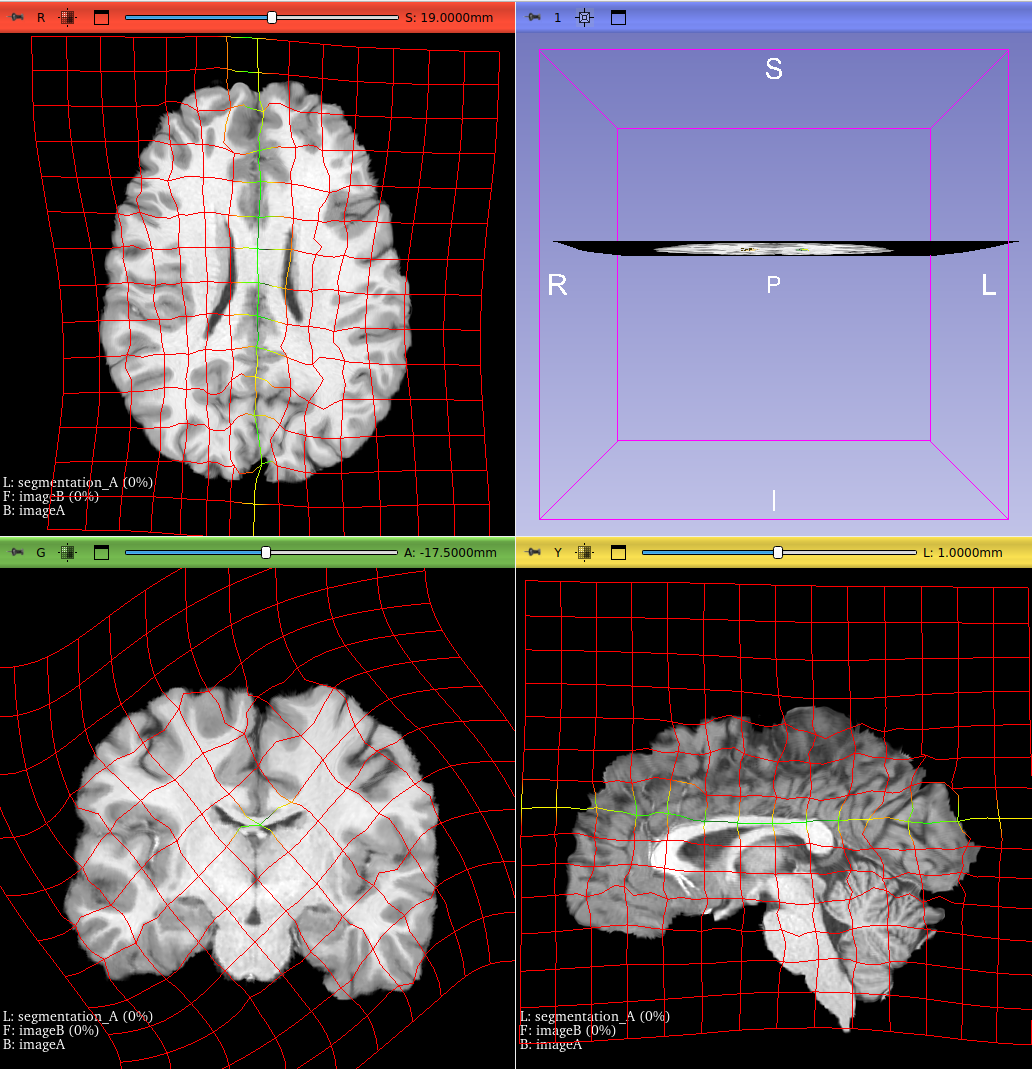} &
			\includegraphics[width=.18\textwidth, trim={0cm 19cm 19cm 1.5cm}, clip]{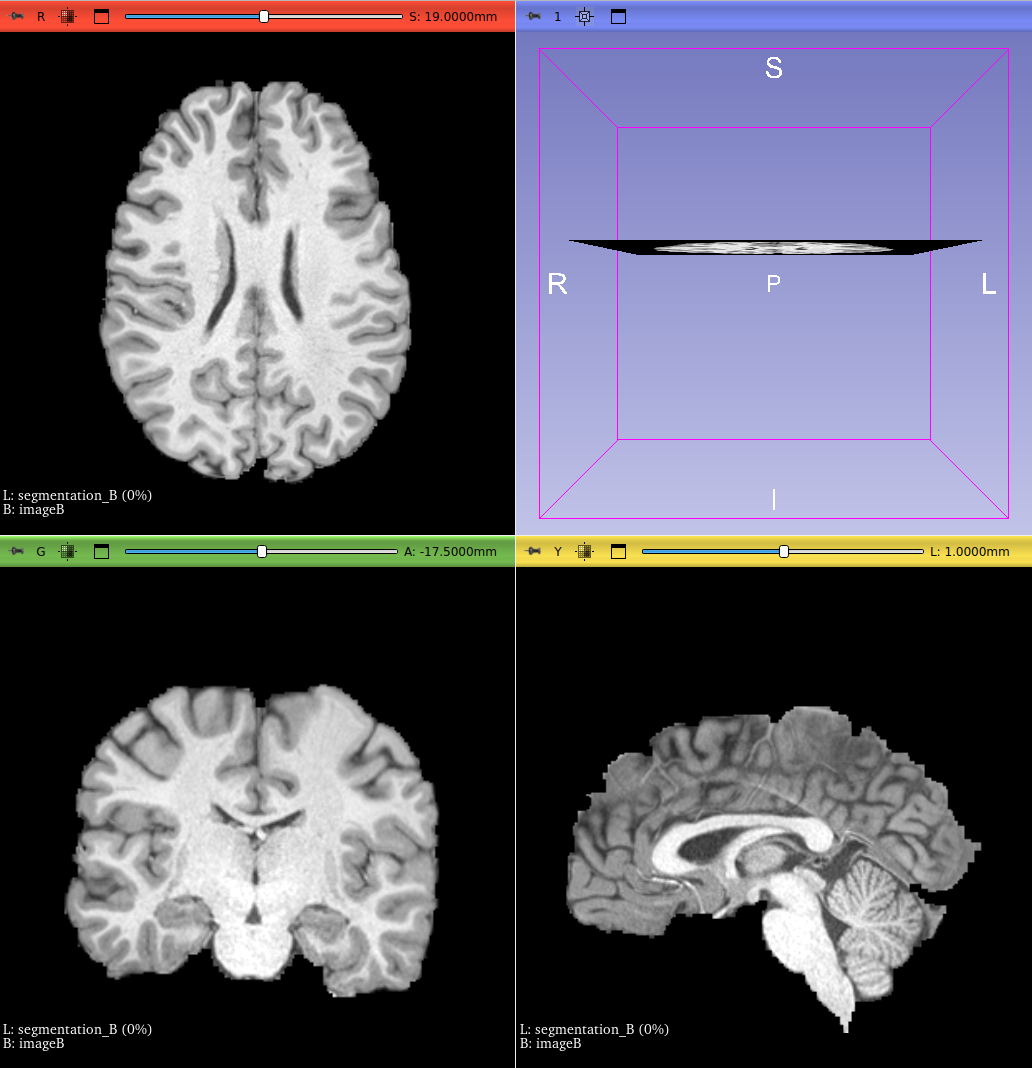}	\\
            \includegraphics[width=.18\textwidth, trim={0cm 1cm 19cm 20cm}, clip]{supfigs/hcp_norot_A.png}  &
			\includegraphics[width=.18\textwidth, trim={0cm 1cm 19cm 20cm}, clip]{supfigs/hcp_norot_warped_A.png} &
			\includegraphics[width=.18\textwidth, trim={0cm 1cm 19cm 20cm}, clip]{supfigs/hcp_norot_grid_A.png} &
			\includegraphics[width=.18\textwidth, trim={0cm 1cm 19cm 20cm}, clip]{supfigs/hcp_norot_B.png}	\\
            \includegraphics[width=.18\textwidth, trim={19cm 1cm 1cm 20cm}, clip]{supfigs/hcp_norot_A.png}  &
			\includegraphics[width=.18\textwidth, trim={19cm 1cm 1cm 20cm}, clip]{supfigs/hcp_norot_warped_A.png} &
			\includegraphics[width=.18\textwidth, trim={19cm 1cm 1cm 20cm}, clip]{supfigs/hcp_norot_grid_A.png} &
			\includegraphics[width=.18\textwidth, trim={19cm 1cm 1cm 20cm}, clip]{supfigs/hcp_norot_B.png}	\\
		\end{tabular}

	\caption{Detailed figures of our results on HCP with the moving image synthetically rotated by 45 degrees. Both CARL(IO) and CARL\{ROT\}(IO) handle a 45 degree rotation well. This is especially remarkable for CARL(IO), which is far out of its training distribution. 45 degrees is empirically the limit of CARL's tolerance for rotations, while CARL\{ROT\}'s DICE is unaffected by arbitrary rotations, as seen in \cref{fig:hcp_translation}. Also, observe that CARL\{ROT\}'s formal equivariance to rotation causes its deformation grid to move rigidly with the brain in the negative space surrounding it.}
\end{figure}

\begin{figure}[htp]
	\centering

		\begin{tabular}{cccc}
			Moving Image & Warped (GradICON IO) & Grid (GradICON IO) & Fixed Image   \\ 
			\includegraphics[width=.20\textwidth, trim={0cm 19cm 19cm 1.5cm}, clip]{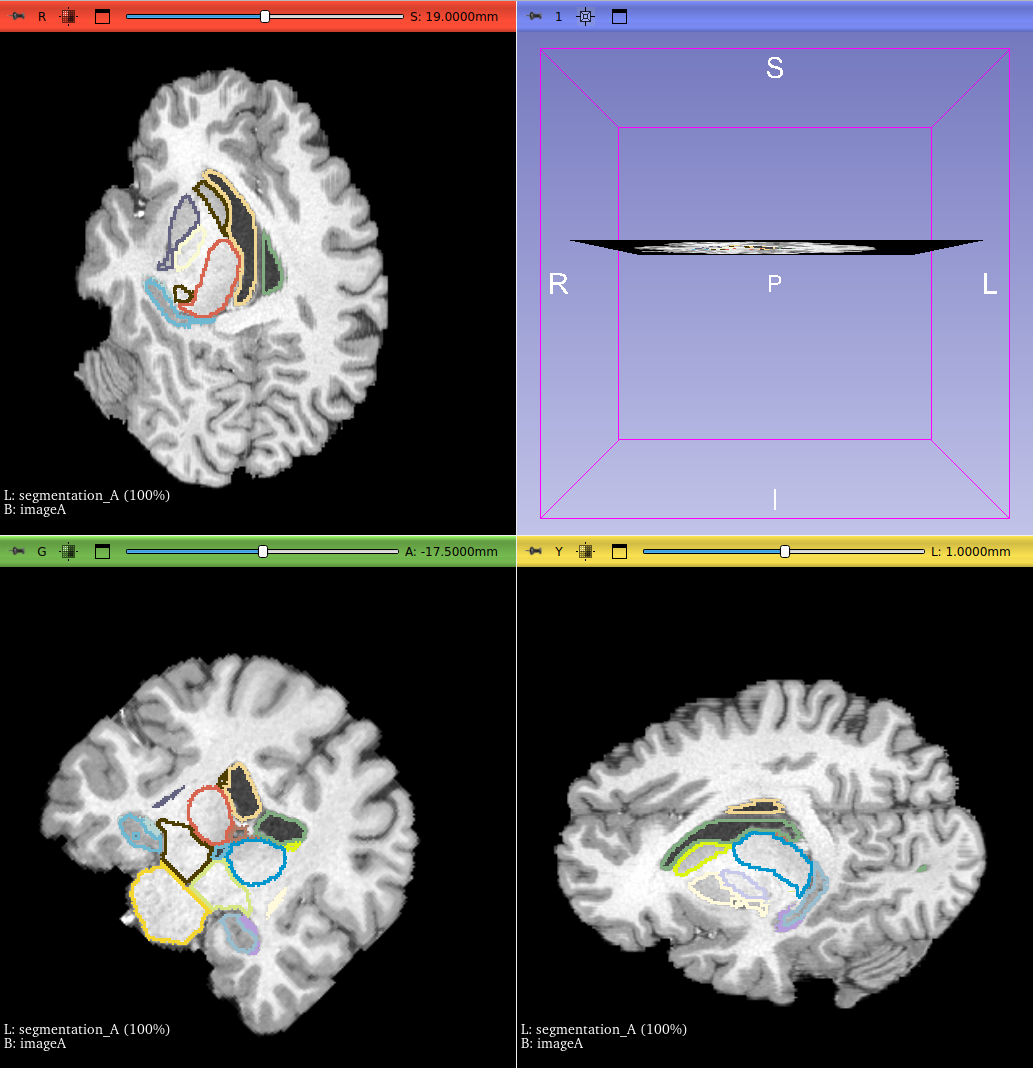}  &
			\includegraphics[width=.20\textwidth, trim={0cm 19cm 19cm 1.5cm}, clip]{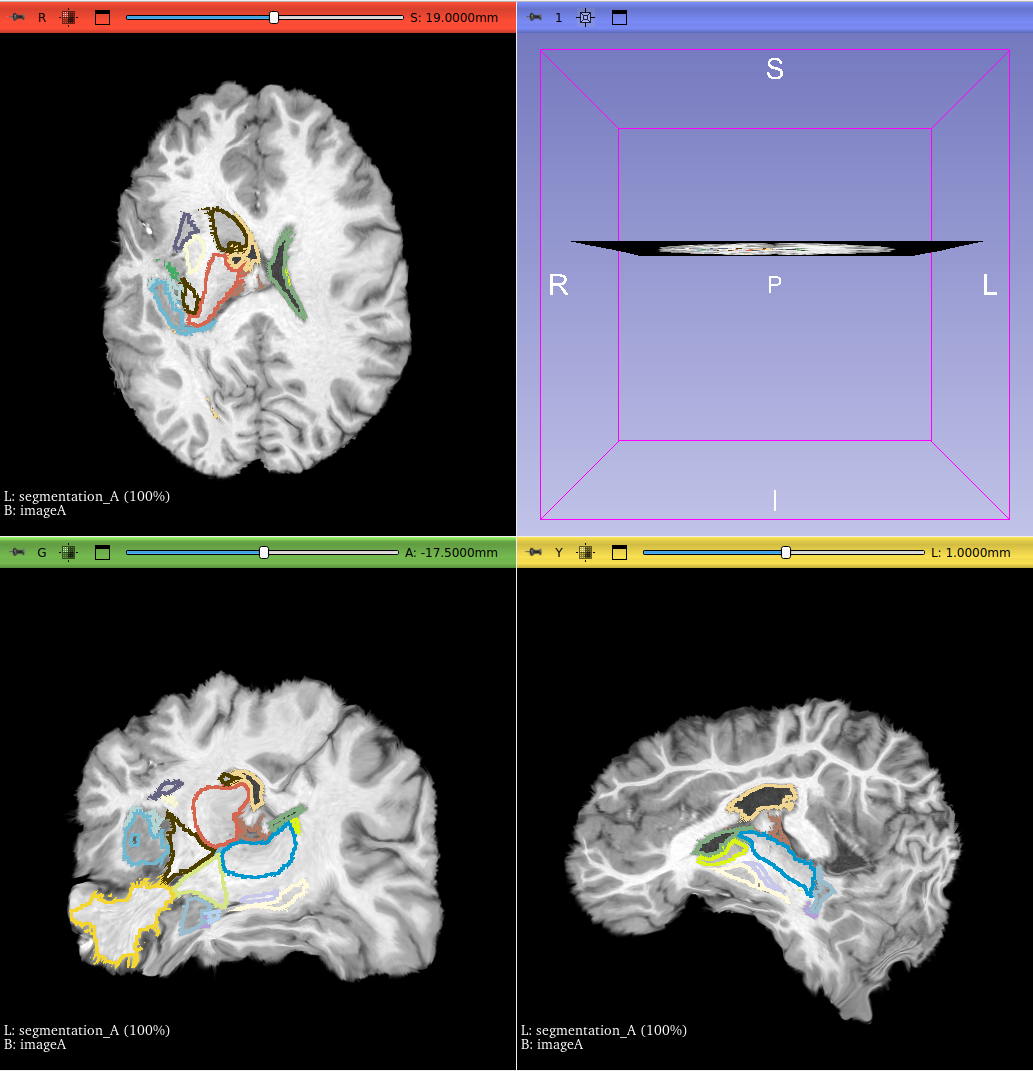} &
			\includegraphics[width=.20\textwidth, trim={0cm 19cm 19cm 1.5cm}, clip]{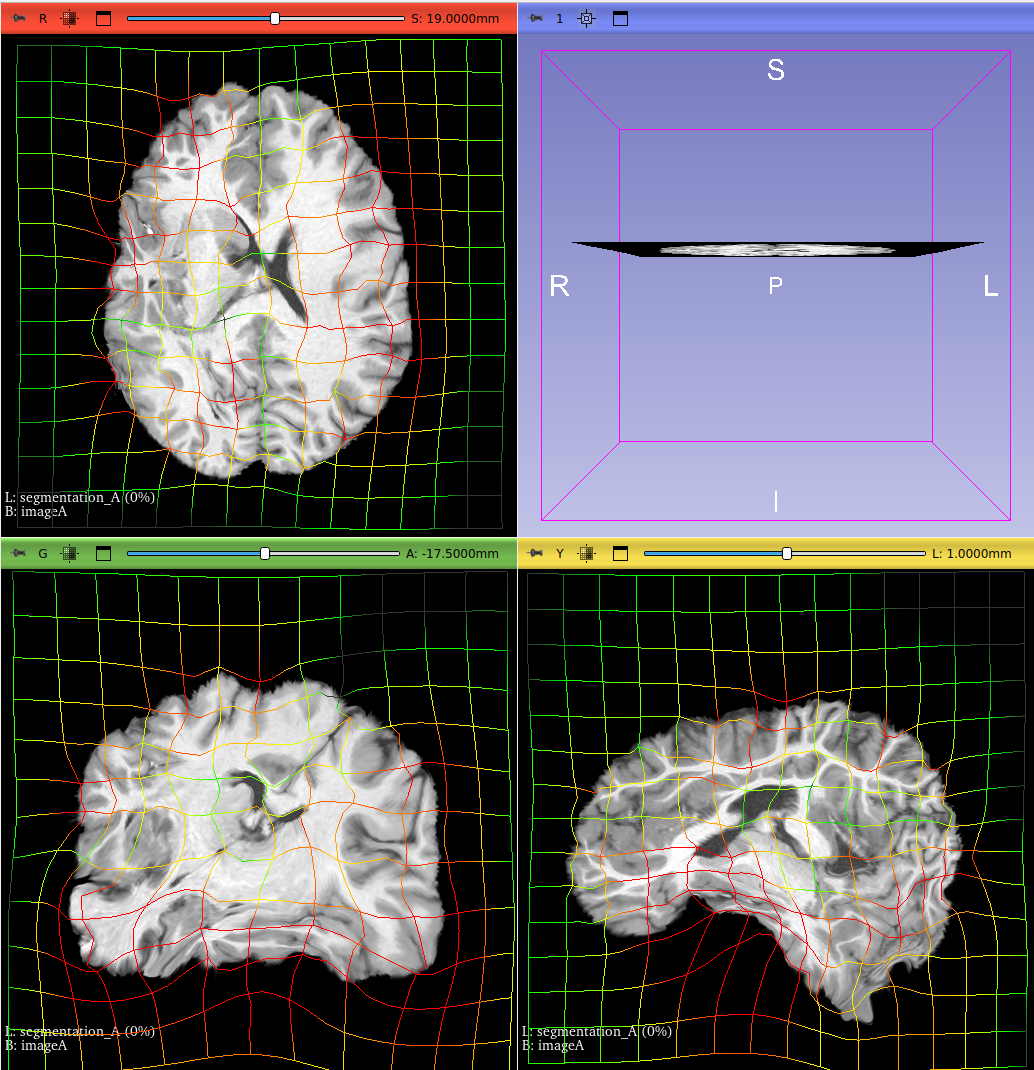} &
			\includegraphics[width=.20\textwidth, trim={0cm 19cm 19cm 1.5cm}, clip]{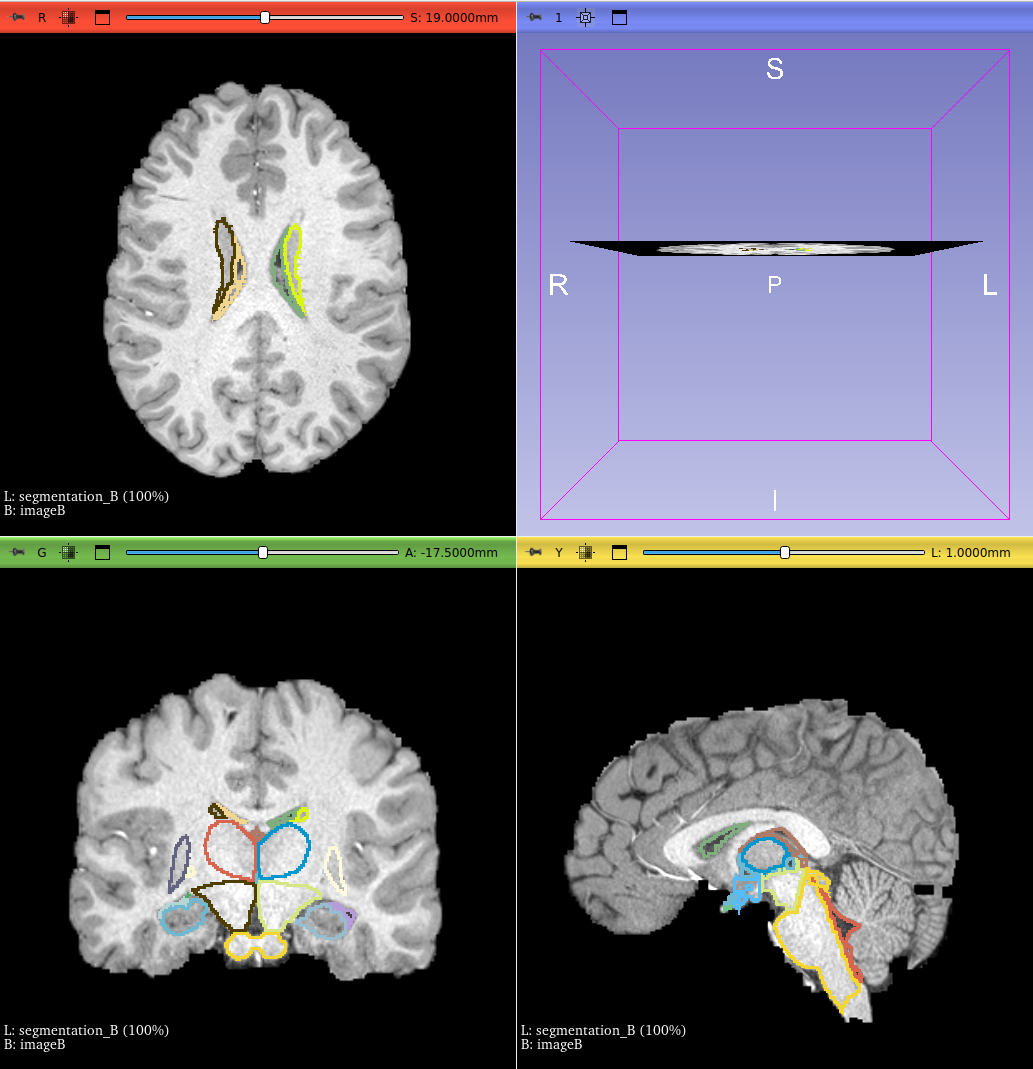}	\\
            \includegraphics[width=.20\textwidth, trim={0cm 1cm 19cm 20cm}, clip]{supfigs/gradicon_A.png}  &
			\includegraphics[width=.20\textwidth, trim={0cm 1cm 19cm 20cm}, clip]{supfigs/gradicon_warped_A.png} &
			\includegraphics[width=.20\textwidth, trim={0cm 1cm 19cm 20cm}, clip]{supfigs/gradicon_grid_A.png} &
			\includegraphics[width=.20\textwidth, trim={0cm 1cm 19cm 20cm}, clip]{supfigs/gradicon_B.png}	\\
            \includegraphics[width=.20\textwidth, trim={19cm 1cm 1cm 20cm}, clip]{supfigs/gradicon_A.png}  &
			\includegraphics[width=.20\textwidth, trim={19cm 1cm 1cm 20cm}, clip]{supfigs/gradicon_warped_A.png} &
			\includegraphics[width=.20\textwidth, trim={19cm 1cm 1cm 20cm}, clip]{supfigs/gradicon_grid_A.png} &
			\includegraphics[width=.20\textwidth, trim={19cm 1cm 1cm 20cm}, clip]{supfigs/gradicon_B.png}	\\
			
		\end{tabular}

	\caption{In contrast, GradICON cannot adapt to a 45 degree rotation which was outside its training distribution.}
\end{figure}
